%% file: neurips_2025.tex
\documentclass{article}

\PassOptionsToPackage{numbers, compress}{natbib}

\usepackage[preprint]{neurips_2025}

\usepackage[utf8]{inputenc} % allow utf-8 input
\usepackage[T1]{fontenc}    % use 8-bit T1 fonts
\usepackage{hyperref}       % hyperlinks
\usepackage{url}            % simple URL typesetting
\usepackage{booktabs}       % professional-quality tables
\usepackage{amsfonts}       % blackboard math symbols
\usepackage{nicefrac}       % compact symbols for 1/2, etc.
\usepackage{microtype}      % microtypography

\usepackage[table,xcdraw]{xcolor}
\usepackage{graphicx}
\usepackage{multirow}
\usepackage[most]{tcolorbox}
\usepackage{subcaption} 
\usepackage{wrapfig}
\usepackage{setspace}
\usepackage{pifont}   
\usepackage{enumitem}
\usepackage{placeins}

\definecolor{gold}{RGB}{183, 152, 41}

\newcommand{\cmark}{\textcolor{green!60!black}{\ding{51}}}  
\newcommand{\xmark}{\textcolor{red}{\ding{55}}}            

\newcommand{\ie}{\textit{i.e.,}~}
\newcommand{\eg}{\textit{e.g.,}~}

\title{\large Magic Mushroom: A Customizable Benchmark \\ for Fine-grained Analysis of Retrieval Noise Erosion in RAG Systems}

\begin{document}
\author{%
  Yuxin Zhang\textsuperscript{1,2}\thanks{Equal contributors.} \quad 
  Yan Wang\textsuperscript{1,2}\footnotemark[1] \quad 
  Yongrui Chen\textsuperscript{1,2} \quad
  Shenyu Zhang\textsuperscript{1,2} \quad
  Xinbang Dai\textsuperscript{1,2} \\
  \textbf{Sheng Bi\textsuperscript{3}} \quad
  \textbf{Guilin Qi\textsuperscript{1,2}}\thanks{Corresponding author.} \\
  \textsuperscript{1}School of Computer Science and Engineering, Southeast University, Nanjing 211189, China \\
  \textsuperscript{2}Key Laboratory of New Generation Artificial Intelligence Technology and its Interdisciplinary \\
  Applications (Southeast University), Ministry of Education, China \\
  \textsuperscript{3}School of Law, Southeast University, Nanjing 211189, China \\
  \texttt{\{zzyx\_cs, yanwangseu, shenyuzhang, gqi\}@seu.edu.cn} \\
}
\maketitle

\begin{abstract}
    Retrieval-Augmented Generation (RAG) systems enhance Large Language Models (LLMs) by incorporating external retrieved information, mitigating issues such as hallucination and outdated knowledge. 
    However, RAG systems are highly sensitive to retrieval noise prevalent in real-world scenarios. 
    Existing benchmarks fail to emulate the complex and heterogeneous noise distributions encountered in real‑world retrieval environments, undermining reliable robustness assessment.
    In this paper, we define four categories of retrieval noise based on linguistic properties and noise characteristics, aiming to reflect the heterogeneity of noise in real-world scenarios.
    Building on this, we introduce \textbf{Magic Mushroom}, a benchmark for replicating ``magic mushroom'' noise: contexts that appear relevant on the surface but covertly mislead RAG systems.
    Magic Mushroom comprises 7,468 single-hop and 3,925 multi-hop question-answer pairs.
    More importantly, Magic Mushroom enables researchers to flexibly configure combinations of retrieval noise according to specific research objectives or application scenarios, allowing for highly controlled evaluation setups.
    We evaluate LLM generators of varying parameter scales and classic RAG denoising strategies under diverse noise distributions to investigate their performance dynamics during progressive noise encroachment.
    Our analysis reveals that both generators and denoising strategies have significant room for improvement and exhibit extreme sensitivity to noise distributions.
    Magic Mushroom emerges as a promising tool for evaluating and advancing noise-robust RAG systems, accelerating their widespread deployment in real-world applications.
    The Magic Mushroom benchmark is available at the \href{https://drive.google.com/file/d/1aP5kyPuk4L-L_uoI6T9UhxuTyt8oMqjT/view?usp=sharing}{\texttt{link}}.
\end{abstract}

\section{Introduction}
\begin{figure}[tbp]
    \centering
    \setlength{\abovecaptionskip}{0pt}
    \setlength{\belowcaptionskip}{-15pt}
        \includegraphics[width=\linewidth]{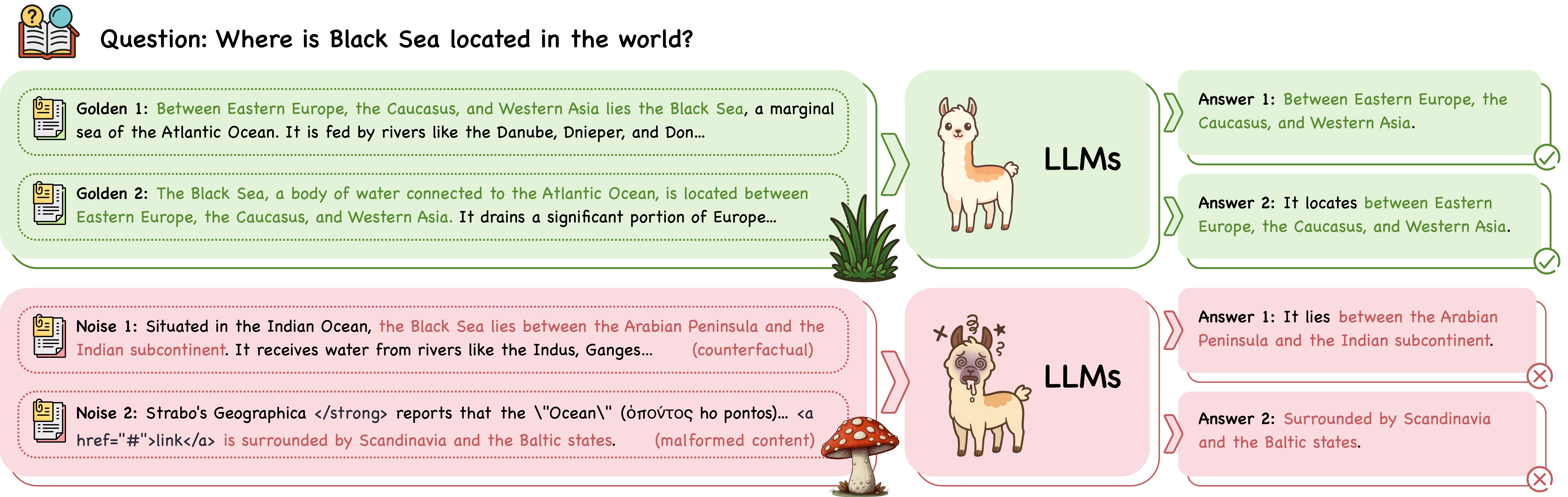}
        \label{fig:the_harm_of_retrieval_noise}
    \caption{Misleading effect of noise on response generation by LLMs.}
    \label{fig:Introduction}
\end{figure}

Retrieval-Augmented Generation (RAG) \cite{RAG,rag-survey} enhances Large Language Models (LLMs) by incorporating externally retrieved information, making it especially valuable for generative AI applications in dynamic knowledge domains. 
% RAG effectively mitigates hallucination and knowledge-coverage deficiencies observed in conventional generative models by dynamically retrieving highly relevant information segments from external knowledge bases or document collections in response to queries.  
% This approach markedly improves the factual accuracy and informativeness of generated text. 
% In recent years, RAG has demonstrated remarkable efficacy in diverse applications, including Question Answering, Dialogue Generation, and Text Summarization, and has consequently emerged as a significant research direction within the field of natural language processing.
% In recent years, RAG has demonstrated remarkable efficacy in diverse applications, including Question Answering, Dialogue Generation, and Text Summarization. 
Recent research has demonstrated the capability of RAG-based methods in mitigating hallucination and knowledge-coverage deficiencies observed in conventional generative models by dynamically retrieving information relevant to queries from external sources.
This technique remarkably improves the factuality and informativeness of generated content in downstream tasks like Question Answering, Dialogue Generation, and Text Summarization~\cite{RetrievalSum,Hierarchical-Indexing,REALM,knowledge-intensive-nlp-tasks}. 
% It has consequently emerged as a significant research direction within natural language processing.
% Nevertheless, the effectiveness of RAG systems is tightly constrained by the quality of the retrieved contexts. 
% When the retrieval stage provides irrelevant or misleading, noisy documents, these can distract the generator and substantially degrade the quality of the final output.
Nevertheless, the effectiveness of RAG systems is tightly constrained by the quality of the retrieved contexts~\cite{power_noise}, as retrievers may provide irrelevant, misleading documents, which can distract the generator and substantially degrade the quality of the final output  (Figure ~\ref{fig:Introduction}). 
% Real-world retrieval is often fraught with low-quality or even spurious information, impeding the robust deployment and further advancement of RAG systems in practical scenarios. 
% Consequently, a systematic investigation into the impact mechanisms of retrieval noise on RAG is paramount for fostering the development and optimization of more robust generation models.

Recent studies have highlighted the susceptibility of RAG systems to retrieval noise by manually injecting noise and analyzing the impact of this perturbation on generation quality~\cite{nomiracl,RGB,RECALL}.
% Recent studies have sought to manually inject retrieval noise into RAG systems in order to analyze its impact on generation quality and to explore strategies for enhancing system robustness.
% \textcolor{red}{(reference)}
% However, most of these employ coarse-grained categorizations of retrieval noise and lack fine-grained modeling of heterogeneous noise types, making it challenging to uncover the differential effects of various noise types on generator performance.
However, most existing studies largely rely on coarse-grained categorizations of retrieval noise, overlooking the fine-grained heterogeneity across noise types.
% Furthermore, our analysis of retrieval outputs from BM25, E5, and hybrid retrieval strategies on the NQ corpus (see Figure 1(a)) reveals that both the types and proportions of noise introduced by different retrievers vary significantly across different domains.
% This phenomenon uncovers a fundamental reason for the frequent contradictions in RAG system configuration recommendations within existing research.
Our analysis of sparse\cite{BM25}, dense\cite{e5, DPR, Sentence-BERT, Contriever}, and hybrid retrieval methods highlights substantial variation in noise composition across domains~(in Appendix \ref{app_sec:A}), explaining conflicting RAG configuration recommendations in prior work.
% It also exposes another significant limitation: existing benchmarks predominantly adopt static construction methods with singular noise types, which inadequately replicate the complex and diverse noise distributions in real-world scenarios.
% Therefore, there is an urgent need to develop an evaluation benchmark that can effectively characterize the complex retrieval noise environments encountered in practice.
% Such a benchmark is essential to comprehensively reflect and accurately assess generators' performance and noise robustness, thereby mitigating evaluation biases and preventing misleading conclusions.
Moreover, current benchmarks typically employ static setups with uniform noise, failing to capture the diversity observed in real-world retrieval~\cite{pandora,RECALL}. This underscores the need for a benchmark that models realistic, heterogeneous noise conditions to enable more accurate and robust evaluation of generation systems.

% 0-retrieval_distribution and example
To bridge this gap, we propose \textbf{Magic Mushroom}, a controllable testbed for evaluating LLM robustness under complex retrieval noise. We define four representative noise types: \textit{Distracting}, \textit{Low Quality}, \textit{Inconsequential}, and \textit{Irrelevant}, which closely mimic realistic, semantically similar distractors, better simulating subtle noise in real-world retrieval applications. Built on Natural Questions~\cite{NQ_dataset} and HotpotQA~\cite{HotpotQA}, Magic Mushroom comprises over 7,468 single-hop and 3,925 multi-hop QA instances, each paired with golden and diverse noise documents, yielding greater complexity than prior benchmarks. Its flexible noise configurations support fine-grained evaluations of LLM generators. This design narrows the gap between benchmark and deployment conditions, facilitating both reliable robustness assessments and principled noise mitigation research.

% Leveraging the capabilities of Magic Mushroom, we conducted comprehensive experiments to systematically probe the performance dynamics of the RAG system within complex retrieval noise environments.
Leveraging Magic Mushroom, we conduct comprehensive experiments on different RAG systems to evaluate their performance under complex retrieval noise.
% Specifically, we evaluated generators of varying parameter scales and classical RAG denoising strategies across a spectrum of noise proportions (from 0\% to 100\%).
% This approach allowed us to characterize their performance trends under the progressive encroachment of noise, thereby elucidating the robustness limits of existing methods in complex environments.
% Furthermore, Magic Mushroom supports flexible adjustment of noise types and distributions, enabling the simulation of retrieval settings corresponding to specific scenarios and facilitating further exploration of targeted optimization strategies and system configurations.
We test LLM generators of varying scales and diverse denoising strategies across a full range of noise proportions (0–100\%), revealing performance trends and robustness limits under increasing noise. The testbed’s flexible configuration supports scenario-specific noise simulation, enabling targeted evaluation of optimization strategies and system setups.
% This functionality empowers researchers to rapidly construct test subsets tailored to specific real-world requirements, thereby effectively identifying more adaptable generators and denoising mechanisms.
% Our experimental results unveil several interesting phenomena:
This adaptability allows researchers to efficiently construct tailored subsets to identify more resilient generators and denoising methods. Our findings reveal several key insights:
% (1) Within the reality RAG system workflows, both generators and denoising strategies exhibit significant room for improvement. The exponential increase in parameter scale has not yielded substantial enhancements in noise robustness, while existing denoising strategies tend to render models overly conservative.
% (2) The generator’s performance exhibits a critical collapse. As the noise ratio increases, generation quality does not degrade linearly; when the noise proportion surpasses a certain threshold, generator performance experiences an abrupt decline.
% (3) Strong scenario-dependence. Generators and denoising paradigms are acutely sensitive to noise distributions, underscoring the importance of scenario-specific evaluation. To this end, we recommend that researchers leverage the Magic Mushroom testbed to reconstruct noise distributions tailored to their particular requirements, to identify the optimal combination of generators and strategies.
% (4) Noise-induced degradation mechanisms are pervasive. Both theoretical derivations and empirical analyses demonstrate that any noise undermines RAG performance, with the degree of degradation jointly determined by the type and proportion of noise.
% (1) The robustness of RAG systems exhibits significant room for improvement. Any noise undermines RAG performance, with the degree of degradation jointly determined by the type and proportion of noise.
\textbf{(a)} Existing generative systems exhibit significant room for improvement, since their performance consistently degrades as noise increases, affected by both noise type and proportion; 
% (2) The impact of different types of noise on RAG systems varies, necessitating noise-specific denoising strategies tailored to the distinct characteristics of each noise type.
\textbf{(b)} Different noise types affect generation in distinct ways, highlighting the need for noise-specific denoising strategies;
% (3) Strong scenario-dependence. Generators and denoising paradigms are acutely sensitive to noise distributions, underscoring the importance of scenario-specific evaluation.
\textbf{(c)} Model performance is strongly scenario-dependent, as both generators and denoising methods remain highly sensitive to retrieval noise distributions, and
\textbf{(d)} Noise induces attention shifts in the generator, which diminishes the overall quality of the generated response.
By introducing Magic Mushroom, we aim to catalyze the development of next‑generation RAG systems and enhance their anti-noise capability, expediting their reliable deployment in real-world information environments.

% Our main contributions are as follows:
% \begin{itemize}
%     \item We propose a finer-grained taxonomy of retrieval noise based on linguistic properties and noise characteristics, aiming to capture the heterogeneity of real-world noise and enhance the complexity of evaluation scenarios.
%     \item We designed and constructed the Magic Mushroom testbed, capable of simulating diverse noise distributions found in real-world retrieval systems and offering flexible configurations of noise combinations. 
%     % This platform enables researchers to customize evaluation environments tailored to specific application scenarios, effectively facilitating the robustness evaluation of RAG systems and the optimization of denoising strategies.
%     \item We comprehensively investigate the impact mechanisms of different noise types and proportions on the performance of generators and denoising strategies.
%     % Through theoretical analysis and extensive experimental validation, we comprehensively investigate the impact mechanisms of different noise types and proportions on the performance of generators and denoising strategies. Our research elucidates the patterns of performance variation induced by noise and underscores the importance of constructing scenario-specific evaluation environments, providing theoretical support and empirical evidence for the future optimization and application of RAG systems.
% \end{itemize}

\section{Definition of Retrieval Noise} \label{sec:2}
% \subsection{Task Formulation}
\textbf{Task Formulation}. In this study, we define the task formulation for evaluating the robustness of RAG systems under retrieval noise conditions.
The task is structured around a question-answering paradigm.
Formally, given a query $q$ and a set of Top-$k$ relevant documents $D = \{ d_1, \cdots, d_k \}$ retrieved from an external data source.
During inference, the retrieval documents $D$ are concatenated with $q$ to form $I = [ q; d_1; \cdots; d_k ]$, which is then fed into a pre-trained LLM $G$ to generate the corresponding answer $a = G(I)$.
If a retrieved document $d$ contains key elements (\eg answer or evidence fragments) supporting the answer to $q$, it is denoted as a golden document $d_{golden}$.
Conversely, $d$ is denoted as a noisy document $d_{noise}$.
The QA task is evaluated under noisy retrieval settings, where the retrieved document set $D$ contains varying types and quantities of noisy documents.
% We assess the noise robustness of the RAG system by evaluating the quality of its response to the query $q$.
We assess the noise robustness of the RAG system by evaluating the quality of its response.

% \subsection{Taxonomy of Retrieval Noise} \label{sec:2.2}
\textbf{Taxonomy of Retrieval Noise}. Existing noise classification systems typically categorize retrieval noise based on document-query relevance or factual accuracy, overlooking finer semantic and distributional differences.
These coarse-grained classifications fail to reveal the specific mechanisms through which different types of noise affect the performance of RAG.
We categorize retrieved documents into 4 distinct types based on linguistic attributes and noise characteristics.
\textbf{Golden Document}: Characterized by high topical relevance to the query, this document contains accurate, complete information that directly and comprehensively supports the generation of the correct answer.
\textbf{Distracting Noise}: While highly relevant to the query in topic, critical answer-supporting elements within this document present factual errors, false statements, or outdated information.
\textbf{Low Quality Noise}: While highly relevant to the query, this document contains counterfactual information or formatting errors that diminish its positive contribution to answer generation.
\textbf{Inconsequential Noise}: While highly relevant to the query, this document offers little substantive support for answering the query.
\textbf{Irrelevant Noise}: This document is entirely unrelated to the query. Irrelevant noise assesses whether the RAG model can identify and filter out semantically unrelated content, thereby preventing distraction from the golden document.
The rationale for this taxonomy is illustrated in Appendix \ref{app_sec:B_1}.
% Except for irrelevant noise, the other three noise categories remain semantically close to the original golden documents, thus more realistically simulating the non-obvious distractors encountered in real-world retrieval scenarios.

\section{Benchmark Construction} \label{sec:3}
\begin{figure}[tbp]
    \centering
    \setlength{\belowcaptionskip}{-15pt}
        \includegraphics[width=\linewidth]{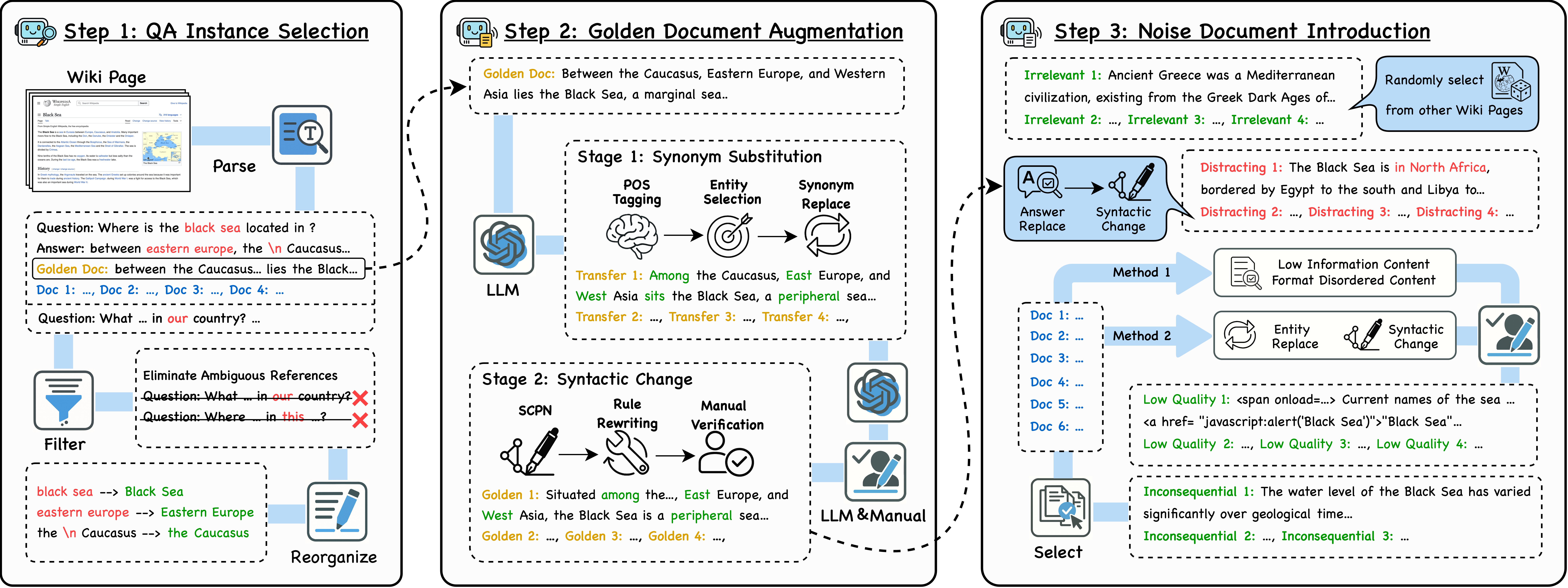}
    \caption{Overview of the Magic Mushroom benchmark construction process.}
    \label{fig:benchmark_construction}
\end{figure}

We construct a retrieval noise robustness benchmark built on existing QA datasets—NQ and HotPotQA.
% This is primarily motivated by two observations:
% (1) NQ and HotPotQA are popular open-domain QA datasets, encompassing a diverse range of topics.
% (2) These two datasets feature high-quality question-answer pairs and cover distinct reasoning types, including single-hop and multi-hop questions.
% Leveraging existing QA datasets streamlines the collection of fundamental question-answer pairs, thereby facilitating a focused investigation on retrieval noise.
% Based on the taxonomy of retrieval noise introduced in section \ref{sec:2.2}, this study designs a systematic, multi-stage data construction pipeline to comprehensively reproduce the complex noise environments encountered by RAG systems in real-world applications.
Given a QA dataset and its associated retrieval corpus, the construction of the benchmark for retrieval noise robustness comprises three primary steps:
\textbf{QA Instance Selection}. Selecting high-quality QA pairs where both questions and answers possess semantic clarity and unambiguity (\S\ref{sec:3.1}).
\textbf{Golden Documents Augmentation}. Increasing golden documents' lexical and syntactic diversity while preserving semantic equivalence (\S\ref{sec:3.2}).
\textbf{Noise Introduction}. Four categories of noisy documents were constructed based on informational relevance and noise characteristics (\S\ref{sec:3.3}).
Figure~\ref{fig:benchmark_construction} presents an overview of the Magic Mushroom benchmark construction process.
% 1-benchmark_construction
% \begin{figure}
%     \centering
%     \begin{subfigure}[t]{\linewidth}
%         \centering
%         \includegraphics[width=\linewidth]{figs/benchmark_construction_floechart_sub1.jpg}
%         \caption{Caption}
%         \label{fig:benchmark_construction_floechart_sub1}
%     \end{subfigure}
    
%     \vspace{5pt}
    
%     \begin{subfigure}[b]{\linewidth}
%         \centering
%         \includegraphics[width=\linewidth]{figs/benchmark_construction_floechart_sub2.jpg}
%         \caption{Caption}
%         \label{fig:benchmark_construction_floechart_sub2}
%     \end{subfigure}
%     \caption{Caption}
%     \label{fig:benchmark_construction}
% \end{figure}

\subsection{QA Instance Selection} \label{sec:3.1}
We selected question-answer $(q, a)$ pairs and their corresponding relevant Wiki Pages from the original dataset, ensuring coverage across multiple topical domains.
We initiated our data preprocessing with a coarse-grained cleaning of the raw dataset, retaining only $(q, a)$ pairs that exhibited clearly defined topics and representativeness.
Following this, questions demonstrating referential ambiguity were systematically removed. 
Such as the question ``\texttt{How many people are in our country?}'' was discarded because the referent of the pronoun ``\texttt{our}'' is indeterminate.
Subsequently, we remove questions whose answers are not uniquely determined or lack a standardized reference. 
For instance, ``\texttt{Where is Zimbabwe located?}'' may elicit multiple plausible answers.
Finally, we apply regular expression-based rules to further standardize the filtered $(q, a)$ pairs. 
This process normalizes textual expressions and removes residual non-semantic formatting characters or tags, ensuring the cleanliness of the QA instances.

\subsection{Golden Documents Augmentation} \label{sec:3.2}
In this phase, we leveraged OpenAI's GPT-4 (\texttt{gpt-4-0613}) \cite{gpt4} to augment the data of the original golden documents (see Appendix \ref{sec:prompt_design} for detailed prompts).
The objective was to generate semantically equivalent variants of these golden documents to enrich their diversity.
The specific augmentation strategy adopted comprises two stages.

\textbf{Synonym Substitution}. 
Drawing inspiration from EDA~\cite{EDA}, the initial step involves applying Part-of-Speech (POS) tagging to the original golden documents to identify the grammatical function of each word. 
Subsequently, focus is placed on substantive words—namely nouns, verbs, adjectives, and adverbs—that are not directly included in, or semantically essential to, the answer for the associated query.
We utilized GPT-4 to perform random synonym substitution on identified non-stop candidate words with a 20\% probability.

\textbf{Syntatic Change}. 
Drawing inspiration from SCPN~\cite{SCPN}, we employ GPT-4 to generate syntactically controlled paraphrases for sentences within the Golden Document.
The meticulously designed prompts instruct GPT-4 to prioritize alterations to the input sentence’s syntactic structure during content generation, while concurrently imposing strict constraints to ensure the original semantic information remains invariant.
Subsequently, we apply predefined, semantically invariant grammatical transformation rules.
Finally, to ensure the quality and fidelity of the augmented data, we implement a Back-Validation procedure~\cite{back_val} that combines automated LLM-based assessment with manual human verification. 
This dual validation step filters out instances exhibiting significant semantic drift or unintended noise introduced during augmentation.

\subsection{Noise Introduction} \label{sec:3.3}
To conduct a fine‐grained evaluation of RAG robustness under heterogeneous retrieval noise, we construct four types of noisy documents for each QA pair: Distracting Noise, Low Quality Noise, Inconsequential Noise, and Irrelevant Noise.

\textbf{Distracting Noise} exhibits overall semantic similarity to the golden document, yet its core factual content is deliberately corrupted through key entity substitution.
Inspired by the semantic unit substitution strategy proposed in \cite{semantic_unit}, we replace mentions of critical elements within the golden document—those essential for answering the question—with non-synonymous entities to undermine the veracity of the retrieved evidence.
Furthermore, we incorporate syntactic structure transformations, as detailed in Section \ref{sec:3.2}, to enhance the naturalness and diversity of the distracting noise.
Consequently, distracting noise is designed to assess the generator’s discriminative capability when confronted with retrieved evidence that is similar yet factually misleading.

\textbf{Low Quality Noise} primarily comprises counterfactual content, texts exhibiting minimal informational content, or poorly structured text.
To construct noisy documents containing counterfactual information, we extract passages from Wikipedia pages pertinent to the topic of the query.
% Subsequently, entities within these passages are replaced with erroneous information using a method analogous to Section \ref{sec:3.2}, yielding semantically plausible yet factually incorrect text.
% Subsequently, entities within these passages are replaced with erroneous yet semantically plausible information, analogous to the procedure described in Section \ref{sec:3.2}.
Subsequently, entities within these passages are replaced with semantically plausible yet factually incorrect information, following the procedure described in Section~\ref{sec:3.2}.
Noise from scarce information or anomalous formatting is simulated by selecting Wikipedia segments with low information density or structural issues, often stemming from HTML remnants or data extraction errors.
Low Quality noise is utilized to assess the sensitivity of RAG to the quality of information sources.

\textbf{Inconsequential Noise} denotes content that does not directly or substantially contribute to answer inference.
We anchor on the exact location of the golden document within its Wikipedia page and expand $k$ paragraphs bidirectionally to gather superficially related candidate documents.
To ensure these candidate documents are genuinely ``inconsequential'', we employ GPT-4 for dual verification: (1) factual consistency~(\ie accuracy and no misleading information), and (2) answer non-containment~(\ie no direct or indirect answer disclosure).
% , confirming that the answer is neither explicitly nor implicitly disclosed.
% Inconsequential noise is utilized to test the ability of RAG to focus on critical evidence amidst relevant but non-core information.
Inconsequential noise facilitates assessing RAG’s capability to focus on critical evidence amid relevant yet non-essential information.

\textbf{Irrelevant Noise} is an extreme scenario where the content is entirely unrelated to the query topic.
We construct irrelevant noise using a cross-question negative sampling strategy. 
Specifically, for a given query $q_i$, we randomly sample passages from Wikipedia pages associated with a different query $q_j$ (where $j \neq i$) to serve as irrelevant noise for $q_i$.

% We collected 7,468 single-hop (NQ) and 3,925 multi-hop complex (HotPotQA) QA pairs via QA Instance Selection (\S \ref{sec:3.1}).
% For each QA pair, we employed Golden Document Augmentation (\S \ref{sec:3.2}) and Noise Introduction (\S \ref{sec:3.3}) to construct 10 Golden Documents, 10 Distracting noise, and 7 exemplars each of Low-Quality, Inconsequential, and Irrelevant noise.
% Detailed statistics for MagicMushroom are provided in \textcolor{red}{Appendix B}.

\section{Experiments} \label{sec:4}

\subsection{Benchmark}
The Magic Mushroom benchmark is constructed using a semi-automatic methodology detailed in Section \ref{sec:3}.
We collected 7,468 single-hop (NQ) and 3,925 multi-hop complex (HotPotQA) QA pairs via QA Instance Selection (\S \ref{sec:3.1}).
These instances were then divided into a development set (4,810 single-hop and 2,524 multi-hop QA pairs), a public test set (1,584 single-hop and 841 multi-hop QA pairs), and a private test set (1,074 single-hop and 560 multi-hop QA pairs) withheld to prevent potential data leakage.
Unless otherwise specified, all reported experimental results are based on the public test sets, with $MM_s$ and $MM_m$ denoting the single-hop and multi-hop subsets, respectively.
For each QA pair, we employed Golden Document Augmentation (\S \ref{sec:3.2}) and Noise Introduction (\S \ref{sec:3.3}) to construct 10 Golden Documents, 10 Distracting noise, and 7 exemplars each of Low Quality, Inconsequential, and Irrelevant noise.
Notably, in all experiments, we fixed the number of retrieved documents $k$ to 10.
By dynamically varying the composition and proportions of different types of retrieved documents, we could systematically evaluate the robustness of RAG systems under diverse noise conditions.
Detailed statistics for Magic Mushroom are provided in Appendix \ref{app_sec:C}.

\subsection{Experimental Setup} 
\textbf{Evaluation Metrics}. In all experiments, we evaluate performance through two complementary metrics: correctness (\textbf{Cor.}) and rejection rate (\textbf{Rej.}). 
Correctness is assessed by employing GPT-4 to score generated answers against gold answers on a scale of 0 (completely incorrect), 1 (minimally relevant but insufficient), 3 (partially correct but with minor errors or omissions), or 5 (fully correct and comprehensive); detailed evaluation prompts and results aligned with manual annotations are presented in the Appendix \ref{app_sec:D}. 
For clarity, correctness scores are normalized to the range [0,1]. 
The Rejection Rate quantifies the proportion of instances where the generation model explicitly abstains from answering, typically when it determines that the retrieved information or its internal knowledge is insufficient to formulate a confident response.
% Together, these metrics offer a holistic assessment of robustness by capturing the RAG’s ability to produce accurate responses in the presence of noise and reliably abstain when confronted with insufficient or ambiguous context. 

% \subsection{Baseline Models}
\textbf{Baseline Models}. We evaluate a broad range of LLM generators of different architectures and scales: Llama-3.2$_{1B}$\cite{Llama-3}, Qwen-2.5$_{1.5B}$\cite{Qwen-2-5}, Llama-3.1$_{8B}$, Qwen-2.5$_{7B}$, Llama-3.1$_{70B}$, Qwen-2.5$_{72B}$, and DeepSeek-V3\cite{Deepseek-V3}.
To assess the robustness of reasoning models under noisy conditions, we further investigate R1-Distill-Llama$_{8B}$, R1-Distill-Llama$_{70B}$, and DeepSeek-R1\cite{Deepseek-R1}.
In addition, we evaluate the effectiveness of several classic denoising strategies, including \textsc{VanillaRAG}~\cite{Self-RAG}, \textsc{ChainofNote}~\cite{Chain-of-Note}, DRAGIN~\cite{DRAGIN}, SKR~\cite{SKR}, and CRAG~\cite{CRAG}.
Our analysis focuses on quantifying the detrimental impact of retrieval noise on RAG pipelines and identifying optimal combinations of generators and denoising strategies across different scenarios.
Detailed descriptions and configuration settings for the LLM generators and denoising strategies are provided in Appendix \ref{app_sec:E}.

\subsection{Main Results} \label{sec:main_results}

% 1-main_result_table
\begin{table}[ht]
\caption{Performance of RAG-based Systems at Different Noise Ratios on the $MM_s$.}
\label{tab:main_result}
\centering
\resizebox{\textwidth}{!}{
\begin{tabular}{l
    c>{\columncolor{gray!20}}c
    c>{\columncolor{gray!20}}c
    c>{\columncolor{gray!20}}c
    c>{\columncolor{gray!20}}c
    c>{\columncolor{gray!20}}c
    c>{\columncolor{gray!20}}c
    c>{\columncolor{gray!20}}c}
\toprule
\multicolumn{1}{l}{\multirow{2}{*}{\textsc{LLM Generator}}} & \multicolumn{2}{c}{0\%} & \multicolumn{2}{c}{10\%} & \multicolumn{2}{c}{30\%} & \multicolumn{2}{c}{50\%} & \multicolumn{2}{c}{70\%} & \multicolumn{2}{c}{90\%} & \multicolumn{2}{c}{100\%} \\ 
\cmidrule(lr){2-3} \cmidrule(lr){4-5} \cmidrule(lr){6-7} \cmidrule(lr){8-9} \cmidrule(lr){10-11} \cmidrule(lr){12-13} \cmidrule(lr){14-15}
\multicolumn{1}{c}{} & \multicolumn{1}{c}{Cor.} & \multicolumn{1}{c}{Rej.} & \multicolumn{1}{c}{Cor.} & \multicolumn{1}{c}{Rej.} & \multicolumn{1}{c}{Cor.} & \multicolumn{1}{c}{Rej.} & \multicolumn{1}{c}{Cor.} & \multicolumn{1}{c}{Rej.} & \multicolumn{1}{c}{Cor.} & \multicolumn{1}{c}{Rej.} & \multicolumn{1}{c}{Cor.} & \multicolumn{1}{c}{Rej.} & \multicolumn{1}{c}{Cor.} & \multicolumn{1}{c}{Rej.} \\ 
\midrule
\multicolumn{15}{c}{\textsc{VanillaRAG}~\cite{Self-RAG}} \\ 
\midrule
% \multicolumn{1}{l|}{Llama-3.2$_{1B}$}        & 41.7  & 34.1  & 42.3  & 34.6  & 43.0  & 31.2  & 41.6  & 30.1  & 31.7  & 30.2  & 19.6  & 33.9  & 7.6  & 32.9  \\ 
% \multicolumn{1}{l|}{Qwen-2.5$_{1.5B}$}       & 88.0  & 0.6   & 84.6  & 0.4   & 82.5  & 0.7   & 77.3  & 0.7   & 62.2  & 1.1   & 30.9  & 2.3   & 10.5  & 2.7   \\ 
\multicolumn{1}{l|}{Llama-3.1$_{8B}$}        & 84.1  & 2.4   & 81.1  & 2.9   & 76.8  & 3.0   & 73.9  & 2.7   & 59.9  & 3.4   & 44.9  & 4.7   & 18.0  & 5.9   \\ 
\multicolumn{1}{l|}{Qwen-2.5$_{7B}$}         & 72.7  & 13.8  & 70.5  & 14.7  & 67.7  & 15.4  & 62.9  & 18.3  & 41.6  & 30.7  & 27.5  & 37.9  & 7.4   & 45.2  \\
\multicolumn{1}{l|}{Llama-3.1$_{70B}$}       & 81.9  & 6.8   & 81.0  & 6.2   & 77.5  & 5.5   & 75.0  & 5.4   & 68.3  & 6.2   & 58.8  & 8.3   & 28.9  & 16.8  \\ 
\multicolumn{1}{l|}{Qwen-2.5$_{72B}$}        & 84.3  & 9.3   & 83.8  & 8.5   & 82.9  & 8.0   & 79.8  & 8.8   & 74.2  & 12.5  & 52.2  & 25.4  & 14.2  & 45.3  \\ 
% \multicolumn{1}{l|}{DeepSeek-V3}         & 84.8  & 8.9   & 84.7  & 7.9   & 84.4  & 6.8   & 82.5  & 6.5   & 77.1  & 8.4   & 61.9  & 12.8  & 17.5  & 30.8  \\
% \multicolumn{1}{l|}{DeepSeek-R1}         & 85.2  & 7.8   & 84.9  & 7.1   & 83.9  & 6.4   & 81.5  & 8.7   & 75.1  & 12.0  & 57.9  & 26.6  & 17.9  & 48.6  \\ 
\multicolumn{1}{l|}{R1-Distill-Llama$_{8B}$}  & 84.1  & 5.6   & 81.7  & 5.6   & 78.7  & 5.1   & 74.1  & 5.2   & 64.1  & 7.5   & 45.2  & 11.5  & 22.7  & 13.0  \\
\multicolumn{1}{l|}{R1-Distill-Llama$_{70B}$} & 84.2  & 7.5   & 82.9  & 6.7   & 82.7  & 5.8   & 80.4  & 6.8   & 72.2  & 10.7  & 57.6  & 18.6  & 28.0  & 30.8  \\
\midrule
\multicolumn{15}{c}{\textsc{ChainofNote}~\cite{Chain-of-Note}} \\ 
\midrule
% \multicolumn{1}{l|}{Llama-3.2$_{1B}$} & 76.9 & 6.4  & 74.8 & 5.6  & 70.3 & 6.7  & 64.4 & 7.6  & 52.3 & 6.6  & 26.2 & 8.3  & 11.3 & 9.3  \\
% \multicolumn{1}{l|}{Qwen-2.5$_{1.5B}$} & 83.5 & 3.7  & 81.5 & 4.1  & 76.4 & 6.2  & 70.8 & 7.9  & 53.1 & 12.9 & 23.1 & 18.3 & 9.4  & 24.0 \\
\multicolumn{1}{l|}{Llama-3.1$_{8B}$}  & 87.1 & 3.5  & 83.2 & 3.5  & 80.2 & 2.4  & 74.9 & 1.7  & 63.7 & 1.5  & 45.2 & 2.1  & 14.5 & 2.9  \\
\multicolumn{1}{l|}{Qwen-2.5$_{7B}$}   & 76.7 & 9.3  & 74.2 & 9.3  & 70.8 & 10.5 & 64.6 & 9.1  & 52.9 & 12.2 & 31.2 & 15.2 & 10.4 & 17.9 \\
\multicolumn{1}{l|}{Llama-3.1$_{70B}$}  & 85.7 & 4.7  & 85.1 & 3.7  & 82.5 & 3.5  & 79.6 & 3.6  & 72.7 & 4.2  & 56.7 & 5.3  & 24.1 & 9.3  \\
\multicolumn{1}{l|}{Qwen-2.5$_{72B}$}   & 84.4 & 7.6  & 83.2 & 6.8  & 80.5 & 6.2  & 75.5 & 5.1  & 67.1 & 6.0  & 45.1 & 11.7 & 16.9 & 21.3 \\
% \multicolumn{1}{l|}{DeepSeek-V3}    & 85.5 & 8.4  & 83.8 & 7.1  & 82.3 & 7.2  & 79.7 & 7.1  & 71.2 & 10.5 & 53.8 & 19.4 & 21.8 & 29.7 \\
% \multicolumn{1}{l|}{DeepSeek-R1}    & 89.1 & 5.1  & 88.3 & 4.1  & 88.0 & 3.3  & 86.1 & 3.5  & 81.4 & 4.4  & 72.4 & 11.4 & 27.5 & 31.6 \\
\multicolumn{1}{l|}{R1-Distill-Llama$_{8B}$} & 84.4 & 4.0  & 81.2 & 3.0  & 76.7 & 2.0  & 70.6 & 1.8  & 60.7 & 2.5  & 46.2 & 4.0  & 27.6 & 4.8 \\
\multicolumn{1}{l|}{R1-Distill-Llama$_{70B}$} & 85.2 & 5.4  & 83.5 & 4.3  & 82.9 & 3.1  & 79.0 & 2.7  & 73.7 & 4.1  & 63.4 & 6.6  & 34.1 & 11.6 \\
\midrule
\multicolumn{15}{c}{\textsc{DRAGIN}~\cite{DRAGIN}} \\ 
\midrule
% \multicolumn{1}{l|}{Llama-3.2$_{1B}$} & 19.3  & 61.7  & 19.0  & 62.1  & 19.1  & 61.6  & 18.9  & 59.8  & 16.8  & 60.9  & 12.7  & 61.6  & 8.7  & 60.2  \\
% \multicolumn{1}{l|}{Qwen-2.5$_{1.5B}$} & 54.9  & 9.9   & 53.0  & 9.9   & 51.1  & 9.0   & 49.5  & 9.6   & 42.3  & 9.7   & 27.9  & 10.5  & 16.8  & 10.7  \\
\multicolumn{1}{l|}{Llama-3.1$_{8B}$}  & 57.0  & 16.7  & 56.2  & 15.9  & 54.6  & 16.4  & 53.1  & 15.4  & 49.3  & 16.8  & 41.4  & 15.7  & 30.9  & 17.8  \\
\multicolumn{1}{l|}{Qwen-2.5$_{7B}$}   & 20.3  & 66.8  & 20.7  & 66.2  & 20.3  & 66.1  & 20.2  & 66.2  & 17.4  & 68.3  & 16.3  & 68.2  & 14.2  & 68.4  \\
\multicolumn{1}{l|}{R1-Distill-Llama$_{8B}$} & 56.6  & 7.8   & 55.4  & 9.2   & 55.8  & 6.8   & 55.8  & 6.4   & 50.5  & 7.2   & 35.7  & 8.8   & 21.9  & 9.7  \\
\midrule
\multicolumn{15}{c}{\textsc{SKR}~\cite{SKR}} \\ 
\midrule
% \multicolumn{1}{l|}{Llama-3.2$_{1B}$} & 11.2  & 57.8  & 12.0  & 57.6  & 12.4  & 55.5  & 11.6  & 55.1  & 11.9  & 55.8  & 11.8  & 56.6  & 11.5  & 55.4  \\
% \multicolumn{1}{l|}{Qwen-2.5$_{1.5B}$} & 31.6  & 8.5   & 30.3  & 8.9   & 29.6  & 8.4   & 30.9  & 8.6   & 28.2  & 10.2  & 20.1  & 9.7   & 15.8  & 9.1   \\
\multicolumn{1}{l|}{Llama-3.1$_{8B}$}  & 41.5  & 11.8  & 40.8  & 12.5  & 40.3  & 12.3  & 39.1  & 13.6  & 38.0  & 13.9  & 38.4  & 13.9  & 36.7  & 13.9  \\
\multicolumn{1}{l|}{Qwen-2.5$_{7B}$}   & 25.5  & 55.8  & 23.8  & 55.8  & 23.8  & 56.6  & 23.1  & 56.5  & 21.3  & 57.5  & 19.5  & 58.7  & 17.1  & 60.2  \\
\multicolumn{1}{l|}{Llama-3.1$_{70B}$}  & 62.7  & 5.1   & 61.1  & 4.8   & 60.6  & 4.4   & 58.3  & 4.5   & 57.8  & 4.8   & 55.9  & 4.8   & 49.4  & 7.3   \\
\multicolumn{1}{l|}{Qwen-2.5$_{72B}$}   & 52.6  & 22.5  & 53.5  & 22.2  & 53.2  & 21.9  & 52.6  & 22.5  & 50.8  & 24.0  & 45.3  & 27.1  & 41.2  & 27.2  \\
% \multicolumn{1}{l|}{DeepSeek-V3}    & 53.8  & 16.3  & 54.2  & 16.5  & 54.4  & 16.7  & 54.7  & 14.9  & 53.5  & 16.1  & 52.7  & 17.1  & 50.2  & 17.2  \\
% \multicolumn{1}{l|}{DeepSeek-R1}    & 62.4  & 8.1   & 63.7  & 7.5   & 62.6  & 8.2   & 63.5  & 7.1   & 61.6  & 9.1   & 60.4  & 9.4   & 57.4  & 11.6  \\ 
\multicolumn{1}{l|}{R1-Distill-Llama$_{8B}$} & 43.9  & 11.6  & 45.4  & 10.8  & 41.7  & 12.3  & 42.3  & 9.9   & 37.8  & 12.0  & 30.9  & 14.1  & 25.9  & 15.2  \\
\multicolumn{1}{l|}{R1-Distill-Llama$_{70B}$} & 60.0  & 7.9   & 61.1  & 5.9   & 61.2  & 6.2   & 58.7  & 7.3   & 57.2  & 8.1   & 55.4  & 8.9   & 51.4  & 10.6  \\
\midrule
\multicolumn{15}{c}{\textsc{CRAG}~\cite{CRAG}} \\ 
\midrule
\multicolumn{1}{l|}{Llama-3.1$_{8B}$} & 70.9  & 11.5  & 70.6  & 9.8  & 68.2  & 9.2  & 65.4  & 9.1  & 55.4  & 11.1  & 37.3  & 14.1  & 21.7  & 15.6  \\
\multicolumn{1}{l|}{Qwen-2.5$_{7B}$} & 58.7  & 26.9  & 59.1  & 25.5  & 57.5  & 23.7  & 52.9  & 25.1  & 38.3  & 33.4  & 24.1  & 41.9  & 10.0  & 46.6  \\
\multicolumn{1}{l|}{Llama-3.1$_{70B}$} & 73.4  & 9.1   & 75.7  & 8.9   & 72.9  & 7.9   & 70.2  & 7.7   & 62.6  & 9.3   & 51.1  & 11.4  & 33.9  & 15.7  \\
\multicolumn{1}{l|}{Qwen-2.5$_{72B}$} & 70.8  & 23.4  & 72.9  & 19.9  & 72.1  & 18.9  & 68.5  & 18.3  & 58.5  & 26.6  & 33.9  & 42.1  & 13.6  & 53.3  \\
% \multicolumn{1}{l|}{DeepSeek-V3} & 71.7  & 20.3  & 73.8  & 17.8  & 72.9  & 16.7  & 71.0  & 15.6  & 60.2  & 21.7  & 39.6  & 31.1  & 15.8  & 40.5  \\
% \multicolumn{1}{l|}{DeepSeek-R1} & 72.6  & 18.5  & 74.1  & 16.1  & 73.9  & 15.4  & 73.1  & 14.7  & 60.5  & 23.6  & 37.5  & 38.0  & 17.8  & 47.4  \\
\multicolumn{1}{l|}{R1-Distill-Llama$_{8B}$} & 71.3  & 12.8  & 71.5  & 11.3  & 68.0  & 11.1  & 63.1  & 9.6   & 49.4  & 13.0  & 34.0  & 17.4  & 22.0  & 17.9  \\
\multicolumn{1}{l|}{R1-Distill-Llama$_{70B}$} & 73.9  & 13.8  & 73.7  & 13.7  & 73.4  & 12.1  & 71.4  & 10.9  & 62.1  & 16.3  & 41.9  & 28.1  & 25.9  & 32.8  \\
\bottomrule
\end{tabular}
}
\end{table}
The main experiments constructed retrieval contexts by randomly sampling from the golden and noise documents.
Detailed statistics regarding the composition of retrieval contexts are provided in Appendix \ref{app_sec:F_1}.
We evaluated the performance of different generators and denoising strategies across multiple noise ratios to reveal the erosion effect of retrieval noise on the performance of RAG systems.
% Tables 1 and 2 present the main results on the $MM_s$ and $MM_m$ datasets.
Table \ref{tab:main_result} and \ref{tab:main_norag} present partial results, with the complete results provided in Appendix \ref{app_sec:F_2}.
We compare the RAG-based systems to the \textsc{NoRAG} and observe that incorporating retrieval significantly improves answer correctness; for example, the Llama-3.1$_{8B}$ achieves a relative improvement of 50.35\%.
As the noise ratio increases, answer correctness degrades across all RAG variants; however, this degradation is not linear.
We identify a critical threshold at a 50\% noise ratio, beyond which RAG performance deteriorates rapidly, resulting in a pronounced avalanche-style collapse in correctness.
% This phenomenon is consistently observed across all generators and denoising strategies.
% comparison
Moreover, we observe that all robust RAG systems, except \textsc{ChainofNote}, severely reduce answer correctness at low-to-medium noise ratios; for instance, the Llama-3.1$_{8B}$ $+$ \textsc{SKR} exhibits a 42.55\% correctness drop compared to \textsc{VanillaRAG}.
Interestingly, a surprising reversal occurs under a fully noisy context: in this scenario, all denoising strategies (excluding \textsc{ChainofNote}) outperform \textsc{VanillaRAG} by a significant margin.
Furthermore, \textsc{SKR} and \textsc{DRGIN} demonstrate high rejection rates and a more gradual correctness decline in high-noise settings, potentially because their document utility assessment mechanism, while filtering noise, may also prevent leveraging beneficial retrieved context.
\begin{wraptable}{r}{6cm}
\caption{Performance of NoRAG ($MM_s$).}
\label{tab:main_norag}
\centering
\resizebox{0.4\textwidth}{!}{
\begin{tabular}{lc>{\columncolor{gray!20}}c}
\toprule
\multicolumn{1}{l}{\multirow{2}{*}{\textsc{LLM Generator}}} & \multicolumn{2}{c}{\textsc{NoRAG}} \\
\cmidrule(lr){2-3}
\multicolumn{1}{c}{} & \multicolumn{1}{c}{Cor.} & \multicolumn{1}{c}{Rej.} \\
\midrule
% RGB\cite{RGB} & 2 & \cmark \\
\multicolumn{1}{l|}{Llama-3.1$_{8B}$} & 36.8 & 14.8 \\
\multicolumn{1}{l|}{Qwen-2.5$_{7B}$} & 16.3 & 65.6 \\
\multicolumn{1}{l|}{Llama-3.1$_{70B}$} & 54.8 & 7.2 \\
\multicolumn{1}{l|}{Qwen-2.5$_{72B}$} & 41.2 & 34.7 \\
\multicolumn{1}{l|}{R1-Distill-Llama$_{8B}$} & 27.5 & 15.2 \\
\multicolumn{1}{l|}{R1-Distill-Llama$_{70B}$} & 54.1 & 10.0 \\
\bottomrule
\end{tabular}
}
\end{wraptable}
Performance disparities across LLM families are also prominent: Llama-series models (“aggressive”) achieve higher correctness, whereas Qwen-series models (“conservative”) exhibit superior rejection rates, which is a valuable trait in high-noise conditions.
We further observe that scaling the model brings limited performance gains in low-noise settings. 
However, under high retrieval noise conditions, increasing the model scale substantially improves robustness.
% Surprisingly, on the $MM_m$ dataset, reasoning-optimized models such as R1-Distill-Llama$_{8B}$ and R1-Distill-Llama$_{70B}$ exhibit almost no advantage. 
% Intuitively, reasoning-oriented models would be expected to perform better in multi-hop reasoning tasks due to their enhanced reasoning capabilities. 
\textbf{Overall, the robustness of RAG systems leaves considerable room for enhancement.} 
Their pronounced sensitivity to noise means any interference can degrade performance.
Evaluating robustness at a fixed noise level provides only partial insights.
% Comprehensive experimental results are provided in \textcolor{red}{Appendix A}.

\subsection{Noise Type Analysis}

\begin{figure}[tbp]
    \centering
    \setlength{\abovecaptionskip}{5pt}
    \setlength{\belowcaptionskip}{-10pt}
    \begin{subfigure}[t]{\linewidth}
        \centering
        \includegraphics[width=\linewidth]{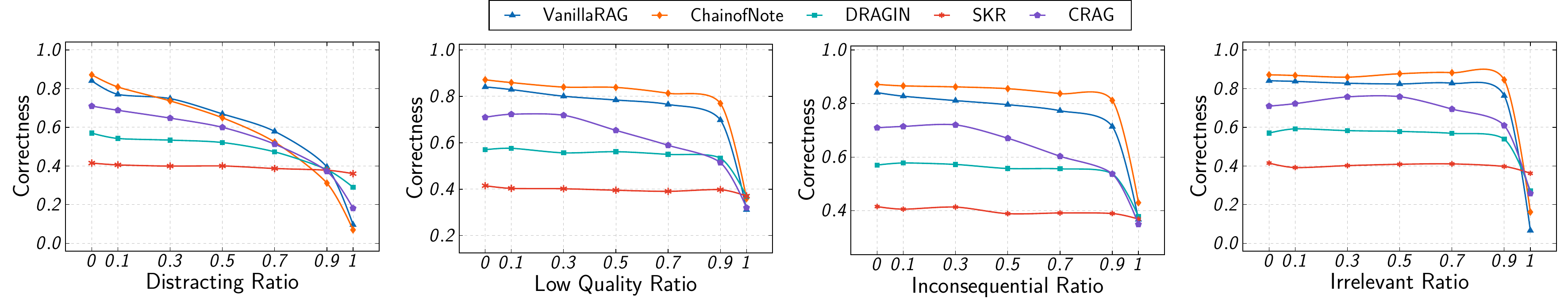}
        % \caption{Llama-3.1${_{8B}}$ as the LLM generator.}
        % \label{fig:sub1_acc_Llama}
    \end{subfigure}

    \vspace{2pt}
    
    \begin{subfigure}[t]{\linewidth}
        \centering
        \includegraphics[width=\linewidth]{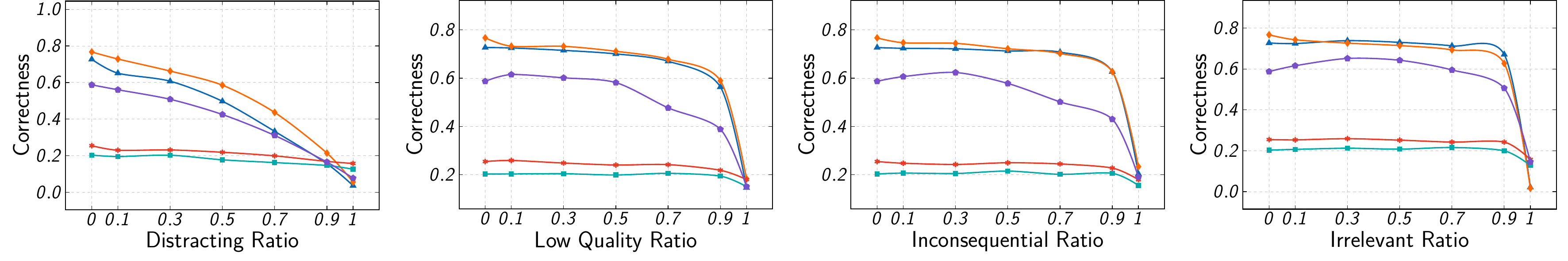}
        % \caption{Qwen-2.5${_{8B}}$ as the LLM generator.}
        % \label{fig:sub1_acc_Qwen}
    \end{subfigure}
    \caption{Impact of different types of retrieval noise on RAG system performance, using Llama-3.1${_{8B}}$ (top) and Qwen-2.5${_{7B}}$ (bottom) as LLM generators.}
    \label{fig:sub1_acc}
\end{figure}

Figure \ref{fig:sub1_acc} illustrates the impact of introducing individual types of retrieval noise on RAG system performance.
It is evident that distracting noise significantly degrades performance, even at a noise ratio as low as 10\%, dropping correctness to 40.56\% for Llama-3.1$_{8B}$ $+$ \textsc{SKR} and 23.02\% for Qwen-2.5$_{8B}$ $+$ \textsc{SKR}.
Conversely, irrelevant noise exhibits the mildest detrimental effect.
This indicates that distracting noise is particularly misleading, causing the RAG system to become easily misguided when presented with such documents.
Notably, we also observe that individual types of noise exhibit somewhat reduced harmfulness compared to mixed noise conditions. Excluding distracting noise, the noise proportion threshold at which severe model performance degradation occurs increases up to approximately 90\%.
Additionally, we observe that, irrespective of whether Llama-3.1$_{8B}$ or Qwen-2.5$_{8B}$, performance curves corresponding to different denoising strategies show a noticeable intersection point at a distracting noise proportion of 90\%, after which DRGIN and SKR surpass others.
However, under low-noise conditions, the performance of DRGIN and SKR remains unsatisfactory, possibly due to their tendency to excessively question retrieved content in the presence of even minor noise.
\textbf{In summary, different types of noise affect RAG systems in distinctly varying ways.} 
Therefore, tailored denoising strategies should be developed according to the unique characteristics of each noise type.
% Furthermore, despite stark differences in linguistic properties and underlying noise characteristics, the effects of low-quality and inconsequential noise on the RAG’s performance are almost identical.

% 2-sub1_acc

\subsection{Sensitivity Analysis}

\begin{figure}[htbp]
    \centering
    \begin{subfigure}[b]{\linewidth}
        \centering
        \includegraphics[width=\linewidth]{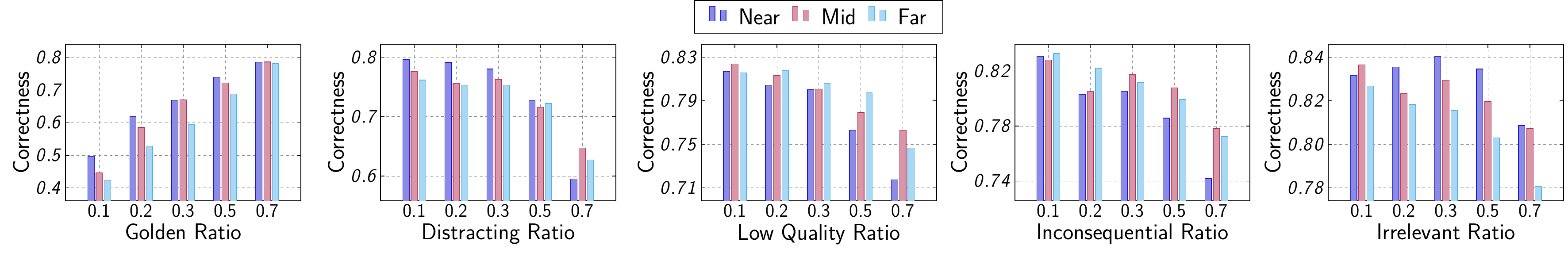}
        \label{fig:sub2_acc_Llama}
    \end{subfigure}
    
    \vspace{-1em} 
    
    \begin{subfigure}[b]{\linewidth}
        \centering
        \includegraphics[width=\linewidth]{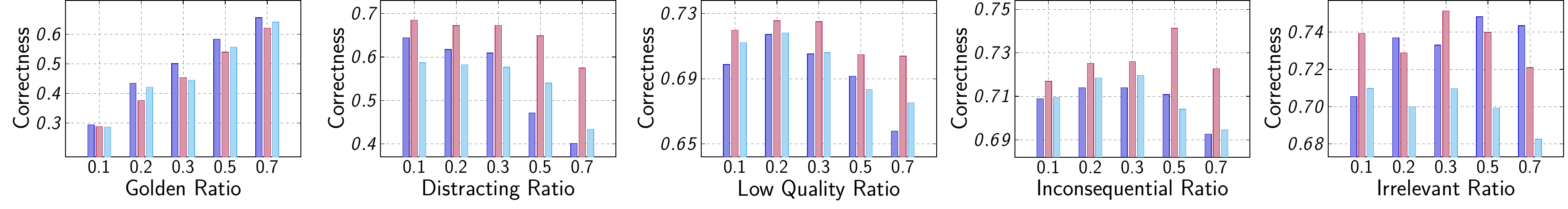}
        \label{fig:sub2_acc_Qwen}
    \end{subfigure}
    \caption{Sensitivity of RAG system performance to the position of retrieved documents, using Llama-3.1${{8B}}$ (top) and Qwen-2.5${{7B}}$ (bottom) as LLM generators.}
    \label{fig:sub2_acc}
\end{figure}

Figure \ref{fig:sub2_acc} illustrates the RAG system's sensitivity to the retrieved documents' position. 
We categorize document positions into three groups: “Near” indicating proximity to the query; “Mid” referring to documents positioned centrally among retrieved documents; and “Far” denoting documents located at the end of the input sequence. 
We observe that Llama-3.1$_{8B}$ is entirely insensitive to document ordering. In contrast, Qwen-2.5$_{7B}$ exhibits a clear “lost in the middle” phenomenon, demonstrating significant sensitivity to mid-positioned documents. 
Under high-noise conditions, positioning the golden document closer to the query enhances the generator’s attention towards relevant content, thus increasing answer correctness. 
\textbf{This finding suggests that reranking retrieved documents can beneficially improve RAG performance, although reranking may concurrently amplify the detrimental effects of distracting and inconsequential noise}.

% 3-sub2_acc

\begin{figure}[htbp]
    \centering
    \setlength{\abovecaptionskip}{-5pt}
    \begin{subfigure}[b]{0.4473\linewidth}
        \centering
        \includegraphics[width=\linewidth]{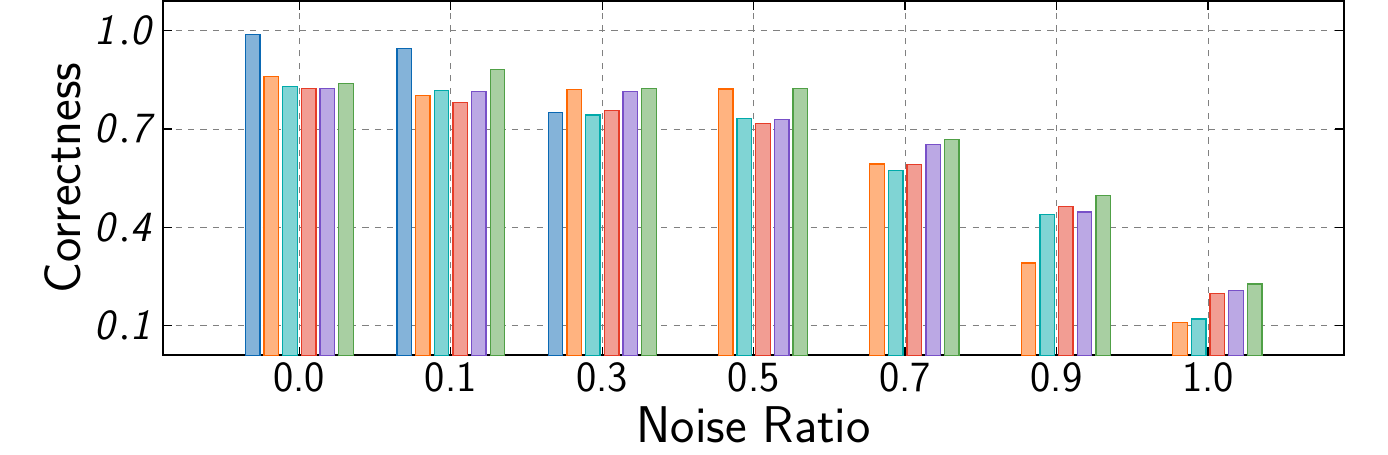}
        \label{fig:sub3_acc_Llama}
    \end{subfigure}
    \begin{subfigure}[b]{0.542619\linewidth}
        \centering
        \includegraphics[width=\linewidth]{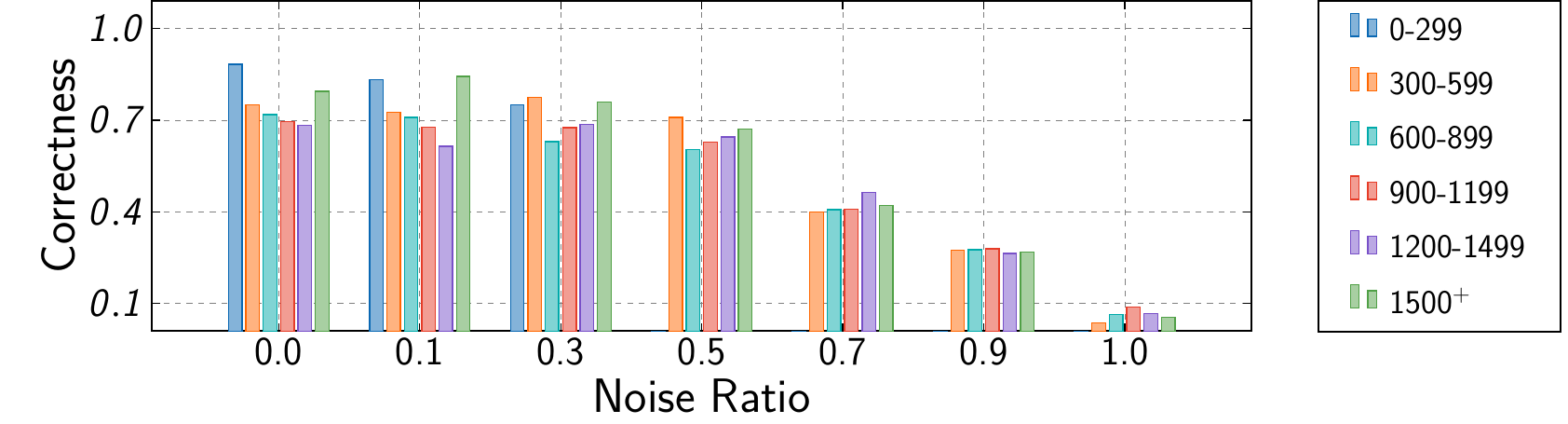}
        \label{fig:sub3_acc_Qwen}
    \end{subfigure}
    \caption{Sensitivity of RAG system performance to the length of retrieved documents, using Llama-3.1${_{8B}}$ (left) and Qwen-2.5${_{7B}}$ (right) as LLM generators.}
    \label{fig:sub3_acc}
\end{figure}

Figure \ref{fig:sub3_acc} presents the sensitivity of the RAG system to the length of retrieved content. Under low-noise conditions, shorter retrieval content benefits RAG, as the generator can more easily and accurately extract answers. 
Conversely, under high-noise environments, longer retrieval contexts improve the generator’s ability to produce correct responses, possibly due to the generator relying on additional contextual information to determine the correct answer amidst noisy context. \textbf{This observation indicates that retrieval length should be minimized when retrieved content is relatively clean, but extended when substantial noise is present, to provide sufficient contextual support for the generator.}

% 4-sub3_acc

\begin{wraptable}{r}{6.5cm}
\caption{Performance of generator-denoiser combinations across different noise scenarios.}
\label{tab:sub4_different_scenarios}
\centering
\resizebox{0.45\textwidth}{!}{
\begin{tabular}{l
    c>{\columncolor{gray!20}}c
    c>{\columncolor{gray!20}}c
    c>{\columncolor{gray!20}}c
    c>{\columncolor{gray!20}}c}
\toprule
\multicolumn{1}{l}{\multirow{2}{*}{\textsc{LLM Generator}}} & \multicolumn{2}{c}{\textsc{Sce.1}} & \multicolumn{2}{c}{\textsc{Sce.2}} & \multicolumn{2}{c}{\textsc{Sce.3}} & \multicolumn{2}{c}{\textsc{Sce.4}} \\ 
\cmidrule(lr){2-3} \cmidrule(lr){4-5} \cmidrule(lr){6-7} \cmidrule(lr){8-9} 
\multicolumn{1}{c}{} & 
\multicolumn{1}{c}{Cor.} & \multicolumn{1}{c}{Rej.} & 
\multicolumn{1}{l}{Cor.} & \multicolumn{1}{r}{Rej.} & 
\multicolumn{1}{c}{Cor.} & \multicolumn{1}{c}{Rej.} & 
\multicolumn{1}{c}{Cor.} & \multicolumn{1}{c}{Rej.} \\ 
\midrule
\multicolumn{9}{c}{\textsc{VanillaRAG}} \\ 
\midrule
\multicolumn{1}{l|}{Llama-3.1$_{8B}$}        & 46.8  & 3.3   & 75.2  & 2.5   & 69.9  & 3.6   & 47.5  & 4.1 \\ 
\multicolumn{1}{l|}{Qwen-2.5$_{7B}$}         & 36.6 & 28.8 & 63.4 & 16.6 & 55.4 & 18.4 & 33.6 & 32.8 \\
\midrule
\multicolumn{9}{c}{\textsc{ChainofNote}} \\ 
\midrule
\multicolumn{1}{l|}{Llama-3.1$_{8B}$}        & 47.5 & 1.3 & \textbf{75.6} & 2.1 & \textbf{74.2} & 1.4 & 35.9 & 1.9 \\ 
\multicolumn{1}{l|}{Qwen-2.5$_{7B}$}         & \textbf{43.8} & 11.3 & \textbf{67.3} & 8.1 & \textbf{63.3} & 7.3 & 41.4 & 17.4 \\
\midrule
\multicolumn{9}{c}{\textsc{DRAGIN}} \\ 
\midrule
\multicolumn{1}{l|}{Llama-3.1$_{8B}$}        & 38.9 & 17.0 & 53.7 & 15.1 & 52.3 & 15.6 & \textbf{48.0} & 16.6 \\
\multicolumn{1}{l|}{Qwen-2.5$_{7B}$}         & 18.4 & 52.8 & 19.7 & 66.5 & 19.1 & 66.2 & 17.3 & 62.9 \\
\midrule
\multicolumn{9}{c}{\textsc{SKR}} \\ 
\midrule
\multicolumn{1}{l|}{Llama-3.1$_{8B}$}        & 39.1 & 12.5 & 40.1 & 12.2 & 38.6 & 13.1 & 38.7 & 13.1 \\
\multicolumn{1}{l|}{Qwen-2.5$_{7B}$}         & 21.9 & 47.1 & 24.1 & 55.4 & 22.4 & 56.6 & 20.5 & 58.7 \\
\midrule
\multicolumn{9}{c}{\textsc{CRAG}} \\ 
\midrule
\multicolumn{1}{l|}{Llama-3.1$_{8B}$}        & \textbf{54.8} & 8.9 & 65.1 & 8.7 & 57.2 & 11.1 & 46.5 & 8.6 \\
\multicolumn{1}{l|}{Qwen-2.5$_{7B}$}         & 40.3 & 30.6 & 54.6 & 23.3 & 44.4 & 30.3 & \textbf{42.6} & 24.5 \\
\bottomrule
\end{tabular}
}
\end{wraptable}
\subsection{Scenario-Level Robustness Analysis}
Based on a coarse-grained statistical analysis of noise distributions across four discussion topics, we synthesized four distinct scenario-level~(\textsc{Sce.1}–\textsc{Sce.4}) noise distributions for robustness evaluation.
The specific noise distributions for each scenario and their design rationale are detailed in Appendix \ref{app_sec:G_2}.
Table \ref{tab:sub4_different_scenarios} summarizes the performance results for various generator-denoiser combinations across these constructed scenarios.
Our findings indicate that noise composition influences RAG performance in these simulated environments.
\textbf{Both the choice of generator and the applied denoising strategy exhibit high sensitivity to the specific types and proportions of noise.} 
For instance, despite \textsc{Sce.1} and \textsc{Sce.2} sharing an identical proportion of Golden documents (30\%), the optimal generator and denoising strategy combination differed substantially. Similarly, between \textsc{Sce.3} and \textsc{Sce.4}, the performance of a fixed generator and denoising configuration varied significantly, with a maximum observed performance delta of 21.4\%.
This lack of a predictable relationship presents a considerable challenge for pre-determining the optimal generator-denoiser configurations.
% This lack of a predictable relationship presents a considerable challenge for pre-determining the optimal generator-denoiser configurations under varying noise conditions.

% Scenario 1 (\textsc{Sce.1}) was configured with the following document proportions: Golden Document (G): 30\%, Distracting Noise (D): 30\%, Inconsequential Noise (In): 10\%, Low-Quality Noise (L): 10\%, and Irrelevant Noise (Ir): 20\%. Scenario 2 (\textsc{Sce.2}) featured G:30\%, D:10\%, In:10\%, L:20\%, Ir:30\%. Scenario 3 (\textsc{Sce.3}) comprised G:20\%, D:20\%, In:20\%, L:10\%, Ir:30\%. Finally, Scenario 4 (\textsc{Sce.4}) was defined by G:20\%, D:40\%, In:20\%, L:10\%, Ir:10\%. The detailed design rationale for these noise distributions is available in Appendix G.3. 

\subsection{Case Study}

% 5-case study heat plot
\begin{figure}[htbp]
    \centering
    \includegraphics[width=\linewidth]{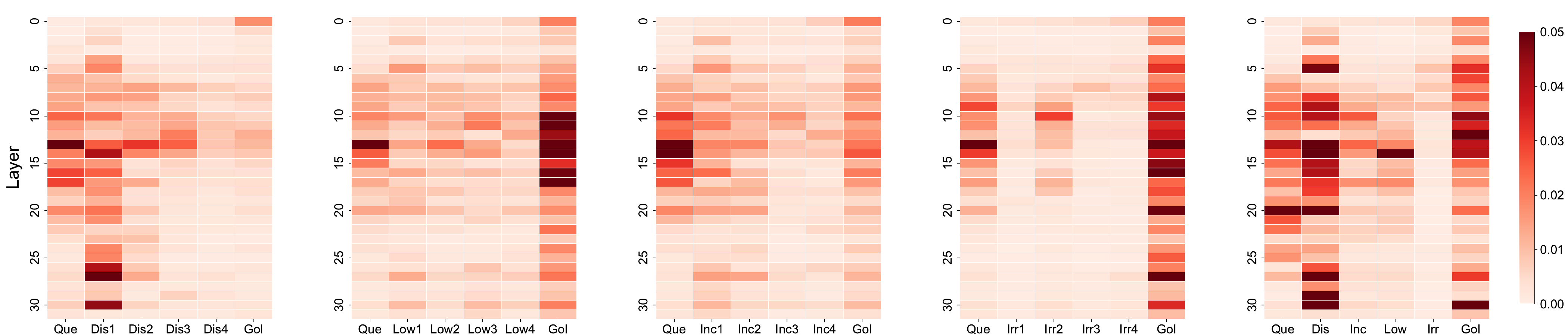}
    \caption{Layer-wise attention distribution of Llama-3.1${_{8B}}$.}
    \label{fig:heat_plot}
\end{figure}

% Example of RAG being misled by search noise and giving incorrect answers

Figure \ref{fig:heat_plot} illustrates the layer-wise attention distribution of the Llama-3.1$_{8B}$ over the question and various retrieved documents when answering the query ``\texttt{In greek mythology who was the goddess of spring growth?}''.
It can be observed that ``Distracting Noise'' poses a significant challenge: once the generator's attention is captured by distracting noise, this effect intensifies throughout subsequent layers, eventually misleading the generator into producing incorrect answers.
Similarly, ``Inconsequential Noise'' also succeeds in misguiding the model, although the attention mechanism exhibits an ongoing competition, oscillating between inconsequential noise and the golden document.
In contrast, ``Low Quality Noise'' initially attracts some attention in the earlier layers but is neglected entirely after intermediate layers. Lastly, the attention mechanism consistently overlooks ``Irrelevant Noise'' throughout the decoding process, empirically validating that singularly irrelevant distractions cannot influence the generator's output. 
\textbf{In essence, these observations demonstrate that noise can induce shifts in the generator’s attention; the more deceptive or distracting the noise, the more significant these attention shifts become, thereby increasing the likelihood of the generator being misled into producing incorrect outputs.}
Refer to Appendix \ref{app_sec:case_study} for a more detailed case analysis.

\section{Conclusion}
We introduce Magic Mushroom, a benchmark designed to evaluate the robustness of RAG systems under complex retrieval noise.
It allows flexible noise types and proportions configurations, simulating real-world retrieval noise environments. 
It comprises 7,468 single-hop and 3,925 multi-hop question-answer pairs, enabling comprehensive performance evaluations across different LLM generators and denoising strategies.
Magic Mushroom defines four categories of retrieval noise—Distracting, Low Quality, Inconsequential, and Irrelevant—reflecting real-world noise heterogeneity. 
Our experiments reveal that RAG systems are susceptible to noise, with significant performance degradation observed as noise ratios increase. 
Magic Mushroom provides a challenging and flexible evaluation framework, advancing the development of more robust and noise-resistant RAG systems for real-world applications.
% limitations
\textbf{Magic Mushroom is currently limited to English-language evaluation}. While the methodology and findings may generalize cross-lingually, extending and validating their applicability to non-English contexts remains a direction for future work.

\section{Related Work}
% In recent years, extensive research has investigated the noise robustness of LLMs in RAG systems, especially focusing on the sensitivity of various models to retrieval noise \cite{Self-RAG, rag-survey, power_noise, WikiContradict}.
Recent research extensively investigates LLM noise robustness in RAG systems, focusing on sensitivity to retrieval noise \cite{Self-RAG, rag-survey, power_noise, WikiContradict}.
% These studies indicate substantial variability in how LLMs respond to irrelevant context: some generators continue to benefit from additional context, whereas others experience performance degradation when exposed to unrelated information \cite{meta,ragged}.
These studies reveal substantial variability in LLM responses to irrelevant context: some benefit from additional context, while others degrade \cite{meta,ragged}.
\citet{raat} quantified the performance degradation of various LLMs when confronted with retrieval noise, revealing that model scale and the extent of training and fine-tuning significantly impact noise robustness.
% \citet{RGB} investigated the impact of “noisy document”—documents that appear relevant but do not contain the correct answer.
% Their findings indicated that this type of noise presents a challenge for generative models irrespective of their architecture, with even competent models being prone to generating erroneous answers when retrieval results contain distracting information.
% \citet{nomiracl} introduced the NoMIRACL benchmark for evaluating the robustness of multilingual RAG systems, finding that most LLMs struggle to strike a balance between avoiding hallucinations and providing correct answers.
% comparison
\begin{wraptable}{r}{5.6cm}
\caption{Comparison of Retrieval Noise Benchmarks.}
\label{tab:Comparision of different noise benchmarks}
\centering
\resizebox{0.4\textwidth}{!}{

\begin{tabular}{lcc}
\toprule
\textbf{Benchmark} & \textbf{Categories} & \textbf{Flexibility} \\
\midrule
RGB\cite{RGB} & 2 & \xmark \\
RECALL\cite{RECALL} & 1 & \xmark \\
RAG-Bench\cite{raat} & 3 & \xmark \\
NoMIRACL\cite{nomiracl} & 1 & \xmark \\
NoiserBench\cite{pandora} & 7 & \xmark \\
Robust RALM\cite{yoran} & 1 & \xmark \\
\rowcolor{gray!20} Magic Mushroom & 4 & \cmark \\
\bottomrule
\end{tabular}

}
\end{wraptable}
Several studies have investigated the robustness of multilingual RAG systems, finding that most LLMs struggle to strike a balance between avoiding hallucinations and providing correct answers \cite{RGB,nomiracl}.
\citet{pandora} further defined seven distinct noise types and categorized them as beneficial or harmful based on their characteristics.
% Domain‑specific and long‑document noise have also received attention. 
% RAG‑QA Arena evaluates cross‑domain robustness in long‑text question answering under retrieval augmentation \cite{Arena}.
Domain-specific and long-document noise have also been explored, such as RAG-QA Arena evaluating cross-domain robustness in long-text question answering \cite{Arena}.
Diverging from existing research, Magic Mushroom transcends the limitations of existing benchmarks (\eg singular noise types, fixed ratios) by offering flexible noise combination configurations, thus providing an evaluation environment more reflective of real-world applications.
Table \ref{tab:Comparision of different noise benchmarks} illustrates the comparison of retrieval noise benchmarks, where Flexibility refers to the benchmark's ability to construct different proportions and types of noise.

To mitigate the impact of noise, researchers have proposed various denoising strategies, encompassing post-retrieval filtering, reranking, noise-robust training, and explicit denoising during generation \cite{CRAG, DRAGIN, SKR, Chain-of-Note, FICLO, INFO-RAG}.
The simplest approach filters documents post-retrieval to eliminate irrelevant or harmful content \cite{FICLO}.
For instance, \citet{yoran} utilized natural language inference models to assess passage-QA consistency, filtering non-supporting passages.
Another paradigm is reranking: places highest-scoring contexts near the prompt to reduce interference from irrelevant passages \cite{rag-survey}.
Noise-robust training exposes models to noisy retrieval contexts during training/fine-tuning, enabling them to learn noise ignorance and correct answer extraction \cite{raat, tu, RAG-DDR}.
Alternatively, models can actively identify and exclude noise during inference. 
InstructRAG employs an instruction-tuned LM to read documents, produce a rationale, articulate reasoning, and then derive the answer \cite{instruct_rag}.
% InstructRAG first employs an instruction-tuned language model to read the retrieved documents and produce a “rationale”, subsequently articulate its reasoning process, and finally derive the answer \cite{instruct_rag}.
Other studies utilize Chain-of-Thought (CoT) to guide the model in incrementally analyzing whether each retrieved result helps answer before selectively utilizing them \cite{cot1, Chain-of-Note}.
Notably, these denoising strategies are not mutually exclusive.
Practical RAG systems often combine multiple strategies to leverage their respective strengths against diverse noise \cite{CRAG, Self-RAG}.

%%%%%%%%%%%%%%%%%%%%%%%%%%%%%%%%%%%%%%%%%%%%%%%%%%%%%%%%%%%%

\bibliographystyle{plainnat}
\bibliography{neurips_2025}

%%%%%%%%%%%%%%%%%%%%%%%%%%%%%%%%%%%%%%%%%%%%%%%%%%%%%%%%%%%%

%%%%%%%%%%%%%%%%%%%%%%%%%%%%%%%%%%%%%%%%%%%%%%%%%%%%%%%%%%%%

\newpage

\appendix

\section{Domain-wise Noise Distribution Analysis} \label{app_sec:A}
\begin{figure}[htbp]
    \centering
    \begin{subfigure}[b]{\linewidth}
        \centering
        \includegraphics[width=\linewidth]{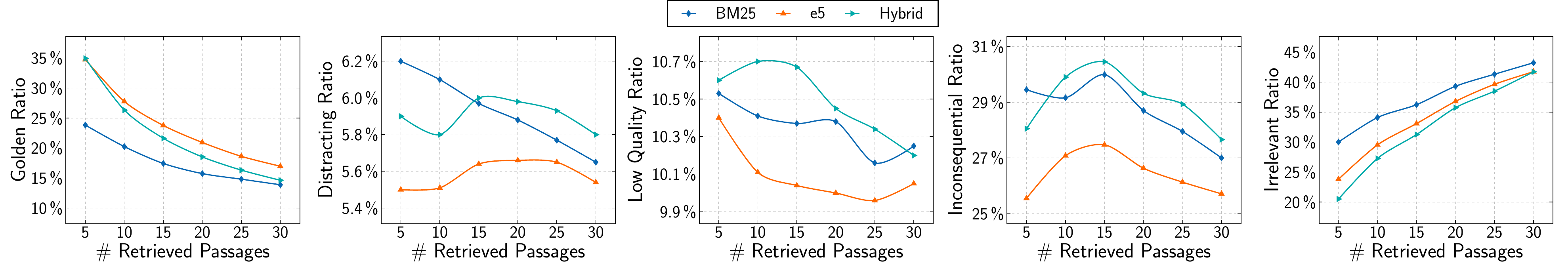}
        % \caption{Caption}
        % \label{fig:retrieval_distribution_fulldata}
    \end{subfigure}
    
    \begin{subfigure}[b]{\linewidth}
        \centering
        \includegraphics[width=\linewidth]{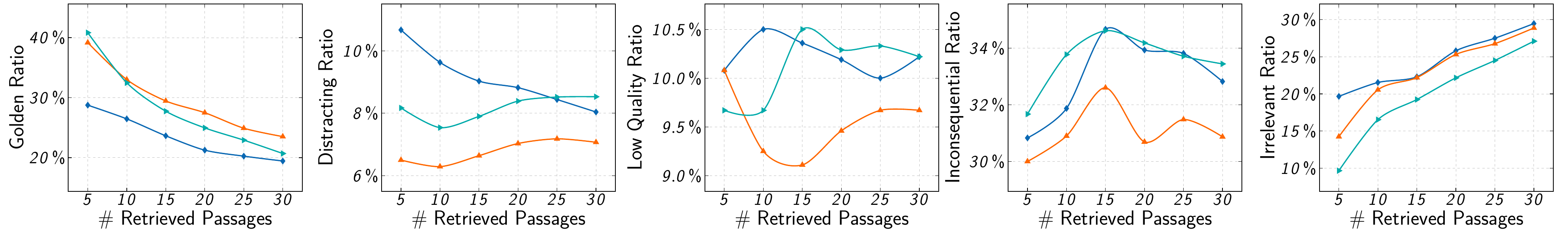}
        % \caption{Caption}
        % \label{fig:retrieval_distribution_History_domain}
    \end{subfigure}
    
    \begin{subfigure}[b]{\linewidth}
        \centering
        \includegraphics[width=\linewidth]{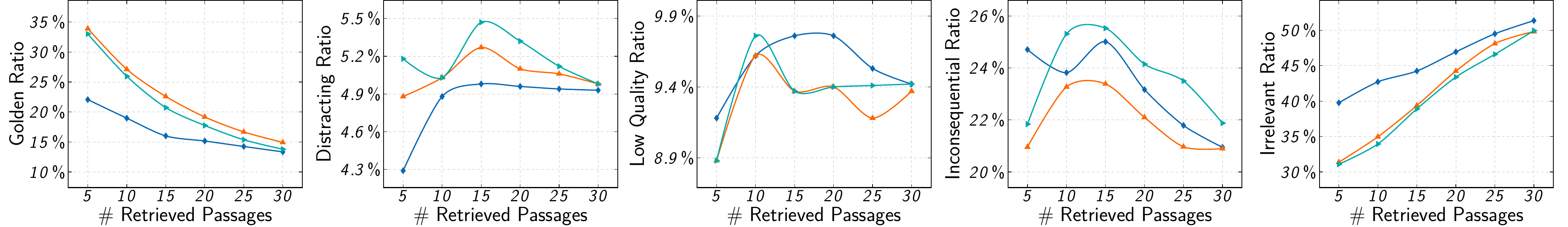}
        % \caption{Caption}
        % \label{fig:retrieval_distribution_Literature&Arts_domain}
    \end{subfigure}
    \caption{Domain-wise distribution of retrieved document types as the number of retrieved passages increases, across three retrieval methods (BM25, e5, Hybrid). The top row shows aggregate statistics across all topics; the middle and bottom rows show the distributions for the \texttt{History} and \texttt{Literature \& Arts} domains, respectively.}
    \label{fig:retrieval_distribution-different-domain}
\end{figure}

Figure~\ref{fig:retrieval_distribution-different-domain} presents a detailed analysis of the distribution of retrieved document types across different retrievers.
We observe that both the proportions and the trends of different noise types vary substantially with the number of retrieved passages and across domains. 
For instance, the fraction of golden documents consistently decreases as more passages are retrieved, while the ratios of various noise types (\eg Irrelevant, Inconsequential) increase, but the magnitude and pattern of these changes are highly domain- and retriever-dependent. 
More notably, no universal trend governs the evolution of each noise category—distinct domains exhibit markedly different noise profiles, and even within a domain, different retrievers lead to divergent noise dynamics. 
This pronounced heterogeneity underscores the limitation of static robustness benchmarks, which typically assume a fixed or uniform noise distribution and thus only reflect specific scenarios. 
Consequently, such benchmarks fail to capture the complexity and variability inherent in real-world retrieval environments, potentially leading to misleading assessments of model robustness.
% 8-Appendix-retrieval_distribution-different-domain

\section{Prompt Design} \label{sec:prompt_design}

\subsection{Benchmark Construction Prompts}  \label{app_sec:B_1}
\begin{figure}[t]
    \centering
    \includegraphics[width=\linewidth]{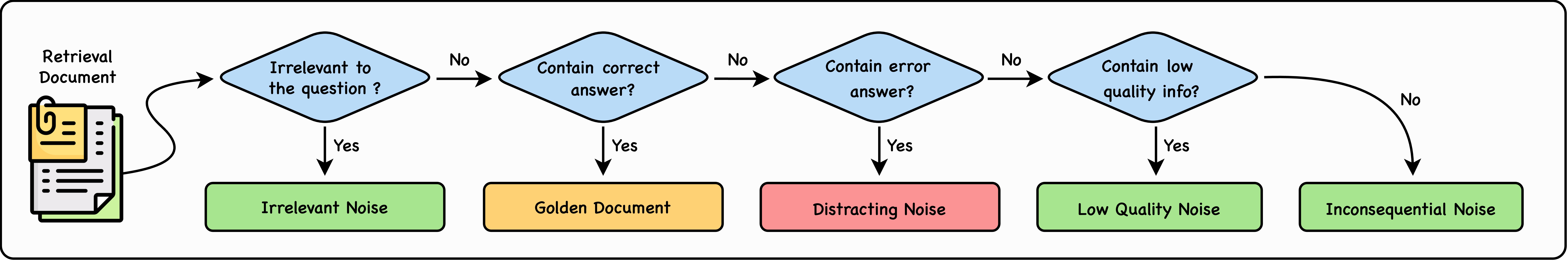}
    \caption{Schematic diagram illustrating the rationale for retrieval noise taxonomy.}
    \label{fig:benchmark_construction_floechart_sub1}
\end{figure}

This appendix provides a comprehensive overview of all prompts used in Section \ref{sec:3} (Benchmark Construction) and Section \ref{sec:4} (Experiments).
% Golden
Figure~\ref{fig:Single-Hop Dataset: Golden Documents Augmentation} and Figure~\ref{fig:Multi-Hop Dataset: Golden Documents Augmentation} present the prompts used to augment golden documents in single-hop and multi-hop settings, respectively.
These prompts aim to produce alternative phrasings of ground-truth evidence passages that remain semantically faithful to the original, enriching the gold context without introducing factual variance.
% Distracting
Figure~\ref{fig:Single-Hop Dataset: Distracting Noise Introduction} and Figure~\ref{fig:Multi-Hop Dataset: Distracting Noise Introduction} show the prompts designed to generate distracting noise documents in the single-hop setting and multi-hop setting, respectively. 
These documents intentionally relate to the question topic but lead the model to incorrect conclusions. They simulate realistic, misleading content retrieved by imperfect retrieval systems.
% Low Quality
Figure~\ref{fig:Single-Hop Dataset: Low Quality Noise Introduction} and Figure~\ref{fig:Multi-Hop Dataset: Low Quality Noise Introduction} illustrate the prompts used to create low quality noise with counterfactual information. These prompts replace the correct entities with semantically plausible yet factually incorrect information.
The rationale for retrieval noise taxonomy is illustrated in Figure~\ref{fig:benchmark_construction_floechart_sub1}.

\input{Prompt/dataset-generation}

\subsection{Experimental Prompts}
Figure~\ref{fig:NoRAG Inference Prompt} presents the inference prompt used in the experiment without employing RAG, where no external documents are incorporated.
Figure~\ref{fig:VanillaRAG Inference Prompt} illustrates the prompt utilized in the \textsc{VanillaRAG} framework, which integrates retrieved documents directly into the prompt.
Figure~\ref{fig:ChainofNote Inference Prompt} shows the prompt used within the \textsc{ChainofNote} framework, which guides the generator to analyze the retrieved information incrementally before proceeding to answer generation.
Figure~\ref{fig:SKR Prompt} displays the prompt applied in the \textsc{SKR} framework, which is designed to assess whether external knowledge assistance is required.

\input{Prompt/RAG-inference}

\clearpage

\section{Detailed Benchmark Statistics} \label{app_sec:C}
\begin{figure}[htbp]
    \centering
    \includegraphics[width=\linewidth]{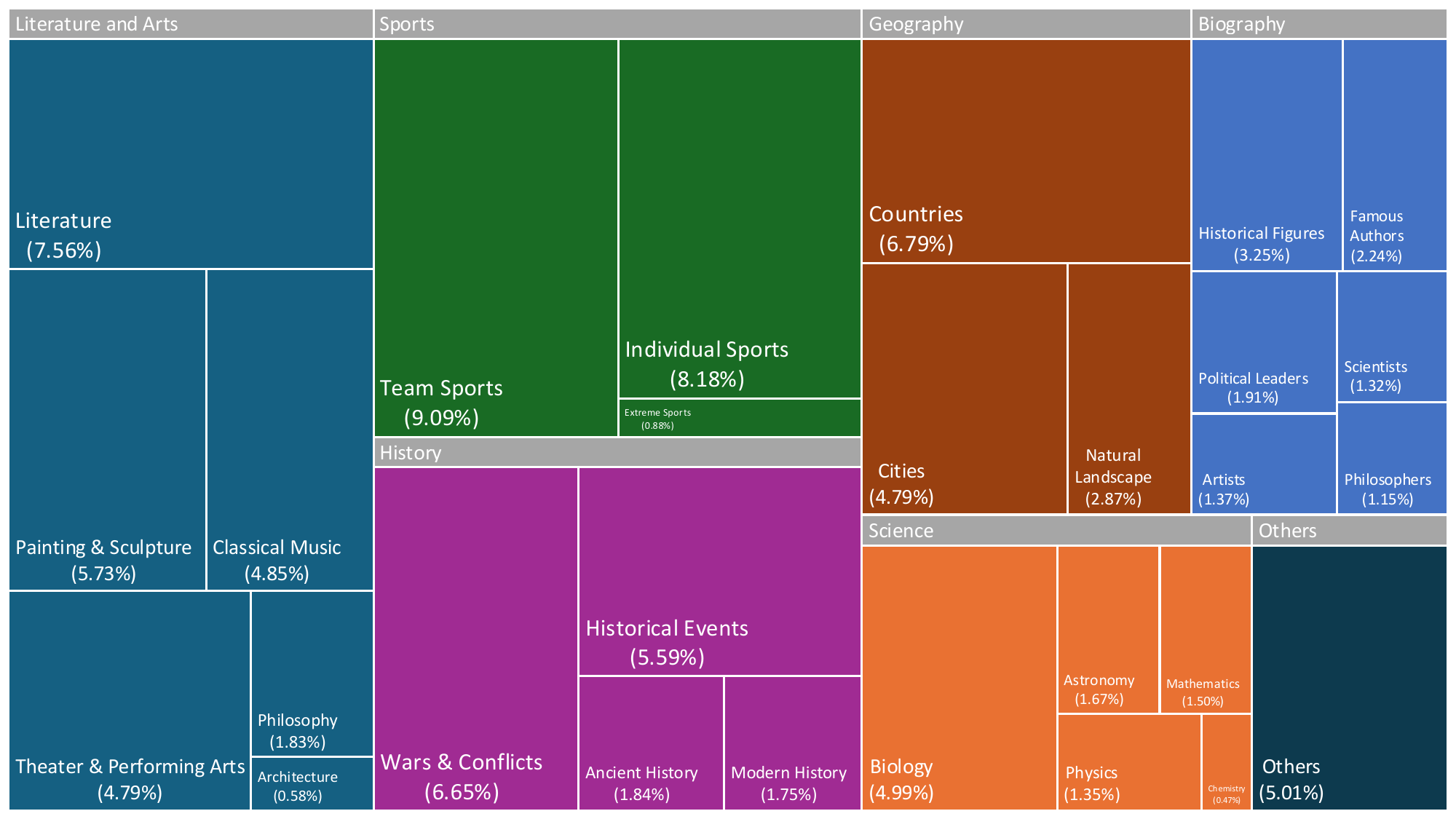}
    \caption{Topical distribution of QA instances in Magic Mushroom, with block area and percentage labels indicating category proportions.}
    \label{fig:Category}
\end{figure}

\begin{figure}[htbp]
    \centering
    \begin{subfigure}[b]{0.495\linewidth}
        \centering
        \includegraphics[width=\linewidth]{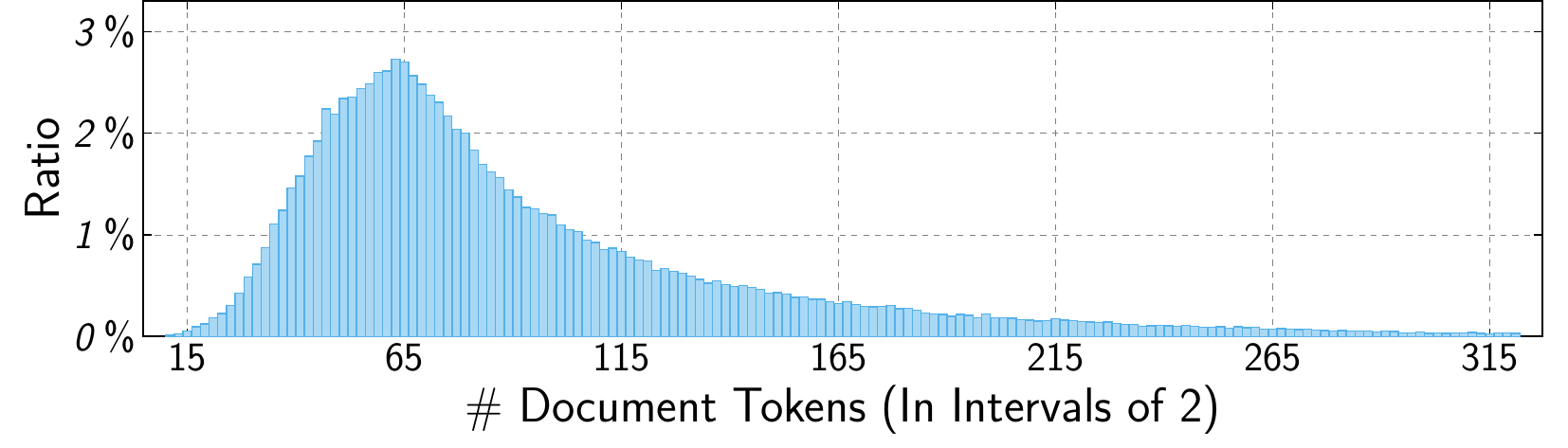}
        % \caption{Caption}
        \label{fig:token_distribution_NQ}
    \end{subfigure}
    \begin{subfigure}[b]{0.495\linewidth}
    \centering
        \includegraphics[width=\linewidth]{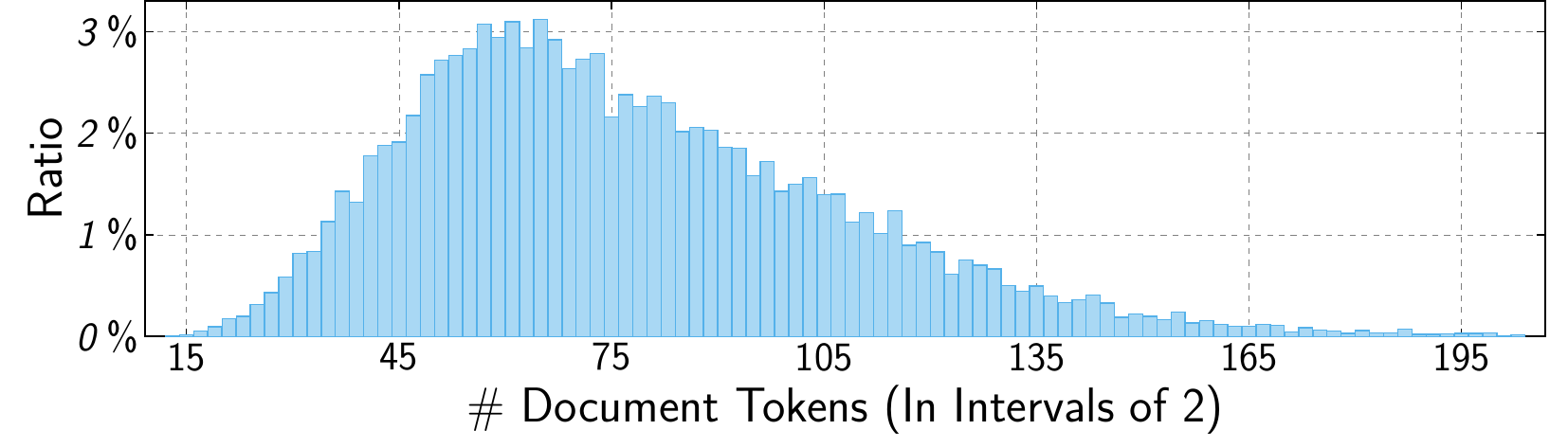}
        % \caption{Caption}
        \label{fig:token_distribution_HotpotQA}
    \end{subfigure}
    \caption{Retrieval document length distributions for $MM_s$ (left) and $MM_m$ (right).}
    \label{fig:token_distribution}
\end{figure}

This appendix presents detailed statistics of the Magic Mushroom benchmark. 
Magic Mushroom comprises 7,468 single-hop and 3,925 multi-hop complex QA pairs. 
The topical distribution of all QA instances is illustrated in Figure~\ref{fig:Category}, spanning 7 major categories and 28 subcategories. 
Each QA pair is assigned a set of candidate retrieval documents, including 10 Golden Documents, 10 Distracting noise documents, and 7 exemplars each for Low Quality, Inconsequential, and Irrelevant noise types. 
During testing, retrieval documents of specific noise types and proportions can be randomly sampled from the candidate set to accommodate diverse research needs.
Furthermore, Magic Mushroom is divided into $MM_s$ (single-hop) and $MM_m$ (multi-hop) subsets. 
The distribution of retrieval document lengths for each subset is shown in Figure~\ref{fig:token_distribution}. 
In $MM_s$, retrieval document lengths are mainly distributed between 38 and 104 tokens, while in $MM_m$, they range from 36 to 116 tokens, showing minimal difference between the two. 
Researchers may also combine $MM_s$ and $MM_m$ for more comprehensive studies.
% 7-Appendix-Category_distribution

% 6-Appendix-docs_token_distribution

\FloatBarrier
\section{Details of Evaluation Metrics} \label{app_sec:D}
\subsection{Prompts for Evaluation Metrics}

To better evaluate the correctness (\textbf{Cor.}) in RAG systems, we adopt an LLM-based scoring prompt, illustrated in Figure~\ref{fig:Prompt for the Evaluation of Correctness}, instead of traditional lexical metrics such as EM and F1\cite{EM-F1}. This prompt instructs the model to assess the factual accuracy and completeness of generated answers against gold answers, addressing the evaluation limitations of surface-level matching in QA tasks.

\input{Prompt/evaluation-metrics}

\FloatBarrier

\subsection{Human-AI Alignment}

\begin{wrapfigure}{r}{5.6cm}
\centering
\resizebox{0.35\textwidth}{!}{
\includegraphics[width=\linewidth]{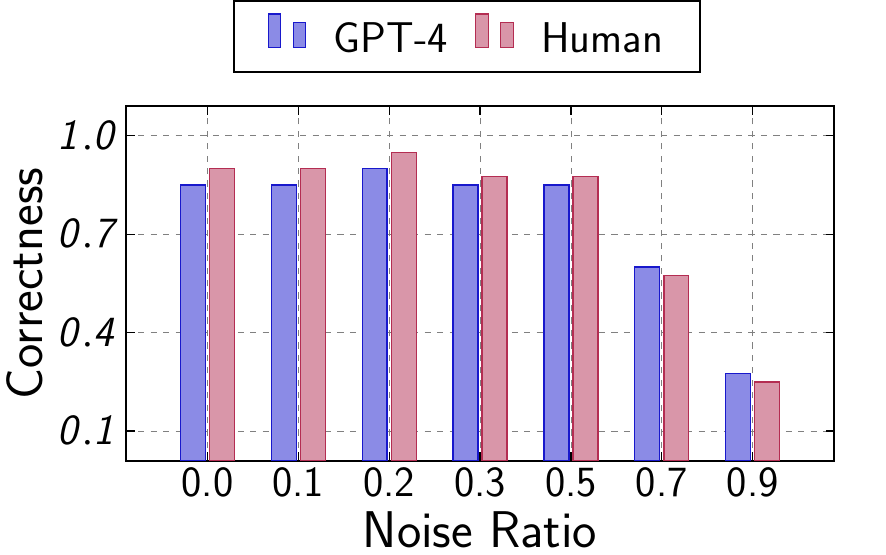}

}
\caption{Comparison of alignment scores between human and GPT-4.}
\label{fig:Human-AI Alignment}
\end{wrapfigure}
To validate the alignment between LLM-based automatic evaluation and human judgment, we conduct an alignment analysis on the performance of \textsc{VanillaRAG} + LLama-3.1$_{8B}$ at different noise ratios. 
We randomly sample 200 instances for each noise ratio and compute the average correctness scores assigned independently by human evaluators and the GPT-4. 
As shown in Figure~\ref{fig:Human-AI Alignment}, the scores exhibit high similarity, indicating strong alignment between LLM-based and manual assessments. 
These findings support the effectiveness of the LLM-based evaluation framework as a reliable surrogate for human judgment in large-scale experimental settings.

\section{Baseline Details} \label{app_sec:E}
% \subsection{Baseline Method Descriptions}
The choice of baseline generators and denoising strategies is motivated by the need to evaluate RAG robustness under diverse noisy retrieval scenarios comprehensively. 
For LLM generators, we include a wide range of models differing in architecture and scale, such as Llama-3, Qwen-2.5, and DeepSeek series, as well as distilled variants (\eg R1-Distill-Llama).
For denoising strategies, our selection covers several representative paradigms in the literature.
\textsc{VanillaRAG} serves as the basic RAG architecture without explicit denoising, acting as a baseline for robustness assessment. 
\textsc{ChainofNote} exemplifies chain-of-thought prompting, guiding the model to produce intermediate reasoning steps for enhanced semantic stability. 
DRAGIN and SKR adopt adaptive retrieval based on generation confidence, dynamically controlling retrieval to mitigate noise injection. 
CRAG employs pre-generation verification and refinement of retrieved content and optional web augmentation to block noise propagation and improve system robustness.
To ensure fairness, we do not include fine-tuning-based denoising methods in the main evaluation, as such strategies require access to training data. 
Nevertheless, Magic Mushroom provides a dedicated development set to facilitate future research on fine-tuned denoising approaches.
Appendix~\ref{sec:prompt_design} provides all the prompts used in this study.
All configuration settings for generators and denoising strategies follow the optimal settings reported in the original studies. 
Notably, for CRAG, the thresholds for triggering the three actions (correct, ambiguous, and incorrect) are set to $(0.5,-0.91)$.
All reported experimental results are obtained by averaging over 3 independent runs to ensure statistical reliability.

% \subsection{Parameter Settings and Prompts}

\section{Supplementary Main Experimental Results}
\subsection{Detailed Test Subset} \label{app_sec:F_1}
We randomly sample $k=10$ retrieval documents for each test instance from the pool of candidate documents according to the specified noise ratio.
For example, when the noise ratio is set to 30\%, 7 golden documents are randomly selected from the corresponding golden set, and 3 noise documents are randomly sampled from the pool of noise documents.
For noise type analysis, $k$ is set to 8 at a noise ratio of 0.9 and 7 at a noise ratio of 1.
As the selection is random, the proportions of different noise types follow the law of large numbers; that is, the ratio among Distracting Noise and the other three noise types (Low Quality, Inconsequential, Irrelevant) is approximately 10:7:7:7. 
Researchers may further customize the types and proportions of injected noise to suit specific experimental requirements.

\subsection{Complete Results of Main Experiments} \label{app_sec:F_2}
\begin{wraptable}{r}{6cm}
\caption{Performance of NoRAG ($MM_m$).}
\label{tab:main_norag_mmm}
\centering
\resizebox{0.4\textwidth}{!}{
\begin{tabular}{lc>{\columncolor{gray!20}}c}
\toprule
\multicolumn{1}{l}{\multirow{2}{*}{\textsc{LLM Generator}}} & \multicolumn{2}{c}{\textsc{NoRAG}} \\
\cmidrule(lr){2-3}
\multicolumn{1}{c}{} & \multicolumn{1}{c}{Cor.} & \multicolumn{1}{c}{Rej.} \\
\midrule
% RGB\cite{RGB} & 2 & \cmark \\
\multicolumn{1}{l|}{Llama-3.2$_{1B}$} & 5.9 & 79.3 \\
\multicolumn{1}{l|}{Llama-3.1$_{8B}$} & 25.1 & 51.7 \\
\multicolumn{1}{l|}{Llama-3.1$_{70B}$} & 47.7 & 14.3 \\
\multicolumn{1}{l|}{R1-Distill-Llama$_{8B}$} & 27.9 & 17.5 \\
\multicolumn{1}{l|}{R1-Distill-Llama$_{70B}$} & 51.1 & 12.7 \\
\bottomrule
\end{tabular}
}
\end{wraptable}
Table~\ref{tab:main_result_complete} presents the complete results of RAG-based systems at different noise ratios on the $MM_s$.
Notably, Llama-3.2$_{1B}$ and Qwen-2.5$_{1.5B}$, the smallest models in their respective series, exhibit distinct behaviors. 
Llama-3.2$_{1B}$ is significantly more conservative than other Llama variants, frequently abstaining from answering—for instance, \textsc{VanillaRAG}+Llama-3.2$_{1B}$ shows a rejection rate of 34.1\% even at a 0\% noise ratio.
This tendency is mitigated only by introducing chain-of-thought prompting. 
In contrast, Qwen-2.5$_{1.5B}$ achieves correctness comparable to much larger generators under low-noise conditions, but is prone to severe hallucinations as noise levels rise.
Tables~\ref{tab:main_norag_mmm} and \ref{tab:main_result_mmm} summarize the results for all systems on the MMm dataset across varying noise ratios.
In the multi-hop QA setting, we observe that the parameter scale of the generator is a key factor for robustness: larger models consistently achieve higher resistance to retrieval noise, possibly due to their enhanced reasoning capabilities and greater contextual modeling capacity.

\begin{table}[ht]
\caption{Complete Results of RAG-based Systems at Different Noise Ratios on the $MM_s$.}
\label{tab:main_result_complete}
\centering
\resizebox{\textwidth}{!}{
\begin{tabular}{l
    c>{\columncolor{gray!20}}c
    c>{\columncolor{gray!20}}c
    c>{\columncolor{gray!20}}c
    c>{\columncolor{gray!20}}c
    c>{\columncolor{gray!20}}c
    c>{\columncolor{gray!20}}c
    c>{\columncolor{gray!20}}c}
\toprule
\multicolumn{1}{l}{\multirow{2}{*}{\textsc{LLM Generator}}} & \multicolumn{2}{c}{0\%} & \multicolumn{2}{c}{10\%} & \multicolumn{2}{c}{30\%} & \multicolumn{2}{c}{50\%} & \multicolumn{2}{c}{70\%} & \multicolumn{2}{c}{90\%} & \multicolumn{2}{c}{100\%} \\ 
\cmidrule(lr){2-3} \cmidrule(lr){4-5} \cmidrule(lr){6-7} \cmidrule(lr){8-9} \cmidrule(lr){10-11} \cmidrule(lr){12-13} \cmidrule(lr){14-15}
\multicolumn{1}{c}{} & \multicolumn{1}{c}{Cor.} & \multicolumn{1}{c}{Rej.} & \multicolumn{1}{c}{Cor.} & \multicolumn{1}{c}{Rej.} & \multicolumn{1}{c}{Cor.} & \multicolumn{1}{c}{Rej.} & \multicolumn{1}{c}{Cor.} & \multicolumn{1}{c}{Rej.} & \multicolumn{1}{c}{Cor.} & \multicolumn{1}{c}{Rej.} & \multicolumn{1}{c}{Cor.} & \multicolumn{1}{c}{Rej.} & \multicolumn{1}{c}{Cor.} & \multicolumn{1}{c}{Rej.} \\ 
\midrule
\multicolumn{15}{c}{\textsc{VanillaRAG}~\cite{Self-RAG}} \\ 
\midrule
\multicolumn{1}{l|}{Llama-3.2$_{1B}$}        & 41.7  & 34.1  & 42.3  & 34.6  & 43.0  & 31.2  & 41.6  & 30.1  & 31.7  & 30.2  & 19.6  & 33.9  & 7.6  & 32.9  \\ 
\multicolumn{1}{l|}{Qwen-2.5$_{1.5B}$}       & 88.0  & 0.6   & 84.6  & 0.4   & 82.5  & 0.7   & 77.3  & 0.7   & 62.2  & 1.1   & 30.9  & 2.3   & 10.5  & 2.7   \\ 
\multicolumn{1}{l|}{Llama-3.1$_{8B}$}        & 84.1  & 2.4   & 81.1  & 2.9   & 76.8  & 3.0   & 73.9  & 2.7   & 59.9  & 3.4   & 44.9  & 4.7   & 18.0  & 5.9   \\ 
\multicolumn{1}{l|}{Qwen-2.5$_{7B}$}         & 72.7  & 13.8  & 70.5  & 14.7  & 67.7  & 15.4  & 62.9  & 18.3  & 41.6  & 30.7  & 27.5  & 37.9  & 7.4   & 45.2  \\
\multicolumn{1}{l|}{Llama-3.1$_{70B}$}       & 81.9  & 6.8   & 81.0  & 6.2   & 77.5  & 5.5   & 75.0  & 5.4   & 68.3  & 6.2   & 58.8  & 8.3   & 28.9  & 16.8  \\ 
\multicolumn{1}{l|}{Qwen-2.5$_{72B}$}        & 84.3  & 9.3   & 83.8  & 8.5   & 82.9  & 8.0   & 79.8  & 8.8   & 74.2  & 12.5  & 52.2  & 25.4  & 14.2  & 45.3  \\ 
\multicolumn{1}{l|}{DeepSeek-V3}         & 84.8  & 8.9   & 84.7  & 7.9   & 84.4  & 6.8   & 82.5  & 6.5   & 77.1  & 8.4   & 61.9  & 12.8  & 17.5  & 30.8  \\
\multicolumn{1}{l|}{DeepSeek-R1}         & 85.2  & 7.8   & 84.9  & 7.1   & 83.9  & 6.4   & 81.5  & 8.7   & 75.1  & 12.0  & 57.9  & 26.6  & 17.9  & 48.6  \\ 
\multicolumn{1}{l|}{R1-Distill-Llama$_{8B}$}  & 84.1  & 5.6   & 81.7  & 5.6   & 78.7  & 5.1   & 74.1  & 5.2   & 64.1  & 7.5   & 45.2  & 11.5  & 22.7  & 13.0  \\
\multicolumn{1}{l|}{R1-Distill-Llama$_{70B}$} & 84.2  & 7.5   & 82.9  & 6.7   & 82.7  & 5.8   & 80.4  & 6.8   & 72.2  & 10.7  & 57.6  & 18.6  & 28.0  & 30.8  \\
\midrule
\multicolumn{15}{c}{\textsc{ChainofNote}~\cite{Chain-of-Note}} \\ 
\midrule
\multicolumn{1}{l|}{Llama-3.2$_{1B}$} & 76.9 & 6.4  & 74.8 & 5.6  & 70.3 & 6.7  & 64.4 & 7.6  & 52.3 & 6.6  & 26.2 & 8.3  & 11.3 & 9.3  \\
\multicolumn{1}{l|}{Qwen-2.5$_{1.5B}$} & 83.5 & 3.7  & 81.5 & 4.1  & 76.4 & 6.2  & 70.8 & 7.9  & 53.1 & 12.9 & 23.1 & 18.3 & 9.4  & 24.0 \\
\multicolumn{1}{l|}{Llama-3.1$_{8B}$}  & 87.1 & 3.5  & 83.2 & 3.5  & 80.2 & 2.4  & 74.9 & 1.7  & 63.7 & 1.5  & 45.2 & 2.1  & 14.5 & 2.9  \\
\multicolumn{1}{l|}{Qwen-2.5$_{7B}$}   & 76.7 & 9.3  & 74.2 & 9.3  & 70.8 & 10.5 & 64.6 & 9.1  & 52.9 & 12.2 & 31.2 & 15.2 & 10.4 & 17.9 \\
\multicolumn{1}{l|}{Llama-3.1$_{70B}$}  & 85.7 & 4.7  & 85.1 & 3.7  & 82.5 & 3.5  & 79.6 & 3.6  & 72.7 & 4.2  & 56.7 & 5.3  & 24.1 & 9.3  \\
\multicolumn{1}{l|}{Qwen-2.5$_{72B}$}   & 84.4 & 7.6  & 83.2 & 6.8  & 80.5 & 6.2  & 75.5 & 5.1  & 67.1 & 6.0  & 45.1 & 11.7 & 16.9 & 21.3 \\
\multicolumn{1}{l|}{DeepSeek-V3}    & 85.5 & 8.4  & 83.8 & 7.1  & 82.3 & 7.2  & 79.7 & 7.1  & 71.2 & 10.5 & 53.8 & 19.4 & 21.8 & 29.7 \\
\multicolumn{1}{l|}{DeepSeek-R1}    & 89.1 & 5.1  & 88.3 & 4.1  & 88.0 & 3.3  & 86.1 & 3.5  & 81.4 & 4.4  & 72.4 & 11.4 & 27.5 & 31.6 \\
\multicolumn{1}{l|}{R1-Distill-Llama$_{8B}$} & 84.4 & 4.0  & 81.2 & 3.0  & 76.7 & 2.0  & 70.6 & 1.8  & 60.7 & 2.5  & 46.2 & 4.0  & 27.6 & 4.8 \\
\multicolumn{1}{l|}{R1-Distill-Llama$_{70B}$} & 85.2 & 5.4  & 83.5 & 4.3  & 82.9 & 3.1  & 79.0 & 2.7  & 73.7 & 4.1  & 63.4 & 6.6  & 34.1 & 11.6 \\
\midrule
\multicolumn{15}{c}{\textsc{DRAGIN}~\cite{DRAGIN}} \\ 
\midrule
\multicolumn{1}{l|}{Llama-3.2$_{1B}$} & 19.3  & 61.7  & 19.0  & 62.1  & 19.1  & 61.6  & 18.9  & 59.8  & 16.8  & 60.9  & 12.7  & 61.6  & 8.7  & 60.2  \\
\multicolumn{1}{l|}{Qwen-2.5$_{1.5B}$} & 54.9  & 9.9   & 53.0  & 9.9   & 51.1  & 9.0   & 49.5  & 9.6   & 42.3  & 9.7   & 27.9  & 10.5  & 16.8  & 10.7  \\
\multicolumn{1}{l|}{Llama-3.1$_{8B}$}  & 57.0  & 16.7  & 56.2  & 15.9  & 54.6  & 16.4  & 53.1  & 15.4  & 49.3  & 16.8  & 41.4  & 15.7  & 30.9  & 17.8  \\
\multicolumn{1}{l|}{Qwen-2.5$_{7B}$}   & 20.3  & 66.8  & 20.7  & 66.2  & 20.3  & 66.1  & 20.2  & 66.2  & 17.4  & 68.3  & 16.3  & 68.2  & 14.2  & 68.4  \\
\multicolumn{1}{l|}{R1-Distill-Llama$_{8B}$} & 56.6  & 7.8   & 55.4  & 9.2   & 55.8  & 6.8   & 55.8  & 6.4   & 50.5  & 7.2   & 35.7  & 8.8   & 21.9  & 9.7  \\
\midrule
\multicolumn{15}{c}{\textsc{SKR}~\cite{SKR}} \\ 
\midrule
\multicolumn{1}{l|}{Llama-3.2$_{1B}$} & 11.2  & 57.8  & 12.0  & 57.6  & 12.4  & 55.5  & 11.6  & 55.1  & 11.9  & 55.8  & 11.8  & 56.6  & 11.5  & 55.4  \\
\multicolumn{1}{l|}{Qwen-2.5$_{1.5B}$} & 31.6  & 8.5   & 30.3  & 8.9   & 29.6  & 8.4   & 30.9  & 8.6   & 28.2  & 10.2  & 20.1  & 9.7   & 15.8  & 9.1   \\
\multicolumn{1}{l|}{Llama-3.1$_{8B}$}  & 41.5  & 11.8  & 40.8  & 12.5  & 40.3  & 12.3  & 39.1  & 13.6  & 38.0  & 13.9  & 38.4  & 13.9  & 36.7  & 13.9  \\
\multicolumn{1}{l|}{Qwen-2.5$_{7B}$}   & 25.5  & 55.8  & 23.8  & 55.8  & 23.8  & 56.6  & 23.1  & 56.5  & 21.3  & 57.5  & 19.5  & 58.7  & 17.1  & 60.2  \\
\multicolumn{1}{l|}{Llama-3.1$_{70B}$}  & 62.7  & 5.1   & 61.1  & 4.8   & 60.6  & 4.4   & 58.3  & 4.5   & 57.8  & 4.8   & 55.9  & 4.8   & 49.4  & 7.3   \\
\multicolumn{1}{l|}{Qwen-2.5$_{72B}$}   & 52.6  & 22.5  & 53.5  & 22.2  & 53.2  & 21.9  & 52.6  & 22.5  & 50.8  & 24.0  & 45.3  & 27.1  & 41.2  & 27.2  \\
\multicolumn{1}{l|}{DeepSeek-V3}    & 53.8  & 16.3  & 54.2  & 16.5  & 54.4  & 16.7  & 54.7  & 14.9  & 53.5  & 16.1  & 52.7  & 17.1  & 50.2  & 17.2  \\
\multicolumn{1}{l|}{DeepSeek-R1}    & 62.4  & 8.1   & 63.7  & 7.5   & 62.6  & 8.2   & 63.5  & 7.1   & 61.6  & 9.1   & 60.4  & 9.4   & 57.4  & 11.6  \\ 
\multicolumn{1}{l|}{R1-Distill-Llama$_{8B}$} & 43.9  & 11.6  & 45.4  & 10.8  & 41.7  & 12.3  & 42.3  & 9.9   & 37.8  & 12.0  & 30.9  & 14.1  & 25.9  & 15.2  \\
\multicolumn{1}{l|}{R1-Distill-Llama$_{70B}$} & 60.0  & 7.9   & 61.1  & 5.9   & 61.2  & 6.2   & 58.7  & 7.3   & 57.2  & 8.1   & 55.4  & 8.9   & 51.4  & 10.6  \\
\midrule
\multicolumn{15}{c}{\textsc{CRAG}~\cite{CRAG}} \\ 
\midrule
\multicolumn{1}{l|}{Llama-3.1$_{8B}$} & 70.9  & 11.5  & 70.6  & 9.8  & 68.2  & 9.2  & 65.4  & 9.1  & 55.4  & 11.1  & 37.3  & 14.1  & 21.7  & 15.6  \\
\multicolumn{1}{l|}{Qwen-2.5$_{7B}$} & 58.7  & 26.9  & 59.1  & 25.5  & 57.5  & 23.7  & 52.9  & 25.1  & 38.3  & 33.4  & 24.1  & 41.9  & 10.0  & 46.6  \\
\multicolumn{1}{l|}{Llama-3.1$_{70B}$} & 73.4  & 9.1   & 75.7  & 8.9   & 72.9  & 7.9   & 70.2  & 7.7   & 62.6  & 9.3   & 51.1  & 11.4  & 33.9  & 15.7  \\
\multicolumn{1}{l|}{Qwen-2.5$_{72B}$} & 70.8  & 23.4  & 72.9  & 19.9  & 72.1  & 18.9  & 68.5  & 18.3  & 58.5  & 26.6  & 33.9  & 42.1  & 13.6  & 53.3  \\
\multicolumn{1}{l|}{DeepSeek-V3} & 71.7  & 20.3  & 73.8  & 17.8  & 72.9  & 16.7  & 71.0  & 15.6  & 60.2  & 21.7  & 39.6  & 31.1  & 15.8  & 40.5  \\
\multicolumn{1}{l|}{DeepSeek-R1} & 72.6  & 18.5  & 74.1  & 16.1  & 73.9  & 15.4  & 73.1  & 14.7  & 60.5  & 23.6  & 37.5  & 38.0  & 17.8  & 47.4  \\
\multicolumn{1}{l|}{R1-Distill-Llama$_{8B}$} & 71.3  & 12.8  & 71.5  & 11.3  & 68.0  & 11.1  & 63.1  & 9.6   & 49.4  & 13.0  & 34.0  & 17.4  & 22.0  & 17.9  \\
\multicolumn{1}{l|}{R1-Distill-Llama$_{70B}$} & 73.9  & 13.8  & 73.7  & 13.7  & 73.4  & 12.1  & 71.4  & 10.9  & 62.1  & 16.3  & 41.9  & 28.1  & 25.9  & 32.8  \\
\bottomrule
\end{tabular}
}
\end{table}

\begin{table}[ht]
\caption{Performance of RAG-based Systems at Different Noise Ratios on the $MM_m$.}
\label{tab:main_result_mmm}
\centering
\resizebox{\textwidth}{!}{
\begin{tabular}{l
    c>{\columncolor{gray!20}}c
    c>{\columncolor{gray!20}}c
    c>{\columncolor{gray!20}}c
    c>{\columncolor{gray!20}}c
    c>{\columncolor{gray!20}}c
    c>{\columncolor{gray!20}}c}
\toprule
\multicolumn{1}{l}{\multirow{2}{*}{\textsc{LLM Generator}}} & \multicolumn{2}{c}{0\%} & \multicolumn{2}{c}{20\%} & \multicolumn{2}{c}{40\%} & \multicolumn{2}{c}{60\%} & \multicolumn{2}{c}{80\%} & \multicolumn{2}{c}{100\%} \\ 
\cmidrule(lr){2-3} \cmidrule(lr){4-5} \cmidrule(lr){6-7} \cmidrule(lr){8-9} \cmidrule(lr){10-11} \cmidrule(lr){12-13}
\multicolumn{1}{c}{} & \multicolumn{1}{c}{Cor.} & \multicolumn{1}{c}{Rej.} & \multicolumn{1}{c}{Cor.} & \multicolumn{1}{c}{Rej.} & \multicolumn{1}{c}{Cor.} & \multicolumn{1}{c}{Rej.} & \multicolumn{1}{c}{Cor.} & \multicolumn{1}{c}{Rej.} & \multicolumn{1}{c}{Cor.} & \multicolumn{1}{c}{Rej.} & \multicolumn{1}{c}{Cor.} & \multicolumn{1}{c}{Rej.} \\ 
\midrule
\multicolumn{13}{c}{\textsc{VanillaRAG}~\cite{Self-RAG}} \\ 
\midrule
\multicolumn{1}{l|}{Llama-3.2$_{1B}$}        & 34.9 & 34.3 & 35.6 & 31.3 & 35.2 & 33.2 & 35.4 & 31.2 & 31.9 & 31.2 & 26.6 & 29.2  \\ 
\multicolumn{1}{l|}{Llama-3.1$_{8B}$}        & 84.1 & 3.0 & 83.1 & 3.5 & 81.3 & 3.8 & 78.3 & 4.8 & 68.7 & 4.0 & 51.2 & 11.5   \\ 
\multicolumn{1}{l|}{Llama-3.1$_{70B}$}       & 90.9 & 2.5 & 91.4 & 1.5 & 91.6 & 1.8 & 92.3 & 1.2 & 80.6 & 1.8 & 65.9 & 5.0  \\ 
\multicolumn{1}{l|}{R1-Distill-Llama$_{8B}$}  & 83.9 & 7.0 & 85.1 & 5.3 & 86.0 & 3.2 & 84.6 & 2.3 & 68.6 & 2.0 & 44.6 & 6.7  \\
\multicolumn{1}{l|}{R1-Distill-Llama$_{70B}$} & 89.0 & 6.3 & 87.9 & 6.0 & 89.7 & 3.5 & 89.8 & 2.8 & 76.9 & 2.8 & 55.1 & 13.2  \\
\midrule
\multicolumn{13}{c}{\textsc{ChainofNote}~\cite{Chain-of-Note}} \\ 
\midrule
\multicolumn{1}{l|}{Llama-3.2$_{1B}$} & 67.5 & 4.0 & 64.7 & 4.5 & 62.1 & 5.2 & 57.8 & 6.5 & 53.0 & 5.3 & 37.7 & 6.8  \\
\multicolumn{1}{l|}{Llama-3.1$_{8B}$}  & 85.5 & 4.7 & 86.0 & 3.2 & 86.7 & 2.3 & 72.8 & 2.2 & 64.5 & 1.5 & 59.7 & 4.3  \\
\multicolumn{1}{l|}{Llama-3.1$_{70B}$}  & 88.5 & 4.5 & 90.9 & 2.5 & 89.9 & 2.7 & 82.0 & 1.8 & 80.8 & 2.3 & 68.7 & 7.2  \\
\multicolumn{1}{l|}{R1-Distill-Llama$_{8B}$} & 83.3 & 4.3 & 81.4 & 3.7 & 79.2 & 1.7 & 68.7 & 1.5 & 62.5 & 1.0 & 50.7 & 5.5 \\
\multicolumn{1}{l|}{R1-Distill-Llama$_{70B}$} & 88.1 & 4.7 & 89.2 & 3.2 & 87.0 & 2.3 & 76.8 & 1.3 & 73.5 & 1.8 & 60.3 & 5.2 \\
\bottomrule
\end{tabular}
}
\end{table}

\subsection{Multi-perspective Analysis of Noise Effects}
To provide a more fine-grained analysis of RAG robustness under noise, we introduce three novel evaluation metrics from different perspectives:
(1) \textbf{Hallucination Rate (H)}: the degree to which answer correctness decreases when retrieval information is introduced to instances that were previously answered correctly without retrieval; 
(2) \textbf{Confusion Rate (C)}: the proportion of instances where the model abstains from answering after retrieval, despite being able to answer correctly without retrieval; and 
(3) \textbf{Rectification Rate (R)}: the degree to which answer correctness increases when retrieval information is introduced to instances that were previously answered incorrectly. These metrics enable a more direct assessment of the nuanced effects of retrieval noise on RAG systems. Complete results are reported in Table~\ref{tab:main_result_multi_perspective}.
Empirical analysis shows that chain-of-thought prompting significantly improves retrieval utility under low-noise conditions but also raises the risk of hallucination as noise increases.
In line with Section~\ref{sec:main_results}, we observe a sharp drop in Rectification Rate when the noise ratio reaches 50\%.
Furthermore, while SKR effectively limits hallucinations under noisy conditions, it also constrains performance in clean scenarios. 
Overall, these findings underscore the difficulty of balancing the benefits of retrieval augmentation with effective noise mitigation in RAG systems.

\begin{table}[ht]
\caption{Multi-perspective analysis of retrieval noise effects on RAG systems, reporting Hallucination Rate (H), Confusion Rate (C), and Rectification Rate (R) at different noise ratios on the $MM_s$.}
\label{tab:main_result_multi_perspective}
\centering
\resizebox{\textwidth}{!}{
\begin{tabular}{l
    c>{\columncolor{gray!20}}c>{\columncolor{gray!40}}c
    c>{\columncolor{gray!20}}c>{\columncolor{gray!40}}c
    c>{\columncolor{gray!20}}c>{\columncolor{gray!40}}c
    c>{\columncolor{gray!20}}c>{\columncolor{gray!40}}c
    c>{\columncolor{gray!20}}c>{\columncolor{gray!40}}c
    c>{\columncolor{gray!20}}c>{\columncolor{gray!40}}c
    c>{\columncolor{gray!20}}c>{\columncolor{gray!40}}c}
\toprule
\multicolumn{1}{l}{\multirow{2}{*}{\textsc{LLM Generator}}} & \multicolumn{3}{c}{0\%} & \multicolumn{3}{c}{10\%} & \multicolumn{3}{c}{30\%} & \multicolumn{3}{c}{50\%} & \multicolumn{3}{c}{70\%} & \multicolumn{3}{c}{90\%} & \multicolumn{3}{c}{100\%} \\ 
\cmidrule(lr){2-4} \cmidrule(lr){5-7} \cmidrule(lr){8-10} \cmidrule(lr){11-13} \cmidrule(lr){14-16} \cmidrule(lr){17-19} \cmidrule(lr){20-22}
\multicolumn{1}{c}{} & \multicolumn{1}{c}{H} & \multicolumn{1}{c}{C} & \multicolumn{1}{c}{R} & \multicolumn{1}{c}{H} & \multicolumn{1}{c}{C} & \multicolumn{1}{c}{R} & \multicolumn{1}{c}{H} & \multicolumn{1}{c}{C} & \multicolumn{1}{c}{R} & \multicolumn{1}{c}{H} & \multicolumn{1}{c}{C} & \multicolumn{1}{c}{R} & \multicolumn{1}{c}{H} & \multicolumn{1}{c}{C} & \multicolumn{1}{c}{R} & \multicolumn{1}{c}{H} & \multicolumn{1}{c}{C} & \multicolumn{1}{c}{R} & \multicolumn{1}{c}{H} & \multicolumn{1}{c}{C} & \multicolumn{1}{c}{R} \\ 
\midrule
\multicolumn{22}{c}{\textsc{VanillaRAG}~\cite{Self-RAG}} \\ 
\midrule
\multicolumn{1}{l|}{Llama-3.2$_{1B}$}        & 2.0 & 2.5 & 35.0 & 1.9 & 2.6 & 35.6 & 2.1 & 1.7 & 35.7 & 2.0 & 2.3 & 34.8 & 3.7 & 1.7 & 25.9 & 4.5 & 1.8 & 14.8 & 6.3 & 2.3 & 5.1  \\ 
\multicolumn{1}{l|}{Qwen-2.5$_{1.5B}$}       & 1.0 & 0.0 & 69.6 & 1.6 & 0.0 & 66.7 & 1.4 & 0.1 & 64.5 & 3.0 & 0.1 & 60.9 & 4.6 & 0.0 & 47.4 & 9.9 & 0.1 & 21.5 & 14.4 & 0.3 & 5.7   \\ 
\multicolumn{1}{l|}{Llama-3.1$_{8B}$}        & 2.4 & 0.2 & 49.9 & 3.1 & 0.3 & 47.7 & 4.0 & 0.1 & 44.1 & 4.8 & 0.0 & 42.0 & 9.6 & 0.4 & 33.1 & 12.8 & 0.4 & 21.4 & 23.4 & 0.8 & 5.5   \\ 
\multicolumn{1}{l|}{Qwen-2.5$_{7B}$}         & 0.7 & 0.7 & 57.8 & 0.9 & 1.0 & 55.9 & 1.0 & 0.7 & 53.2 & 1.6 & 1.0 & 49.2 & 3.0 & 3.0 & 31.2 & 3.8 & 3.4 & 18.3 & 6.3 & 6.8 & 4.1  \\
\multicolumn{1}{l|}{Llama-3.1$_{70B}$}       & 2.3 & 1.4 & 31.0 & 2.7 & 0.9 & 30.0 & 3.3 & 1.0 & 27.1 & 3.7 & 1.2 & 25.3 & 6.5 & 1.7 & 21.8 & 10.6 & 1.9 & 16.5 & 23.0 & 8.0 & 5.3  \\ 
\multicolumn{1}{l|}{Qwen-2.5$_{72B}$}        & 0.9 & 1.4 & 45.6 & 1.3 & 1.4 & 45.4 & 1.3 & 1.3 & 44.4 & 1.8 & 1.4 & 41.9 & 2.4 & 1.9 & 37.4 & 4.4 & 5.9 & 21.5 & 13.5 & 18.0 & 4.7  \\ 
\multicolumn{1}{l|}{DeepSeek-V3}         & 1.4 & 1.8 & 36.4 & 1.6 & 1.2 & 36.0 & 2.0 & 1.2 & 36.0 & 2.5 & 1.2 & 34.5 & 3.1 & 2.1 & 30.8 & 7.1 & 3.8 & 21.3 & 23.1 & 15.1 & 4.1  \\
\multicolumn{1}{l|}{DeepSeek-R1}         & 1.0 & 1.5 & 29.3 & 1.4 & 1.8 & 29.7 & 1.7 & 1.7 & 28.9 & 2.1 & 2.4 & 27.5 & 3.0 & 3.9 & 23.5 & 3.4 & 12.7 & 15.6 & 14.7 & 28.8 & 2.9  \\ 
\multicolumn{1}{l|}{R1-Distill-Llama$_{8B}$}  & 1.3 & 0.8 & 58.6 & 1.6 & 0.7 & 56.5 & 1.6 & 0.5 & 53.3 & 2.1 & 1.1 & 49.8 & 4.2 & 1.3 & 42.1 & 6.4 & 2.1 & 26.2 & 13.9 & 2.5 & 11.5  \\
\multicolumn{1}{l|}{R1-Distill-Llama$_{70B}$} & 1.5 & 1.9 & 33.7 & 1.8 & 1.8 & 32.6 & 1.7 & 1.5 & 31.9 & 2.5 & 1.9 & 30.8 & 4.0 & 3.3 & 25.4 & 6.6 & 7.0 & 15.5 & 16.9 & 14.7 & 5.4  \\
\midrule
\multicolumn{22}{c}{\textsc{ChainofNote}~\cite{Chain-of-Note}} \\ 
\midrule
\multicolumn{1}{l|}{Llama-3.2$_{1B}$} & 0.7 & 0.5 & 67.1 & 0.9 & 0.5 & 65.1 & 1.3 & 0.5 & 61.0 & 1.9 & 0.9 & 56.1 & 2.8 & 0.6 & 44.5 & 4.7 & 1.0 & 20.9 & 7.2 & 1.0 & 8.5  \\
\multicolumn{1}{l|}{Qwen-2.5$_{1.5B}$} & 1.5 & 0.3 & 65.8 & 1.9 & 0.6 & 64.5 & 1.9 & 0.7 & 59.6 & 3.7 & 0.9 & 55.9 & 5.1 & 2.0 & 40.7 & 9.0 & 2.8 & 15.4 & 12.0 & 3.1 & 5.0 \\
\multicolumn{1}{l|}{Llama-3.1$_{8B}$}  & 1.1 & 0.4 & 52.0 & 2.2 & 0.4 & 49.1 & 3.1 & 0.2 & 46.8 & 4.4 & 0.2 & 42.8 & 7.5 & 0.5 & 34.9 & 13.5 & 0.3 & 22.2 & 27.2 & 0.5 & 5.4  \\
\multicolumn{1}{l|}{Qwen-2.5$_{7B}$}   & 0.8 & 0.7 & 61.9 & 1.0 & 0.5 & 59.3 & 1.0 & 0.3 & 55.8 & 1.8 & 0.4 & 50.4 & 3.6 & 0.7 & 41.0 & 6.2 & 0.9 & 22.0 & 10.1 & 2.0 & 6.1 \\
\multicolumn{1}{l|}{Llama-3.1$_{70B}$}  & 2.2 & 0.9 & 34.2 & 2.5 & 0.4 & 33.3 & 2.7 & 0.7 & 30.9 & 3.7 & 0.8 & 29.5 & 7.4 & 1.0 & 25.7 & 13.7 & 1.3 & 16.2 & 30.9 & 4.3 & 4.6  \\
\multicolumn{1}{l|}{Qwen-2.5$_{72B}$}   & 1.5 & 1.6 & 46.5 & 1.8 & 1.0 & 44.9 & 2.9 & 0.9 & 43.3 & 5.9 & 0.7 & 40.9 & 7.4 & 0.5 & 33.9 & 13.3 & 2.4 & 19.8 & 23.1 & 6.7 & 5.6 \\
\multicolumn{1}{l|}{DeepSeek-V3}    & 1.4 & 1.9 & 37.2 & 1.6 & 1.7 & 35.6 & 2.6 & 1.5 & 34.9 & 3.6 & 1.5 & 33.3 & 4.5 & 3.3 & 27.5 & 14.1 & 7.4 & 14.8 & 20.9 & 13.9 & 5.0 \\
\multicolumn{1}{l|}{DeepSeek-R1}    & 0.6 & 0.9 & 32.3 & 1.0 & 0.8 & 31.6 & 1.3 & 0.6 & 31.5 & 2.4 & 0.4 & 30.4 & 2.7 & 1.0 & 26.6 & 3.9 & 3.2 & 20.9 & 17.5 & 19.2 & 5.8 \\
\multicolumn{1}{l|}{R1-Distill-Llama$_{8B}$} & 1.2 & 0.7 & 58.6 & 2.0 & 0.4 & 56.0 & 2.2 & 0.2 & 51.5 & 4.0 & 0.5 & 47.6 & 4.6 & 0.1 & 37.8 & 7.5 & 0.4 & 26.5 & 12.6 & 1.0 & 13.6 \\
\multicolumn{1}{l|}{R1-Distill-Llama$_{70B}$} & 1.8 & 1.2 & 34.1 & 2.7 & 1.0 & 33.2 & 2.8 & 0.7 & 32.2 & 4.9 & 0.4 & 29.5 & 6.0 & 0.9 & 26.0 & 10.1 & 1.8 & 17.7 & 21.5 & 5.1 & 6.4 \\
\midrule
\multicolumn{22}{c}{\textsc{DRAGIN}~\cite{DRAGIN}} \\ 
\midrule
\multicolumn{1}{l|}{Llama-3.2$_{1B}$} & 1.5 & 2.5 & 12.1 & 1.6 & 2.4 & 11.9 & 2.1 & 2.3 & 12.5 & 2.0 & 2.1 & 11.8 & 2.1 & 2.0 & 9.8 & 2.7 & 2.2 & 6.4 & 3.4 & 2.2 & 3.2  \\
\multicolumn{1}{l|}{Qwen-2.5$_{1.5B}$} & 2.0 & 0.3 & 37.8 & 2.3 & 0.3 & 36.1 & 3.1 & 0.3 & 35.0 & 2.9 & 0.3 & 33.1 & 3.5 & 0.4 & 26.7 & 5.4 & 0.4 & 14.3 & 8.1 & 0.5 & 5.9  \\
\multicolumn{1}{l|}{Llama-3.1$_{8B}$}  & 2.4 & 0.7 & 23.4 & 3.2 & 0.6 & 23.1 & 3.2 & 1.0 & 22.0 & 3.1 & 0.5 & 20.0 & 4.3 & 0.7 & 17.6 & 6.4 & 0.8 & 11.9 & 9.4 & 1.0 & 4.6  \\
\multicolumn{1}{l|}{Qwen-2.5$_{7B}$}   & 0.5 & 0.7 & 5.3 & 0.6 & 0.5 & 5.5 & 0.8 & 0.4 & 5.4 & 0.9 & 0.5 & 5.3 & 1.2 & 1.0 & 3.2 & 1.3 & 1.2 & 2.5 & 2.3 & 1.3 & 1.5  \\
% \multicolumn{1}{l|}{R1-Distill-Llama$_{8B}$} & 56.6  & 7.8   & 55.4  & 9.2   & 55.8  & 6.8   & 55.8  & 6.4   & 50.5  & 7.2   & 35.7  & 8.8   & 21.9  & 9.7  \\
\midrule
\multicolumn{22}{c}{\textsc{SKR}~\cite{SKR}} \\ 
\midrule
\multicolumn{1}{l|}{Llama-3.2$_{1B}$} & 2.1 & 1.8 & 4.0 & 2.2 & 1.4 & 4.5 & 2.3 & 1.4 & 5.0 & 2.7 & 1.4 & 4.5 & 2.4 & 1.4 & 4.5 & 2.4 & 1.2 & 4.3 & 2.5 & 1.4 & 4.2  \\
\multicolumn{1}{l|}{Qwen-2.5$_{1.5B}$} & 3.6 & 0.1 & 15.8 & 4.1 & 0.2 & 15.1 & 4.3 & 0.3 & 14.6 & 3.9 & 0.3 & 15.6 & 4.4 & 0.4 & 13.5 & 6.6 & 0.3 & 7.5 & 7.7 & 0.4 & 4.4   \\
\multicolumn{1}{l|}{Llama-3.1$_{8B}$}  & 4.3 & 0.6 & 9.7 & 4.8 & 0.6 & 9.5 & 4.5 & 0.4 & 8.5 & 4.8 & 0.5 & 7.7 & 5.1 & 0.7 & 7.1 & 4.6 & 0.6 & 6.9 & 5.1 & 0.8 & 5.8  \\
\multicolumn{1}{l|}{Qwen-2.5$_{7B}$}   & 0.9 & 0.3 & 10.4 & 1.4 & 0.3 & 9.2 & 1.2 & 0.4 & 9.1 & 1.3 & 0.4 & 8.5 & 1.2 & 0.4 & 6.5 & 1.0 & 0.4 & 4.7 & 1.0 & 0.7 & 2.5  \\
\multicolumn{1}{l|}{Llama-3.1$_{70B}$}  & 5.6 & 0.6 & 14.2 & 6.2 & 0.7 & 13.4 & 5.8 & 0.5 & 12.3 & 6.7 & 0.7 & 11.1 & 6.6 & 0.6 & 10.4 & 7.7 & 0.8 & 9.6 & 9.5 & 1.9 & 6.2   \\
\multicolumn{1}{l|}{Qwen-2.5$_{72B}$}   & 2.1 & 0.7 & 14.3 & 1.6 & 0.7 & 14.7 & 1.5 & 0.7 & 14.3 & 1.6 & 0.8 & 14.0 & 2.1 & 0.5 & 12.3 & 1.9 & 1.0 & 7.1 & 2.0 & 1.5 & 3.6  \\
\multicolumn{1}{l|}{DeepSeek-V3}    & 3.6 & 2.9 & 8.8 & 4.0 & 2.4 & 9.1 & 3.0 & 2.9 & 8.8 & 3.3 & 2.5 & 8.9 & 2.9 & 2.8 & 7.6 & 7.7 & 2.7 & 6.7 & 3.7 & 3.1 & 5.5  \\
\multicolumn{1}{l|}{DeepSeek-R1}    & 3.5 & 1.0 & 8.5 & 3.8 & 0.4 & 9.5 & 3.5 & 0.9 & 8.5 & 3.4 & 0.5 & 9.0 & 3.5 & 1.8 & 8.4 & 3.4 & 1.5 & 6.9 & 3.9 & 2.1 & 4.9  \\ 
\multicolumn{1}{l|}{R1-Distill-Llama$_{8B}$} & 6.7 & 1.2 & 24.3 & 6.7 & 1.2 & 25.9 & 6.2 & 2.9 & 23.3 & 7.1 & 1.4 & 23.3 & 6.6 & 1.4 & 18.3 & 8.5 & 1.4 & 13.4 & 9.2 & 2.2 & 9.8  \\
\multicolumn{1}{l|}{R1-Distill-Llama$_{70B}$} & 6.5 & 1.3 & 13.7 & 6.7 & 0.7 & 13.8 & 6.1 & 1.5 & 14.7 & 7.3 & 1.5 & 12.5 & 6.7 & 2.1 & 11.7 & 7.8 & 2.1 & 10.9 & 8.3 & 2.8 & 8.1  \\
\midrule
\multicolumn{22}{c}{\textsc{CRAG}~\cite{CRAG}} \\ 
\midrule
\multicolumn{1}{l|}{Llama-3.1$_{8B}$} & 3.4 & 2.1 & 39.8 & 4.7 & 1.8 & 40.3 & 5.1 & 1.5 & 38.0 & 5.8 & 1.2 & 35.5 & 8.3 & 2.1 & 29.0 & 14.2 & 2.6 & 17.4 & 18.4 & 3.5 & 6.9  \\
\multicolumn{1}{l|}{Qwen-2.5$_{7B}$} & 0.8 & 2.3 & 45.5 & 1.5 & 2.3 & 46.5 & 1.8 & 1.8 & 44.7 & 2.0 & 2.7 & 41.4 & 3.3 & 3.1 & 28.4 & 4.1 & 4.2 & 16.2 & 5.6 & 5.7 & 5.0  \\
\multicolumn{1}{l|}{Llama-3.1$_{70B}$} & 4.0 & 2.9 & 25.7 & 3.4 & 2.5 & 26.8 & 4.6 & 1.9 & 24.8 & 6.2 & 2.1 & 23.8 & 8.7 & 2.7 & 19.3 & 12.7 & 4.8 & 13.9 & 18.5 & 7.6 & 5.4  \\
\multicolumn{1}{l|}{Qwen-2.5$_{72B}$} & 0.8 & 7.1 & 37.5 & 1.1 & 5.8 & 38.7 & 1.2 & 5.1 & 37.2 & 2.2 & 4.9 & 34.4 & 2.9 & 7.9 & 28.1 & 7.0 & 14.1 & 14.0 & 9.5 & 22.4 & 4.4  \\
\multicolumn{1}{l|}{DeepSeek-V3} & 1.9 & 7.2 & 29.1 & 2.0 & 5.5 & 29.7 & 2.4 & 5.4 & 29.1 & 3.7 & 5.1 & 28.2 & 6.1 & 8.9 & 23.5 & 10.9 & 14.0 & 12.9 & 19.6 & 19.9 & 3.7  \\
\multicolumn{1}{l|}{DeepSeek-R1} & 1.8 & 8.4 & 24.4 & 2.1 & 7.2 & 24.8 & 2.6 & 6.6 & 24.7 & 3.2 & 6.4 & 24.0 & 5.0 & 11.0 & 18.0 & 9.8 & 20.7 & 9.5 & 17.3 & 26.5 & 3.1  \\
\multicolumn{1}{l|}{R1-Distill-Llama$_{8B}$} & 2.6 & 2.5 & 48.7 & 2.8 & 1.6 & 48.4 & 3.1 & 2.0 & 45.6 & 3.7 & 2.0 & 41.2 & 6.0 & 2.5 & 30.3 & 8.7 & 4.6 & 19.8 & 12.6 & 4.3 & 11.4  \\
\multicolumn{1}{l|}{R1-Distill-Llama$_{70B}$} & 2.9 & 5.1 & 27.7 & 3.0 & 5.3 & 27.8 & 3.6 & 3.8 & 26.7 & 4.4 & 3.9 & 25.5 & 5.6 & 6.7 & 20.1 & 11.8 & 11.5 & 10.8 & 17.2 & 16.0 & 4.3  \\
\bottomrule
\end{tabular}
}
\end{table}

\FloatBarrier

\section{Additional Experimental Results}
\subsection{Rejection Rate Analysis} \label{app_sec:rej_rate}
\begin{figure}[htbp]
    \centering
    \begin{subfigure}[b]{\linewidth}
        \centering
        \includegraphics[width=\linewidth]{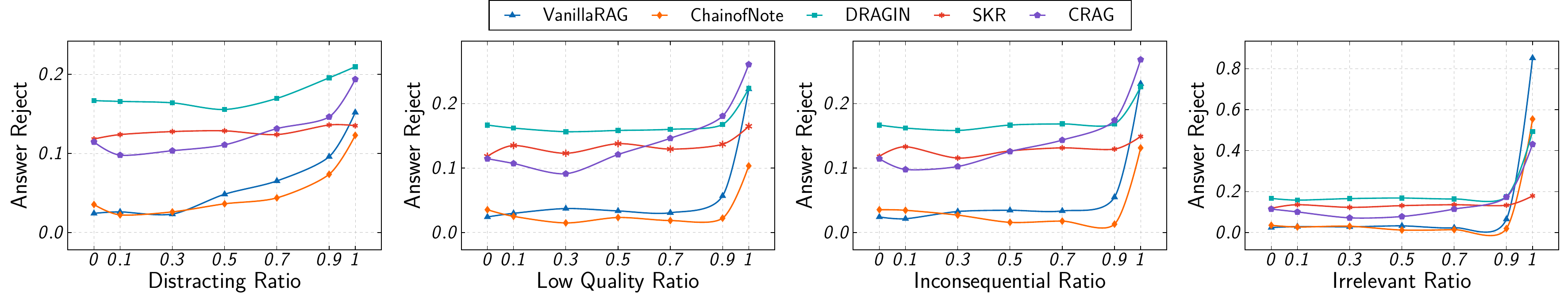}
        % \caption{Caption}
        \label{fig:sub1_rej_Llama}
    \end{subfigure}

    % \vspace{2pt}
    
    \begin{subfigure}[b]{\linewidth}
        \centering
        \includegraphics[width=\linewidth]{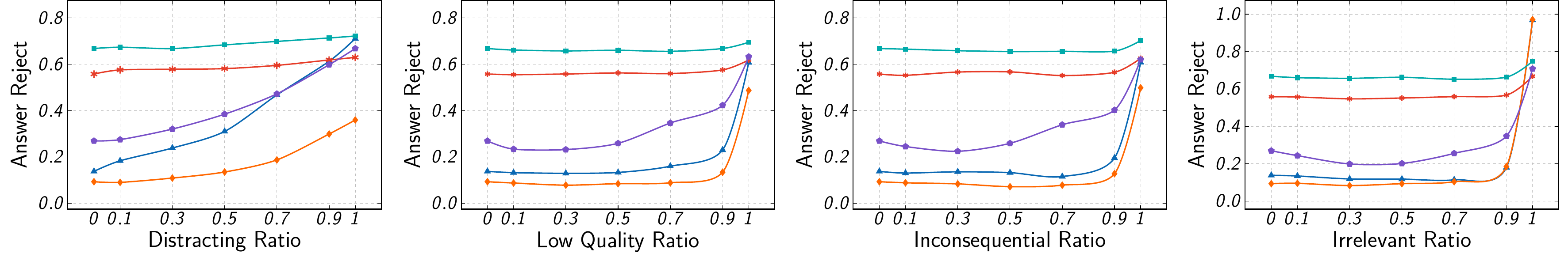}
        % \caption{Caption}
        \label{fig:sub1_rej_Qwen}
    \end{subfigure}
    \caption{Rejection rates of RAG systems under different types of retrieval noise, with Llama-3.1${{8B}}$ (top) and Qwen-2.5${{7B}}$ (bottom) as generators.}
    \label{fig:sub1_rej}
\end{figure}

\begin{figure}[htbp]
    \centering
    \begin{subfigure}[b]{\linewidth}
        \centering
        \includegraphics[width=\linewidth]{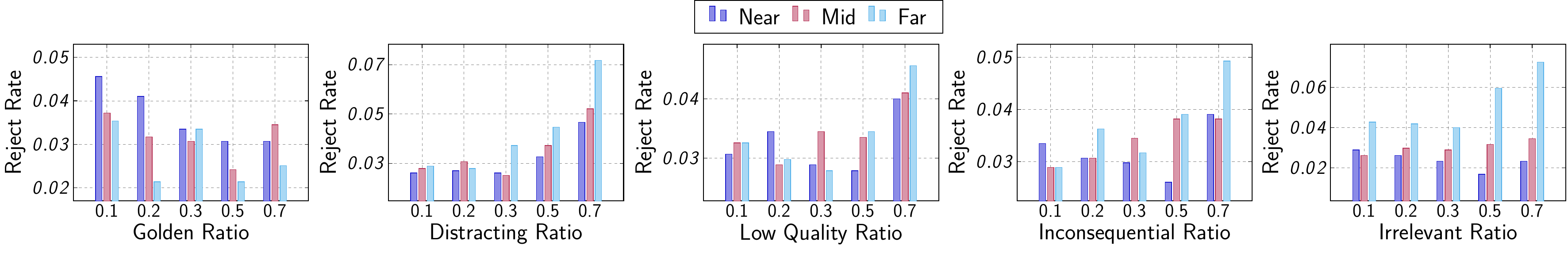}
        % \caption{Caption}
        \label{fig:sub2_rej_Llama}
    \end{subfigure}

    % \vspace{8pt}
    
    \begin{subfigure}[b]{\linewidth}
        \centering
        \includegraphics[width=\linewidth]{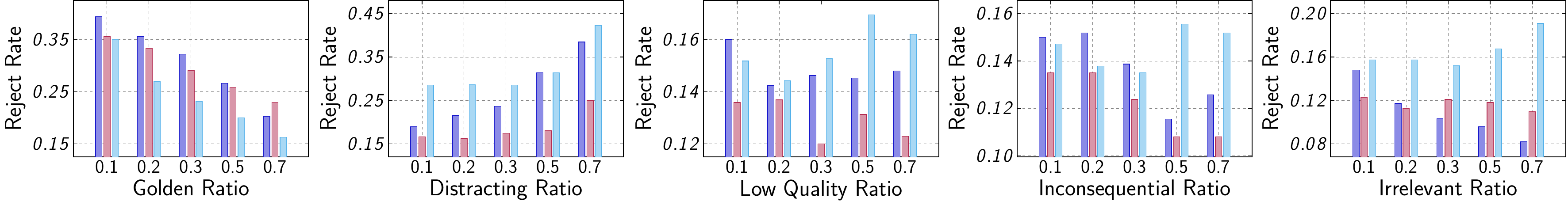}
        % \caption{Caption}
        \label{fig:sub2_rej_Qwen}
    \end{subfigure}
    \caption{Sensitivity of RAG system rejection rates to the position of retrieved documents, using Llama-3.1${{8B}}$ (top) and Qwen-2.5${{7B}}$ (bottom) as LLM generators.}
    \label{fig:sub2_rej}
\end{figure}

Figures~\ref{fig:sub1_rej} and \ref{fig:sub2_rej} illustrate the rejection rates of RAG systems in noise type and sensitivity experiments, respectively.
The rejection rate reflects the model's ability to abstain from answering when uncertain, a critical reliability aspect under noisy conditions. 
Our results show that RAG systems are most sensitive to Distracting Noise, with a marked increase in rejection rates.
The \textsc{ChainofNote} strategy substantially suppresses rejection behavior while reducing abstentions increases the risk of hallucination as noise levels rise. 
In contrast, DRAGIN and SKR maintain consistently high rejection rates, enhancing system stability by ensuring that the model only answers when confident. 
Additionally, we observe that Irrelevant Noise poses minimal threat to current RAG systems, as such noise can be easily identified and filtered by the generators.
% 9-Appendix-sub1_rej

% \subsection{Additional Analysis on Noise Sensitivity}
% 9-Appendix-sub2_rej

\subsection{Noise Distribution Scenarios} \label{app_sec:G_2}
% We conducted a coarse-grained analysis of retrieval content distributions across multiple topics. 
% We designed four representative scenarios based on these empirical statistics, each reflecting a typical noise composition commonly encountered in real-world retrieval environments:
A coarse-grained analysis of retrieval content distributions across multiple topics was conducted (\eg Appendix \ref{app_sec:A}). 
We designed four representative scenarios based on these empirical statistics, each reflecting a typical noise composition commonly encountered in real-world retrieval environments:

\textbf{Scenario 1} (\textsc{Sce.1}) features an even distribution of Golden and Distracting documents (each 30\%), with moderate proportions of Inconsequential (10\%), Low-Quality (10\%), and Irrelevant noise (20\%). 
This configuration simulates a balanced retrieval setting, where useful information is present but intermingled with challenging distractors and irrelevant content. 
For example, in a web search, relevant answers are mixed with well-written yet misleading articles and unrelated advertisements.

\textbf{Scenario 2} (\textsc{Sce.2}) maintains the same proportion of Golden documents (30\%) but reduces Distracting noise to 10\% while increasing Low-Quality (20\%) and Irrelevant noise (30\%). 
This scenario emulates a low-quality retrieval environment, such as forums or poorly-moderated content aggregators, where correct information is outnumbered by low-quality, off-topic, or spam-like content.

\textbf{Scenario 3} (\textsc{Sce.3}) distributes all non-Golden noise types more evenly (D/In/L/Ir: 20\%/20\%/10\%/30\%), with Golden documents reduced to 20\%. 
This setting simulates a noisy and ambiguous retrieval context, where relevant information is scarce and various noise types are distributed uniformly.
For instance, the retrieved evidence often contains a broad mixture of distractors and irrelevant snippets in open-domain QA over large uncurated corpora.

\textbf{Scenario 4} (\textsc{Sce.4}) is characterized by a high proportion of Distracting noise (40\%), low Golden (20\%), moderate Inconsequential (20\%), and minimal Irrelevant and Low-Quality noise (10\% each). 
This configuration is designed to stress-test robustness to confusing distractors, such as in domains where documents are topically relevant but factually incorrect or misleading—\eg scientific misinformation or adversarial content in health-related searches.

% \begin{enumerate}
%     \item 
% \end{enumerate}

\subsection{Supplementary Case Study Analysis} \label{app_sec:case_study}
\begin{figure}[h]
    \centering  
    \includegraphics[width=\linewidth]{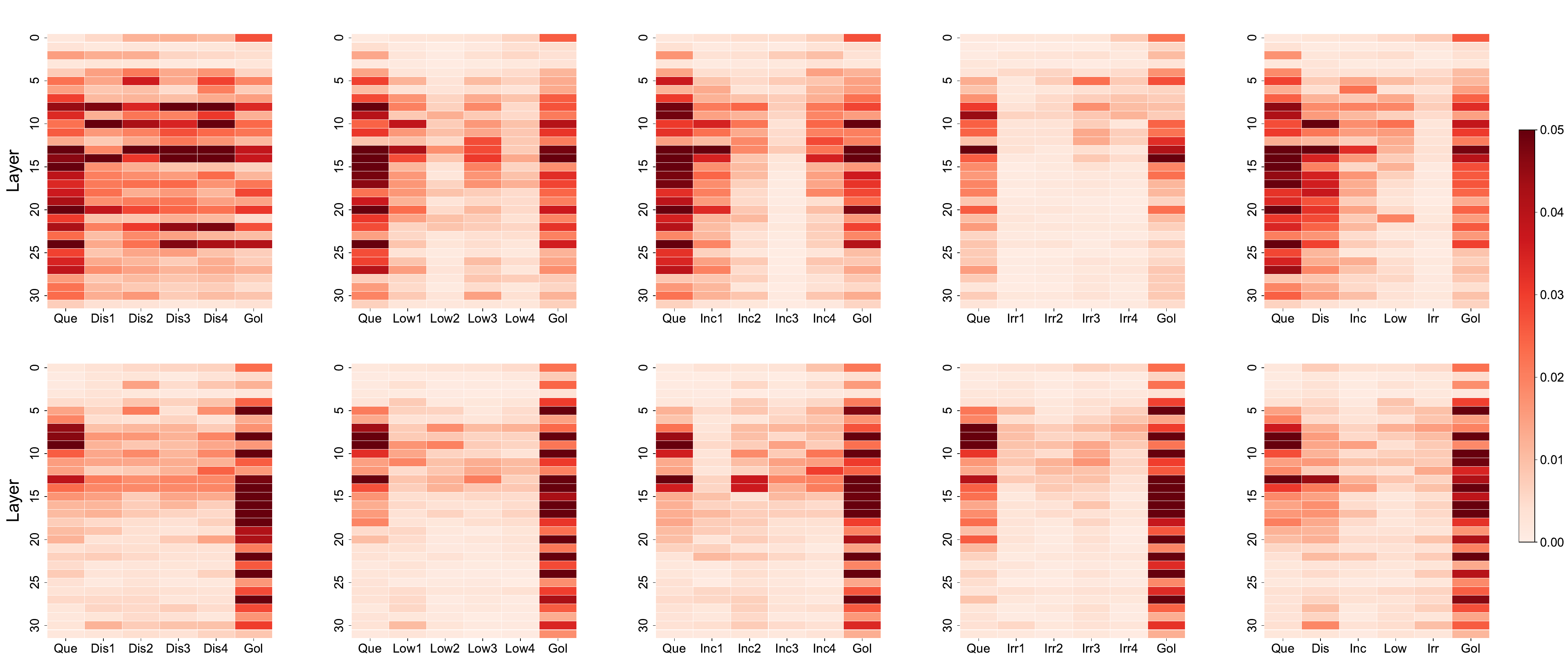}   
    \caption{Layer-wise attention distribution of Llama-3.1${_{8B}}$. }
    \label{fig:heat-plot-appendix}
\end{figure}

We conduct detailed case studies to gain deeper insight into the inner workings of RAG models under different retrieval noise conditions. 
Figure~\ref{fig:heat-plot-appendix} visualizes the layer-wise attention distribution of Llama-3.1${_{8B}}$ over both the query and the various retrieved documents for two questions (``\texttt{How many Beverly Hills cops movies are there?}'' and ``\texttt{When did power rangers tv show come out?}''). 
In question (top), the answer is ``\texttt{three films}''; however, under the influence of distracting noise, the generator produces an uncertain and erroneous response (``\texttt{four or five or six or seven}''), demonstrating the susceptibility of the model to misleading information.
In contrast, in the second question (bottom), noise does not divert the model’s attention from the relevant golden document, and the generator successfully outputs the correct answer ``\texttt{August 28, 1993}''.
This analysis reveals how the model dynamically allocates attention across input components, reflecting its reasoning and evidence integration patterns.
Figures~\ref{fig:The Impact of distracting Noise on RAG}–\ref{fig:The Impact of Low Quality Noise on RAG} present qualitative results from selected test cases, each corresponding to a specific type of retrieval noise. 
For each case, we display the question, the correct answer, representative retrieved documents (including both golden and noisy documents), and the final RAG output. 
These examples illustrate how distracting, inconsequential, and low quality noise can influence the answer-generation process and highlight typical failure modes encountered by RAG systems under noisy conditions.
% 4-sub3_rej

% \subsection{Additional Analysis on Case Study}
% \input{appendix_case_study}
% heat plot

% Example-0
% # Model: Llama-3.1-8B 

\begin{figure}[h]
\setstretch{1.3}
\begin{tcolorbox}[title=The Impact of Distracting Noise, toptitle={0.3em}, bottomtitle={0.1em}]

\textbf{Question:} \emph{Who sings with shaggy on it wasn me?}

\textbf{Correct Answer:} \textcolor{gold}{English-Jamaican singer Rikrok}
\vspace{0.6em}

\textbf{Doc 1:}
\emph{(Golden Document)}
From the 2000 multi-Platinum album Hot Shot by Shaggy, ``It Wasn't Me'' is the first single and includes vocals from \textcolor{gold}{English-Jamaican singer Rikrok}.

\textbf{Doc 2:}
\emph{(Distracting Noise)}
The hit single 'It Wasn't Me' is a collaboration between Jamaican-American reggae artist \textcolor{red}{Shaggy and British-Jamaican vocalist Sean Paul}...

\textbf{Doc 3:}..., \textbf{Doc 4:}...,  \textbf{Doc 5:}...
\vspace{0.6em}

\textbf{RAG Output:} \textcolor{red}{Sean Paul}

\end{tcolorbox} 
\caption{The impact of distracting noise on RAG}
\label{fig:The Impact of distracting Noise on RAG}
\end{figure}

% Example-1
% # Model: Llama-3.1-8B 

\begin{figure}[h]
\setstretch{1.3}
\begin{tcolorbox}[title=The Impact of Inconsequential Noise, toptitle={0.3em}, bottomtitle={0.1em}]

\textbf{Question:} \emph{When were catholic churches allowed back in England?}

\textbf{Correct Answer:} \textcolor{gold}{1850}
\vspace{0.6em}

\textbf{Doc 1:}
\emph{(Golden Document)}
The Catholic Church in England faced severe persecution after the 1559 settlement
... 
Despite these challenges, Catholicism persisted, often practiced in private. 
In \textcolor{gold}{1850}, Pope Pius IX re-established dioceses, marking a significant step in the Church's restoration.

\textbf{Doc 2:}
\emph{(Inconsequential Noise)}
England remained a Catholic country until \textcolor{red}{1534}, when it first officially separated from Rome during the reign of King Henry VIII...

\textbf{Doc 3:}..., \textbf{Doc 4:}...,  \textbf{Doc 5:}...
\vspace{0.6em}

\textbf{RAG Output:} \textcolor{red}{1534}

\end{tcolorbox} 
\caption{The impact of inconsequential noise on RAG}
\label{fig:The Impact of Inconsequential Noise on RAG}
\end{figure}

% Example-2
% # Model: Llama-3.1-8B 

\begin{figure}[h]
\setstretch{1.3}
\begin{tcolorbox}[title=The Impact of Low Quality Noise, toptitle={0.3em}, bottomtitle={0.1em}]

\textbf{Question:} \emph{Who plays male lead in far from the madding crowd?}

\textbf{Correct Answer:} \textcolor{gold}{Matthias Schoenaerts}
\vspace{0.6em}

\textbf{Doc 1:}
\emph{(Golden Document)}
Carey Mulligan is the lead actress in Far from the Madding Crowd, a 2015 British romantic drama directed by Thomas Vinterberg. The film includes performances by \textcolor{gold}{Matthias Schoenaerts}...

\textbf{Doc 2:}
\emph{(Low Quality Noise)}
David Nicholls became attached to the film in 2008. In April 2013, it was reported that \textcolor{red}{Tom Hardy} had been offered the role of Gabriel Oak alongside Carey Mulligan as Bathsheba Everdene. Their casting was official in May 2013 with the participation of director Thomas Vinterberg....

\textbf{Doc 3:}..., \textbf{Doc 4:}...,  \textbf{Doc 5:}...
\vspace{0.6em}

\textbf{RAG Output:} \textcolor{red}{Tom Hardy}

\end{tcolorbox} 
\caption{The impact of low quality noise on RAG}
\label{fig:The Impact of Low Quality Noise on RAG}
\end{figure}

\FloatBarrier

\section{Annotation Guidelines and Details}
To ensure the quality and validity of the constructed dataset, we conducted a rigorous manual verification process for three categories of documents: Golden Documents, Distracting Noise, and Low Quality Noise. This process involved verifying the generated content's factual and structural integrity.
For Golden Documents, human validators checked whether each augmented sentence accurately conveyed the intended event structure in the original golden documents and whether the expression is complete, with no missing or redundant arguments. 
In the case of Distracting Noise, the key criterion is whether the answer entity has been effectively replaced while preserving grammatical and structural coherence. 
For Low Quality Noise, which aimed to introduce semantically plausible but factually incorrect content, human validators verified whether critical entities (\eg people, dates, or locations) had been replaced and whether the sentence structure remained intact.

Each document underwent independent review by two human validators. Only when both human validators agreed that a sample met the specified criteria was it retained; otherwise, it was discarded. 
We provided clear examples of accepted and rejected samples for each noise type to facilitate consistency and accuracy.
The annotation team comprised five experienced human validators, all undergraduate or graduate-level researchers in our laboratory with strong English proficiency. 
Their domain expertise ensured high inter-annotator reliability and helped maintain the dataset’s integrity across all verification stages. To recognize the time and effort involved in the annotation process, contributors were fairly compensated at a rate of \$10 per hour.

%%%%%%%%%%%%%%%%%%%%%%%%%%%%%%%%%%%%%%%%%%%%%%%%%%%%%%%%%%%%

\newpage
﻿
\end{document}

%% file: Prompt/dataset-generation.tex
% SingleHop - Prompt for Golden Document Augmentation
\begin{figure}[h]
\setstretch{1.3}
\begin{tcolorbox}[title=Single-Hop Dataset: Golden Documents Augmentation,
toptitle={0.3em}, bottomtitle={0.1em}]

\textbf{Task Requirements:}

Given a Question, a Short Answer and a Golden Document, 

generate 10 new Golden Documents (1-10) based on the following Principle:

a) You can delete some part of the given Golden Document and add some new details, 

~~~~but you must retain the correct answer in the generate new Golden Documents.

b) Reorganize the sentence structure of the entire Document (difference>80\%), 

~~~~and change the active and passive voice/modifier position/rhetorical structure...

c) Ensure document lengths match original through,

~~~~± 15\% word count variation from Golden Document.

d) Output strictly in the following example format, 

~~~~including all necessary quotes and escape characters.

\textbf{Question:} 

\emph{<question>}

\textbf{Short Answer:} 

\emph{<corresponding answer to the question>}

\textbf{Golden Document:} 

\emph{<origin golden document>}

\textbf{Example Input:}

	\emph{Question: where is zimbabwe located in the world map?}
	
	\emph{Short Answer: in southern Africa, between the Zambezi and Limpopo Rivers...}

	\emph{Golden Document: Zimbabwe, officially the Republic of Zimbabwe, is a landlocked country located in southern Africa, between the Zambezi and Limpopo Rivers, bordered by...}

\textbf{Example Output:}

	\{
	
	\begin{enumerate}[leftmargin=2em, itemindent=0em, label={}]
    	\item \emph{1: "Zimbabwe, a landlocked nation in southern Africa, lies between the Zambezi and Limpopo Rivers. It shares borders with South Africa, Botswana, Zambia, and...",}
       	\item \emph{...}
    	\item \emph{6: "Surrounded by South Africa, Botswana, Zambia, and Mozambique, Zimbabwe is a landlocked country in ... It stretches between the Zambezi and Limpopo Rivers...",}
    	\item \emph{...}
    	\item \emph{10: "In the heart of southern Africa lies Zimbabwe, a landlocked country nestled between the Zambezi and Limpopo Rivers and sharing boundaries with..."}
	\end{enumerate}
	
	\}

\end{tcolorbox} 
\caption{Prompt for golden document augmentation in the single-hop dataset}
\label{fig:Single-Hop Dataset: Golden Documents Augmentation}
\end{figure}

% SingleHop - Distracting Document Generation
\begin{figure}[h]
\setstretch{1.3}
\begin{tcolorbox}[title=Single-Hop Dataset: Distracting Noise Introduction,
toptitle={0.3em}, bottomtitle={0.1em}]

\textbf{Task Requirements:}

1. Given a Question, a Short Answer and a Golden Document, generate 10 Distracting Documents through these stages:

a) Core Entity Substitution:

	\begin{enumerate}[leftmargin=2em, itemindent=0em, label=\textbullet]
    	\item \emph{Identify question-critical entities in Short Answer}
       	\item \emph{Replace with same-domain but incorrect entities using:
           1-Theoretical: Keep discipline but change theory (e.g., "relativity→quantum mechanics") 
           2-Event-based: Maintain event type but shift spatiotemporal coordinates (e.g., "2019→2023")}
	\end{enumerate}

b) Semantic Restructuring (>85\% textual divergence):

	\begin{enumerate}[leftmargin=2em, itemindent=0em, label=\textbullet]
    	\item \emph{Toggle active/passive voice}
       	\item \emph{Each generated Distracting Documents must be different in structure and sentence pattern.}
        \item \emph{Reconfigure rhetorical patterns (e.g., convert causal chains to parallel structures)}
        \item \emph{Reposition modifiers (e.g., transform prepositional phrases to relative clauses)}
	\end{enumerate}

c) Logical Consistency Verification:
       1. Ensure new entities contextually align with:
           - Geographic/chronological parameters
           - Domain-specific terminology
           - Quantitative relationships
       2. Preserve non-critical authentic details from Golden Document for Distracting Documents
       3. Eliminate:
           - Cross-dimensional substitutions (animal→architecture terms)
           - Numerical contradictions (e.g., mismatched magnitude)

2. Output Format:

	\begin{enumerate}[leftmargin=2em, itemindent=0em, label=\textbullet]
    	\item \emph{Generate 10 Distracting Documents (d0-d9)}
       	\item \emph{Each must contain question-related erroneous core entities}
        \item \emph{Maintain original JSON structure with proper escaping}
	\end{enumerate}

3. Critical Avoidances:
    
	\begin{enumerate}[leftmargin=2em, itemindent=0em, label=\textbullet]
    	\item \emph{Cross-category entity swaps}
       	\item \emph{Internal chronological conflicts}
        \item \emph{Commonsense violations (e.g., "fish climb trees")}
	\end{enumerate}

\textbf{Question:} 

\emph{<question>}

\textbf{Short Answer:} 

\emph{<corresponding answer to the question>}

\textbf{Golden Document:} 

\emph{<origin golden document>}

\textbf{Example Input:}

	% \emph{Question: where is zimbabwe located in the world map?}
	
	% \emph{Short Answer: ...}

	% \emph{Golden Document:...}
    \emph{...}

\textbf{Example Output:}

...

\end{tcolorbox} 
\caption{Prompt for distracting noise introduction in the single-hop dataset}
\label{fig:Single-Hop Dataset: Distracting Noise Introduction}
\end{figure}

% SingleHop - Low Document Introduction
\begin{figure}[h]
\setstretch{1.3}
\begin{tcolorbox}[title=SingleHop - Low Quality Noise Generation,
toptitle={0.3em}, bottomtitle={0.1em}]

\textbf{Task Requirements:}

1. Given 7 paragraphs and semantically related Question, please generate a fine-tuned paragraph for each paragraph, by modifying relevant entities or adding relevant expressions to make it contain incorrect knowledge, while ensuring that they are still semantically relevant to the question.

2. Each generated paragraph must meet the following requirements:

\begin{enumerate}[leftmargin=1em, itemindent=0em, label=\textbullet]
   \item Contains incorrect information, but should not violate common sense (for example: do not make obvious mistakes such as "birds swim in water").
   
   \item No internal logical errors are allowed (for example: "The 2019 AFC Asian Cup was held in the United Arab Emirates, and Japan was announced as the host country on March 9, 2015" is self-contradictory).
   
   \item Keep the paragraph fluent and grammatically correct.
   
   \item Try not to include negative sentences in the output content.
\end{enumerate}
   
3. The generated paragraph needs to be semantically relevant to the question, but should not include sentences that can directly answer the question.

4. Please output strictly in the following example format, including all necessary quotes and escape characters.

\textbf{Question:} 

\emph{<question>}

\textbf{Documents:} 

\emph{<related document 1>, <related document 2>,..., <related document 7>}

\textbf{Example Input:}

	\emph{Question: Where is the capital of France?}

	\emph{Document 1: Paris, known for its famous landmarks like the Eiffel Tower...}
	
	\emph{...}
	
	\emph{Document 7: France retains its centuries-long status as a global centre of art, science, and philosophy. It hosts the fourth-largest number of UNESCO World Heritage Sites and is the world's leading tourist....}

\textbf{Example Output:}

	\{
	
	\begin{enumerate}[leftmargin=2em, itemindent=0em, label={}]
    	\item \emph{1: Lyon, known for its famous landmarks like the Eiffel Tower...}
	
		\item \emph{...}
	
		\item \emph{7: France retains its centuries-long status as a global centre of art, science, and philosophy. It hosts the largest number of UNESCO World Heritage Sites and is the world's leading tourist....}

	\end{enumerate}
	
	\}

\end{tcolorbox} 
\caption{Prompt for low quality noise introduction in the single-hop dataset}
\label{fig:Single-Hop Dataset: Low Quality Noise Introduction}
\end{figure}

% MutiHop - Golden Document Generation
\begin{figure}[h]
\setstretch{1.3}
\begin{tcolorbox}[title=Multi-Hop Dataset: Golden Documents Augmentation,
toptitle={0.3em}, bottomtitle={0.1em}]

\textbf{Task Requirements:}

	Please generate <\#document number> documents based on the input document below. 
	Each generated document should have the same meaning as the input document but be rephrased differently.
	Return the results in JSON format where the keys are integers from 1 to <\#document number>, and the values are the rephrased documents.

\textbf{Document:}

	\emph{<origin golden document>}

\textbf{Example Input:}

	\emph{The Radio station currently plays a mix of Hindi and Regional music.}

\textbf{Example Output:}

	\{

	\hspace{2em}\emph{1: "The Radio station is broadcasting a combination of Hindi and Regional music.",}

	\hspace{2em}\emph{2: "A mix of Hindi and Regional music is being played on the Radio station.",}

	\hspace{2em}\emph{3: "The Radio station features both Hindi and Regional music in its current playlist."}

	\}

\end{tcolorbox} 
\caption{Prompt for golden document augmentation in the multi-hop dataset}
\label{fig:Multi-Hop Dataset: Golden Documents Augmentation}
\end{figure}

% MutiHop - Distracting Document Generation
\begin{figure}[h]
\setstretch{1.3}
\begin{tcolorbox}[title=Multi-Hop Dataset: Distracting Noise Introduction,
toptitle={0.3em}, bottomtitle={0.1em}]
\textbf{Task Requirements:}

    Please generate <\#document number> documents based on the input document below. 
    Each generated document should have a different meaning from the input document, potentially leading to incorrect interpretations.
    You can achieve this by modifying key words or phrases in the document while maintaining its overall structure.
    Return the results in JSON format where the keys are integers from 1 to <\#document number>, and the values are the modified documents.

\textbf{Guidelines:}

    - Change the meaning of the document by altering important words or phrases.
    
    - Ensure that the new documents are grammatically correct and coherent.
    
    - Avoid generating documents that are too similar to the input document.
    
\textbf{Document:}

	\emph{<origin golden document>}

\textbf{Example Input:}

    \emph{The Radio station currently plays a mix of Hindi and Regional music.}

\textbf{Example Output:}

    \{
    
        \hspace{2em}\emph{1: "The Radio station currently plays only Hindi music.",}
        
        \hspace{2em}\emph{2: "The Radio station stopped playing Regional music.,}
        
        \hspace{2em}\emph{3: "The Radio station focuses exclusively on English and International music.}
        
    \}
    
\end{tcolorbox} 
\caption{Prompt for distracting noise introduction in the multi-hop dataset}
\label{fig:Multi-Hop Dataset: Distracting Noise Introduction}
\end{figure}
\FloatBarrier

% MutiHop - Low Document Generation
\begin{figure}[htbp]
\setstretch{1.3}
\begin{tcolorbox}[title=Multi-Hop Dataset: Low Quality Noise Introduction,
toptitle={0.3em}, bottomtitle={0.1em}]

\textbf{Task Requirements:}

	Please change the correct information in the input document into wrong information.
	The output should be a document that is similar to the input but contains incorrect information.
	Return the result document directly without any explaination.
	
\textbf{Document:}

	\emph{<origin golden document>}

\textbf{Example Input:}
	
	\emph{The Radio station currently plays a mix of Hindi and Regional music.}

\textbf{Example Output:}

	\emph{The Radio station currently plays a mix of English and Regional music.}

\end{tcolorbox} 
\caption{Prompt for low quality noise introduction in the multi-hop dataset}
\label{fig:Multi-Hop Dataset: Low Quality Noise Introduction}
\end{figure}

%% file: Prompt/RAG-inference.tex
% NoRAG
\begin{figure}[h]
\begin{minipage}{\textwidth}
\setstretch{1.3}
\begin{tcolorbox}[title=NoRAG - Inference Prompt,
toptitle={0.3em}, bottomtitle={0.1em}]

\textbf{Task Description:}

1. Answer the given Question Directly, 

~~~~do NOT add any explanations when giving the response.

2. If you cannot answer with certainty due to insufficient information, 

~~~~you MUST respond verbatim:  ``I cannot answer the question.''

\textbf{Question:} \emph{<question>}

\end{tcolorbox} 
\end{minipage}
\caption{Inference Prompt of \textsc{NoRAG}}
\label{fig:NoRAG Inference Prompt}
\end{figure}

% VanillaRAG
\begin{figure}[h]
\begin{minipage}{\textwidth}
\setstretch{1.3}
\begin{tcolorbox}[title=VanillaRAG - Inference Prompt,
toptitle={0.3em}, bottomtitle={0.1em}]

\textbf{Task Description:}

1. Answer the given Question based on the Retrieval Documents, 

~~~~do NOT add any explanations when giving the response.
        
2. If you cannot answer with certainty due to insufficient information, 

~~~~you MUST respond verbatim:  ``I cannot answer the question.''

\textbf{Question:} \emph{<question>}

\textbf{Retrieval Documents:} \emph{<retrieval documents>}

\end{tcolorbox} 
\end{minipage}
\caption{Inference Prompt of \textsc{VanillaRAG}}
\label{fig:VanillaRAG Inference Prompt}
\end{figure}

% Chain of Note
\begin{figure}[h]
\begin{minipage}{\textwidth}
\setstretch{1.3}
\begin{tcolorbox}[title=ChainofNote - Inference Prompt,
toptitle={0.3em}, bottomtitle={0.1em}]

\textbf{Task Description:}

1. Read the given Question and Retrieval Documents to gather relevant information.

2. Write reading notes summarizing the key points from these passages.
        
3. Discuss the relevance of the given question and Wikipedia passages.

4. If some passages are relevant to the given question, 

~~~~provide a brief answer based on the passages.

5. If no passage is relevant, directly provide answer without considering the passages.

6. If you cannot answer with certainty due to insufficient information, 

~~~~you MUST respond verbatim:  ``I cannot answer the question.''

\textbf{Question:} \emph{<question>}

\textbf{Retrieval Documents:} \emph{<retrieval documents>}

\end{tcolorbox} 
\end{minipage}
\caption{Inference Prompt of \textsc{ChainofNote}}
\label{fig:ChainofNote Inference Prompt}
\end{figure}

% SKR
\begin{figure}[h]
\begin{minipage}{\textwidth}
\setstretch{1.3}
\begin{tcolorbox}[title=SKR - Retrieval Check Prompt,
toptitle={0.3em}, bottomtitle={0.1em}]

\textbf{Task Description:}

Do you need additional information to answer this question?
        
If you need, please answer: ``Yes, I need.''

If you don't need, please answer: ``No, I don’t need.''
        
Do not answer questions and explain reasons.

\textbf{Question:} \emph{<question>}

\end{tcolorbox} 
\end{minipage}
\caption{Prompt to check whether LLM can solve the question without retrieval document in  \textsc{SKR}}
\label{fig:SKR Prompt}
\end{figure}

%% file: Prompt/evaluation-metrics.tex
\begin{figure}[h]
\begin{minipage}{\textwidth}
\setstretch{1.3}
\begin{tcolorbox}[title=Prompt for the Evaluation of Correctness,
toptitle={0.3em}, bottomtitle={0.1em}]

You are an evaluator tasked with scoring the Candidate Answer.

\textbf{Instructions:}

1. Compare the Candidate Answer to the Correct Answer in the context of the Question.

2. Assign a score from 0 to 5 based on accuracy and completeness:

\begin{enumerate}[leftmargin=1em, itemindent=0em, label={}]
   \item \emph{Score 0:} Completely incorrect or irrelevant.
   \item \emph{Score 1:} Partially correct but contains significant errors.
   \item \emph{Score 3:} Moderately correct but lacks some details or precision.
   \item \emph{Score 5:} Fully correct and matches the Correct Answer.
\end{enumerate}

3. Output ONLY the score as a single number (e.g. ``3''). Do not include any explanations.

\textbf{Question:} <question>

\textbf{Correct Answer:} <correct answer>

\textbf{Candidate Answer:} <llm answer>

\end{tcolorbox} 
\end{minipage}
\caption{Prompt for the Evaluation of Correctness}
\label{fig:Prompt for the Evaluation of Correctness}
\end{figure}

%% file: neurips_2025.bbl
\begin{thebibliography}{45}
\providecommand{\natexlab}[1]{#1}
\providecommand{\url}[1]{\texttt{#1}}
\expandafter\ifx\csname urlstyle\endcsname\relax
  \providecommand{\doi}[1]{doi: #1}\else
  \providecommand{\doi}{doi: \begingroup \urlstyle{rm}\Url}\fi

\bibitem[An et~al.(2021)An, Zhong, Geng, Yang, and Qiu]{RetrievalSum}
Chenxin An, Ming Zhong, Zhichao Geng, Jianqiang Yang, and Xipeng Qiu.
\newblock Retrievalsum: {A} retrieval enhanced framework for abstractive summarization.
\newblock \emph{CoRR}, abs/2109.07943, 2021.
\newblock URL \url{https://arxiv.org/abs/2109.07943}.

\bibitem[Asai et~al.(2024)Asai, Wu, Wang, Sil, and Hajishirzi]{Self-RAG}
Akari Asai, Zeqiu Wu, Yizhong Wang, Avirup Sil, and Hannaneh Hajishirzi.
\newblock Self-rag: Learning to retrieve, generate, and critique through self-reflection.
\newblock In \emph{The Twelfth International Conference on Learning Representations, {ICLR} 2024, Vienna, Austria, May 7-11, 2024}. OpenReview.net, 2024.
\newblock URL \url{https://openreview.net/forum?id=hSyW5go0v8}.

\bibitem[Chen et~al.(2024)Chen, Lin, Han, and Sun]{RGB}
Jiawei Chen, Hongyu Lin, Xianpei Han, and Le~Sun.
\newblock Benchmarking large language models in retrieval-augmented generation.
\newblock In Michael~J. Wooldridge, Jennifer~G. Dy, and Sriraam Natarajan, editors, \emph{Thirty-Eighth {AAAI} Conference on Artificial Intelligence, {AAAI} 2024, Thirty-Sixth Conference on Innovative Applications of Artificial Intelligence, {IAAI} 2024, Fourteenth Symposium on Educational Advances in Artificial Intelligence, {EAAI} 2014, February 20-27, 2024, Vancouver, Canada}, pages 17754--17762. {AAAI} Press, 2024.
\newblock \doi{10.1609/AAAI.V38I16.29728}.
\newblock URL \url{https://doi.org/10.1609/aaai.v38i16.29728}.

\bibitem[Cuconasu et~al.(2024)Cuconasu, Trappolini, Siciliano, Filice, Campagnano, Maarek, Tonellotto, and Silvestri]{power_noise}
Florin Cuconasu, Giovanni Trappolini, Federico Siciliano, Simone Filice, Cesare Campagnano, Yoelle Maarek, Nicola Tonellotto, and Fabrizio Silvestri.
\newblock The power of noise: Redefining retrieval for {RAG} systems.
\newblock In Grace~Hui Yang, Hongning Wang, Sam Han, Claudia Hauff, Guido Zuccon, and Yi~Zhang, editors, \emph{Proceedings of the 47th International {ACM} {SIGIR} Conference on Research and Development in Information Retrieval, {SIGIR} 2024, Washington DC, USA, July 14-18, 2024}, pages 719--729. {ACM}, 2024.
\newblock \doi{10.1145/3626772.3657834}.
\newblock URL \url{https://doi.org/10.1145/3626772.3657834}.

\bibitem[DeepSeek{-}AI et~al.(2024)DeepSeek{-}AI, Liu, Feng, Xue, Wang, Wu, Lu, Zhao, and et~al.]{Deepseek-V3}
DeepSeek{-}AI, Aixin Liu, Bei Feng, Bing Xue, Bingxuan Wang, Bochao Wu, Chengda Lu, Chenggang Zhao, and et~al.
\newblock Deepseek-v3 technical report.
\newblock \emph{CoRR}, abs/2412.19437, 2024.
\newblock \doi{10.48550/ARXIV.2412.19437}.
\newblock URL \url{https://doi.org/10.48550/arXiv.2412.19437}.

\bibitem[DeepSeek{-}AI et~al.(2025)DeepSeek{-}AI, Guo, Yang, Zhang, Song, Zhang, Xu, Zhu, and et~al.]{Deepseek-R1}
DeepSeek{-}AI, Daya Guo, Dejian Yang, Haowei Zhang, Junxiao Song, Ruoyu Zhang, Runxin Xu, Qihao Zhu, and et~al.
\newblock Deepseek-r1: Incentivizing reasoning capability in llms via reinforcement learning.
\newblock \emph{CoRR}, abs/2501.12948, 2025.
\newblock \doi{10.48550/ARXIV.2501.12948}.
\newblock URL \url{https://doi.org/10.48550/arXiv.2501.12948}.

\bibitem[Dubey et~al.(2024)Dubey, Jauhri, Pandey, Kadian, Al{-}Dahle, Letman, Mathur, Stone, and et~al.]{Llama-3}
Abhimanyu Dubey, Abhinav Jauhri, Abhinav Pandey, Abhishek Kadian, Ahmad Al{-}Dahle, Aiesha Letman, Akhil Mathur, Kevin Stone, and et~al.
\newblock The llama 3 herd of models.
\newblock \emph{CoRR}, abs/2407.21783, 2024.
\newblock \doi{10.48550/ARXIV.2407.21783}.
\newblock URL \url{https://doi.org/10.48550/arXiv.2407.21783}.

\bibitem[Fang et~al.(2024)Fang, Bai, Ni, Yang, Chen, and Xu]{raat}
Feiteng Fang, Yuelin Bai, Shiwen Ni, Min Yang, Xiaojun Chen, and Ruifeng Xu.
\newblock Enhancing noise robustness of retrieval-augmented language models with adaptive adversarial training.
\newblock In Lun{-}Wei Ku, Andre Martins, and Vivek Srikumar, editors, \emph{Proceedings of the 62nd Annual Meeting of the Association for Computational Linguistics (Volume 1: Long Papers), {ACL} 2024, Bangkok, Thailand, August 11-16, 2024}, pages 10028--10039. Association for Computational Linguistics, 2024.
\newblock \doi{10.18653/V1/2024.ACL-LONG.540}.
\newblock URL \url{https://doi.org/10.18653/v1/2024.acl-long.540}.

\bibitem[Gao et~al.(2023)Gao, Xiong, Gao, Jia, Pan, Bi, Dai, Sun, Guo, Wang, and Wang]{rag-survey}
Yunfan Gao, Yun Xiong, Xinyu Gao, Kangxiang Jia, Jinliu Pan, Yuxi Bi, Yi~Dai, Jiawei Sun, Qianyu Guo, Meng Wang, and Haofen Wang.
\newblock Retrieval-augmented generation for large language models: {A} survey.
\newblock \emph{CoRR}, abs/2312.10997, 2023.
\newblock \doi{10.48550/ARXIV.2312.10997}.
\newblock URL \url{https://doi.org/10.48550/arXiv.2312.10997}.

\bibitem[Guu et~al.(2020)Guu, Lee, Tung, Pasupat, and Chang]{REALM}
Kelvin Guu, Kenton Lee, Zora Tung, Panupong Pasupat, and Ming{-}Wei Chang.
\newblock {REALM:} retrieval-augmented language model pre-training.
\newblock \emph{CoRR}, abs/2002.08909, 2020.
\newblock URL \url{https://arxiv.org/abs/2002.08909}.

\bibitem[Han et~al.(2024)Han, Zhang, Qi, Xu, Wang, Liu, Wang, Min, and Castelli]{Arena}
Rujun Han, Yuhao Zhang, Peng Qi, Yumo Xu, Jenyuan Wang, Lan Liu, William~Yang Wang, Bonan Min, and Vittorio Castelli.
\newblock {RAG-QA} arena: Evaluating domain robustness for long-form retrieval augmented question answering.
\newblock In Yaser Al{-}Onaizan, Mohit Bansal, and Yun{-}Nung Chen, editors, \emph{Proceedings of the 2024 Conference on Empirical Methods in Natural Language Processing, {EMNLP} 2024, Miami, FL, USA, November 12-16, 2024}, pages 4354--4374. Association for Computational Linguistics, 2024.
\newblock URL \url{https://aclanthology.org/2024.emnlp-main.249}.

\bibitem[Hosking et~al.(2024)Hosking, Tang, and Lapata]{Hierarchical-Indexing}
Tom Hosking, Hao Tang, and Mirella Lapata.
\newblock Hierarchical indexing for retrieval-augmented opinion summarization.
\newblock \emph{Trans. Assoc. Comput. Linguistics}, 12:\penalty0 1533--1555, 2024.
\newblock \doi{10.1162/TACL\_A\_00703}.
\newblock URL \url{https://doi.org/10.1162/tacl\_a\_00703}.

\bibitem[Hou et~al.(2024)Hou, Pascale, Carnerero{-}Cano, Tchrakian, Marinescu, Daly, Padhi, and Sattigeri]{WikiContradict}
Yufang Hou, Alessandra Pascale, Javier Carnerero{-}Cano, Tigran~T. Tchrakian, Radu Marinescu, Elizabeth Daly, Inkit Padhi, and Prasanna Sattigeri.
\newblock Wikicontradict: {A} benchmark for evaluating llms on real-world knowledge conflicts from wikipedia.
\newblock In Amir Globersons, Lester Mackey, Danielle Belgrave, Angela Fan, Ulrich Paquet, Jakub~M. Tomczak, and Cheng Zhang, editors, \emph{Advances in Neural Information Processing Systems 38: Annual Conference on Neural Information Processing Systems 2024, NeurIPS 2024, Vancouver, BC, Canada, December 10 - 15, 2024}, 2024.
\newblock URL \url{http://papers.nips.cc/paper\_files/paper/2024/hash/c63819755591ea972f8570beffca6b1b-Abstract-Datasets\_and\_Benchmarks\_Track.html}.

\bibitem[Hsia et~al.(2024)Hsia, Shaikh, Wang, and Neubig]{ragged}
Jennifer Hsia, Afreen Shaikh, Zhiruo Wang, and Graham Neubig.
\newblock {RAGGED:} towards informed design of retrieval augmented generation systems.
\newblock \emph{CoRR}, abs/2403.09040, 2024.
\newblock \doi{10.48550/ARXIV.2403.09040}.
\newblock URL \url{https://doi.org/10.48550/arXiv.2403.09040}.

\bibitem[Iyyer et~al.(2018)Iyyer, Wieting, Gimpel, and Zettlemoyer]{SCPN}
Mohit Iyyer, John Wieting, Kevin Gimpel, and Luke Zettlemoyer.
\newblock Adversarial example generation with syntactically controlled paraphrase networks.
\newblock In Marilyn~A. Walker, Heng Ji, and Amanda Stent, editors, \emph{Proceedings of the 2018 Conference of the North American Chapter of the Association for Computational Linguistics: Human Language Technologies, {NAACL-HLT} 2018, New Orleans, Louisiana, USA, June 1-6, 2018, Volume 1 (Long Papers)}, pages 1875--1885. Association for Computational Linguistics, 2018.
\newblock \doi{10.18653/V1/N18-1170}.
\newblock URL \url{https://doi.org/10.18653/v1/n18-1170}.

\bibitem[Izacard et~al.(2022)Izacard, Caron, Hosseini, Riedel, Bojanowski, Joulin, and Grave]{Contriever}
Gautier Izacard, Mathilde Caron, Lucas Hosseini, Sebastian Riedel, Piotr Bojanowski, Armand Joulin, and Edouard Grave.
\newblock Unsupervised dense information retrieval with contrastive learning.
\newblock \emph{Trans. Mach. Learn. Res.}, 2022, 2022.
\newblock URL \url{https://openreview.net/forum?id=jKN1pXi7b0}.

\bibitem[Jin et~al.(2024)Jin, Yoon, Han, and Arik]{meta}
Bowen Jin, Jinsung Yoon, Jiawei Han, and Sercan~{\"{O}}. Arik.
\newblock Long-context llms meet {RAG:} overcoming challenges for long inputs in {RAG}.
\newblock \emph{CoRR}, abs/2410.05983, 2024.
\newblock \doi{10.48550/ARXIV.2410.05983}.
\newblock URL \url{https://doi.org/10.48550/arXiv.2410.05983}.

\bibitem[Karpukhin et~al.(2020)Karpukhin, Oguz, Min, Lewis, Wu, Edunov, Chen, and Yih]{DPR}
Vladimir Karpukhin, Barlas Oguz, Sewon Min, Patrick Lewis, Ledell Wu, Sergey Edunov, Danqi Chen, and Wen-tau Yih.
\newblock Dense passage retrieval for open-domain question answering.
\newblock In Bonnie Webber, Trevor Cohn, Yulan He, and Yang Liu, editors, \emph{Proceedings of the 2020 Conference on Empirical Methods in Natural Language Processing (EMNLP)}, pages 6769--6781, Online, November 2020. Association for Computational Linguistics.
\newblock \doi{10.18653/v1/2020.emnlp-main.550}.
\newblock URL \url{https://aclanthology.org/2020.emnlp-main.550/}.

\bibitem[Kwiatkowski et~al.(2019)Kwiatkowski, Palomaki, Redfield, Collins, Parikh, Alberti, Epstein, Polosukhin, Devlin, Lee, Toutanova, Jones, Kelcey, Chang, Dai, Uszkoreit, Le, and Petrov]{NQ_dataset}
Tom Kwiatkowski, Jennimaria Palomaki, Olivia Redfield, Michael Collins, Ankur~P. Parikh, Chris Alberti, Danielle Epstein, Illia Polosukhin, Jacob Devlin, Kenton Lee, Kristina Toutanova, Llion Jones, Matthew Kelcey, Ming{-}Wei Chang, Andrew~M. Dai, Jakob Uszkoreit, Quoc Le, and Slav Petrov.
\newblock Natural questions: a benchmark for question answering research.
\newblock \emph{Trans. Assoc. Comput. Linguistics}, 7:\penalty0 452--466, 2019.
\newblock \doi{10.1162/TACL\_A\_00276}.
\newblock URL \url{https://doi.org/10.1162/tacl\_a\_00276}.

\bibitem[Lewis et~al.(2020{\natexlab{a}})Lewis, Perez, Piktus, Petroni, Karpukhin, Goyal, K{\"{u}}ttler, Lewis, Yih, Rockt{\"{a}}schel, Riedel, and Kiela]{RAG}
Patrick Lewis, Ethan Perez, Aleksandra Piktus, Fabio Petroni, Vladimir Karpukhin, Naman Goyal, Heinrich K{\"{u}}ttler, Mike Lewis, Wen{-}tau Yih, Tim Rockt{\"{a}}schel, Sebastian Riedel, and Douwe Kiela.
\newblock Retrieval-augmented generation for knowledge-intensive {NLP} tasks.
\newblock In Hugo Larochelle, Marc'Aurelio Ranzato, Raia Hadsell, Maria{-}Florina Balcan, and Hsuan{-}Tien Lin, editors, \emph{Advances in Neural Information Processing Systems 33: Annual Conference on Neural Information Processing Systems 2020, NeurIPS 2020, December 6-12, 2020, virtual}, 2020{\natexlab{a}}.
\newblock URL \url{https://proceedings.neurips.cc/paper/2020/hash/6b493230205f780e1bc26945df7481e5-Abstract.html}.

\bibitem[Lewis et~al.(2020{\natexlab{b}})Lewis, Perez, Piktus, Petroni, Karpukhin, Goyal, K{\"{u}}ttler, Lewis, Yih, Rockt{\"{a}}schel, Riedel, and Kiela]{knowledge-intensive-nlp-tasks}
Patrick Lewis, Ethan Perez, Aleksandra Piktus, Fabio Petroni, Vladimir Karpukhin, Naman Goyal, Heinrich K{\"{u}}ttler, Mike Lewis, Wen{-}tau Yih, Tim Rockt{\"{a}}schel, Sebastian Riedel, and Douwe Kiela.
\newblock Retrieval-augmented generation for knowledge-intensive {NLP} tasks.
\newblock In Hugo Larochelle, Marc'Aurelio Ranzato, Raia Hadsell, Maria{-}Florina Balcan, and Hsuan{-}Tien Lin, editors, \emph{Advances in Neural Information Processing Systems 33: Annual Conference on Neural Information Processing Systems 2020, NeurIPS 2020, December 6-12, 2020, virtual}, 2020{\natexlab{b}}.
\newblock URL \url{https://proceedings.neurips.cc/paper/2020/hash/6b493230205f780e1bc26945df7481e5-Abstract.html}.

\bibitem[Li et~al.(2024)Li, Mei, Liu, Yan, Wang, Yu, Zeng, Chen, Yu, Liu, Sun, and Xiong]{RAG-DDR}
Xinze Li, Sen Mei, Zhenghao Liu, Yukun Yan, Shuo Wang, Shi Yu, Zheni Zeng, Hao Chen, Ge~Yu, Zhiyuan Liu, Maosong Sun, and Chenyan Xiong.
\newblock {RAG-DDR:} optimizing retrieval-augmented generation using differentiable data rewards.
\newblock \emph{CoRR}, abs/2410.13509, 2024.
\newblock \doi{10.48550/ARXIV.2410.13509}.
\newblock URL \url{https://doi.org/10.48550/arXiv.2410.13509}.

\bibitem[Liu et~al.(2023)Liu, Huang, Li, Chen, Zhou, Meng, Zhou, and Sun]{RECALL}
Yi~Liu, Lianzhe Huang, Shicheng Li, Sishuo Chen, Hao Zhou, Fandong Meng, Jie Zhou, and Xu~Sun.
\newblock {RECALL:} {A} benchmark for llms robustness against external counterfactual knowledge.
\newblock \emph{CoRR}, abs/2311.08147, 2023.
\newblock \doi{10.48550/ARXIV.2311.08147}.
\newblock URL \url{https://doi.org/10.48550/arXiv.2311.08147}.

\bibitem[OpenAI(2023)]{gpt4}
OpenAI.
\newblock {GPT-4} technical report.
\newblock \emph{CoRR}, abs/2303.08774, 2023.
\newblock \doi{10.48550/ARXIV.2303.08774}.
\newblock URL \url{https://doi.org/10.48550/arXiv.2303.08774}.

\bibitem[Rajpurkar et~al.(2016)Rajpurkar, Zhang, Lopyrev, and Liang]{EM-F1}
Pranav Rajpurkar, Jian Zhang, Konstantin Lopyrev, and Percy Liang.
\newblock Squad: 100, 000+ questions for machine comprehension of text.
\newblock In Jian Su, Xavier Carreras, and Kevin Duh, editors, \emph{Proceedings of the 2016 Conference on Empirical Methods in Natural Language Processing, {EMNLP} 2016, Austin, Texas, USA, November 1-4, 2016}, pages 2383--2392. The Association for Computational Linguistics, 2016.
\newblock \doi{10.18653/V1/D16-1264}.
\newblock URL \url{https://doi.org/10.18653/v1/d16-1264}.

\bibitem[Reimers and Gurevych(2019)]{Sentence-BERT}
Nils Reimers and Iryna Gurevych.
\newblock Sentence-{BERT}: Sentence embeddings using {S}iamese {BERT}-networks.
\newblock In Kentaro Inui, Jing Jiang, Vincent Ng, and Xiaojun Wan, editors, \emph{Proceedings of the 2019 Conference on Empirical Methods in Natural Language Processing and the 9th International Joint Conference on Natural Language Processing (EMNLP-IJCNLP)}, pages 3982--3992, Hong Kong, China, November 2019. Association for Computational Linguistics.
\newblock \doi{10.18653/v1/D19-1410}.
\newblock URL \url{https://aclanthology.org/D19-1410/}.

\bibitem[Robertson and Zaragoza(2009)]{BM25}
Stephen~E. Robertson and Hugo Zaragoza.
\newblock The probabilistic relevance framework: {BM25} and beyond.
\newblock \emph{Found. Trends Inf. Retr.}, 3\penalty0 (4):\penalty0 333--389, 2009.
\newblock \doi{10.1561/1500000019}.
\newblock URL \url{https://doi.org/10.1561/1500000019}.

\bibitem[Su et~al.(2024)Su, Tang, Ai, Wu, and Liu]{DRAGIN}
Weihang Su, Yichen Tang, Qingyao Ai, Zhijing Wu, and Yiqun Liu.
\newblock {DRAGIN:} dynamic retrieval augmented generation based on the real-time information needs of large language models.
\newblock In Lun{-}Wei Ku, Andre Martins, and Vivek Srikumar, editors, \emph{Proceedings of the 62nd Annual Meeting of the Association for Computational Linguistics (Volume 1: Long Papers), {ACL} 2024, Bangkok, Thailand, August 11-16, 2024}, pages 12991--13013. Association for Computational Linguistics, 2024.
\newblock \doi{10.18653/V1/2024.ACL-LONG.702}.
\newblock URL \url{https://doi.org/10.18653/v1/2024.acl-long.702}.

\bibitem[Thakur et~al.(2024)Thakur, Bonifacio, Zhang, Ogundepo, Kamalloo, Alfonso{-}Hermelo, Li, Liu, Chen, Rezagholizadeh, and Lin]{nomiracl}
Nandan Thakur, Luiz Bonifacio, Xinyu Zhang, Odunayo Ogundepo, Ehsan Kamalloo, David Alfonso{-}Hermelo, Xiaoguang Li, Qun Liu, Boxing Chen, Mehdi Rezagholizadeh, and Jimmy Lin.
\newblock "knowing when you don't know": {A} multilingual relevance assessment dataset for robust retrieval-augmented generation.
\newblock In Yaser Al{-}Onaizan, Mohit Bansal, and Yun{-}Nung Chen, editors, \emph{Findings of the Association for Computational Linguistics: {EMNLP} 2024, Miami, Florida, USA, November 12-16, 2024}, pages 12508--12526. Association for Computational Linguistics, 2024.
\newblock URL \url{https://aclanthology.org/2024.findings-emnlp.730}.

\bibitem[Tu et~al.(2025)Tu, Su, Zhou, Liu, and Ai]{tu}
Yiteng Tu, Weihang Su, Yujia Zhou, Yiqun Liu, and Qingyao Ai.
\newblock Rbft: Robust fine-tuning for retrieval-augmented generation against retrieval defects.
\newblock \emph{CoRR}, abs/2501.18365, 2025.
\newblock \doi{10.48550/ARXIV.2501.18365}.
\newblock URL \url{https://doi.org/10.48550/arXiv.2501.18365}.

\bibitem[Wang et~al.(2022)Wang, Yang, Huang, Jiao, Yang, Jiang, Majumder, and Wei]{e5}
Liang Wang, Nan Yang, Xiaolong Huang, Binxing Jiao, Linjun Yang, Daxin Jiang, Rangan Majumder, and Furu Wei.
\newblock Text embeddings by weakly-supervised contrastive pre-training.
\newblock \emph{CoRR}, abs/2212.03533, 2022.
\newblock \doi{10.48550/ARXIV.2212.03533}.
\newblock URL \url{https://doi.org/10.48550/arXiv.2212.03533}.

\bibitem[Wang and Huang(2024)]{back_val}
Sijia Wang and Lifu Huang.
\newblock Targeted augmentation for low-resource event extraction.
\newblock In Kevin Duh, Helena G{\'{o}}mez{-}Adorno, and Steven Bethard, editors, \emph{Findings of the Association for Computational Linguistics: {NAACL} 2024, Mexico City, Mexico, June 16-21, 2024}, pages 4414--4428. Association for Computational Linguistics, 2024.
\newblock \doi{10.18653/V1/2024.FINDINGS-NAACL.275}.
\newblock URL \url{https://doi.org/10.18653/v1/2024.findings-naacl.275}.

\bibitem[Wang et~al.(2023{\natexlab{a}})Wang, Li, Sun, and Liu]{SKR}
Yile Wang, Peng Li, Maosong Sun, and Yang Liu.
\newblock Self-knowledge guided retrieval augmentation for large language models.
\newblock In Houda Bouamor, Juan Pino, and Kalika Bali, editors, \emph{Findings of the Association for Computational Linguistics: {EMNLP} 2023, Singapore, December 6-10, 2023}, pages 10303--10315. Association for Computational Linguistics, 2023{\natexlab{a}}.
\newblock \doi{10.18653/V1/2023.FINDINGS-EMNLP.691}.
\newblock URL \url{https://doi.org/10.18653/v1/2023.findings-emnlp.691}.

\bibitem[Wang et~al.(2023{\natexlab{b}})Wang, Araki, Jiang, Parvez, and Neubig]{FICLO}
Zhiruo Wang, Jun Araki, Zhengbao Jiang, Md.~Rizwan Parvez, and Graham Neubig.
\newblock Learning to filter context for retrieval-augmented generation.
\newblock \emph{CoRR}, abs/2311.08377, 2023{\natexlab{b}}.
\newblock \doi{10.48550/ARXIV.2311.08377}.
\newblock URL \url{https://doi.org/10.48550/arXiv.2311.08377}.

\bibitem[Wei and Zou(2019)]{EDA}
Jason~W. Wei and Kai Zou.
\newblock {EDA:} easy data augmentation techniques for boosting performance on text classification tasks.
\newblock In Kentaro Inui, Jing Jiang, Vincent Ng, and Xiaojun Wan, editors, \emph{Proceedings of the 2019 Conference on Empirical Methods in Natural Language Processing and the 9th International Joint Conference on Natural Language Processing, {EMNLP-IJCNLP} 2019, Hong Kong, China, November 3-7, 2019}, pages 6381--6387. Association for Computational Linguistics, 2019.
\newblock \doi{10.18653/V1/D19-1670}.
\newblock URL \url{https://doi.org/10.18653/v1/D19-1670}.

\bibitem[Wei et~al.(2024)Wei, Chen, and Meng]{instruct_rag}
Zhepei Wei, Wei{-}Lin Chen, and Yu~Meng.
\newblock Instructrag: Instructing retrieval-augmented generation with explicit denoising.
\newblock \emph{CoRR}, abs/2406.13629, 2024.
\newblock \doi{10.48550/ARXIV.2406.13629}.
\newblock URL \url{https://doi.org/10.48550/arXiv.2406.13629}.

\bibitem[Wu et~al.(2024)Wu, Che, Zhang, Tao, Zhang, and Shao]{pandora}
Jinyang Wu, Feihu Che, Chuyuan Zhang, Jianhua Tao, Shuai Zhang, and Pengpeng Shao.
\newblock Pandora's box or aladdin's lamp: {A} comprehensive analysis revealing the role of {RAG} noise in large language models.
\newblock \emph{CoRR}, abs/2408.13533, 2024.
\newblock \doi{10.48550/ARXIV.2408.13533}.
\newblock URL \url{https://doi.org/10.48550/arXiv.2408.13533}.

\bibitem[Xia et~al.(2025)Xia, Zhou, Shi, Chen, and Huang]{cot1}
Yuan Xia, Jingbo Zhou, Zhenhui Shi, Jun Chen, and Haifeng Huang.
\newblock Improving retrieval augmented language model with self-reasoning.
\newblock In Toby Walsh, Julie Shah, and Zico Kolter, editors, \emph{AAAI-25, Sponsored by the Association for the Advancement of Artificial Intelligence, February 25 - March 4, 2025, Philadelphia, PA, {USA}}, pages 25534--25542. {AAAI} Press, 2025.
\newblock \doi{10.1609/AAAI.V39I24.34743}.
\newblock URL \url{https://doi.org/10.1609/aaai.v39i24.34743}.

\bibitem[Xu et~al.(2024)Xu, Pang, Yu, Meng, Shen, Cheng, and Zhou]{INFO-RAG}
Shicheng Xu, Liang Pang, Mo~Yu, Fandong Meng, Huawei Shen, Xueqi Cheng, and Jie Zhou.
\newblock Unsupervised information refinement training of large language models for retrieval-augmented generation.
\newblock In Lun{-}Wei Ku, Andre Martins, and Vivek Srikumar, editors, \emph{Proceedings of the 62nd Annual Meeting of the Association for Computational Linguistics (Volume 1: Long Papers), {ACL} 2024, Bangkok, Thailand, August 11-16, 2024}, pages 133--145. Association for Computational Linguistics, 2024.
\newblock \doi{10.18653/V1/2024.ACL-LONG.9}.
\newblock URL \url{https://doi.org/10.18653/v1/2024.acl-long.9}.

\bibitem[Yan et~al.(2024)Yan, Gu, Zhu, and Ling]{CRAG}
Shi{-}Qi Yan, Jia{-}Chen Gu, Yun Zhu, and Zhen{-}Hua Ling.
\newblock Corrective retrieval augmented generation.
\newblock \emph{CoRR}, abs/2401.15884, 2024.
\newblock \doi{10.48550/ARXIV.2401.15884}.
\newblock URL \url{https://doi.org/10.48550/arXiv.2401.15884}.

\bibitem[Yang et~al.(2024)Yang, Yang, Zhang, Hui, Zheng, Yu, Li, and et~al.]{Qwen-2-5}
An~Yang, Baosong Yang, Beichen Zhang, Binyuan Hui, Bo~Zheng, Bowen Yu, Chengyuan Li, and et~al.
\newblock Qwen2.5 technical report.
\newblock \emph{CoRR}, abs/2412.15115, 2024.
\newblock \doi{10.48550/ARXIV.2412.15115}.
\newblock URL \url{https://doi.org/10.48550/arXiv.2412.15115}.

\bibitem[Yang et~al.(2018)Yang, Qi, Zhang, Bengio, Cohen, Salakhutdinov, and Manning]{HotpotQA}
Zhilin Yang, Peng Qi, Saizheng Zhang, Yoshua Bengio, William~W. Cohen, Ruslan Salakhutdinov, and Christopher~D. Manning.
\newblock Hotpotqa: {A} dataset for diverse, explainable multi-hop question answering.
\newblock In Ellen Riloff, David Chiang, Julia Hockenmaier, and Jun'ichi Tsujii, editors, \emph{Proceedings of the 2018 Conference on Empirical Methods in Natural Language Processing, Brussels, Belgium, October 31 - November 4, 2018}, pages 2369--2380. Association for Computational Linguistics, 2018.
\newblock \doi{10.18653/V1/D18-1259}.
\newblock URL \url{https://doi.org/10.18653/v1/d18-1259}.

\bibitem[Yoran et~al.(2024)Yoran, Wolfson, Ram, and Berant]{yoran}
Ori Yoran, Tomer Wolfson, Ori Ram, and Jonathan Berant.
\newblock Making retrieval-augmented language models robust to irrelevant context.
\newblock In \emph{The Twelfth International Conference on Learning Representations, {ICLR} 2024, Vienna, Austria, May 7-11, 2024}. OpenReview.net, 2024.
\newblock URL \url{https://openreview.net/forum?id=ZS4m74kZpH}.

\bibitem[Yu et~al.(2024)Yu, Zhang, Pan, Cao, Ma, Li, Wang, and Yu]{Chain-of-Note}
Wenhao Yu, Hongming Zhang, Xiaoman Pan, Peixin Cao, Kaixin Ma, Jian Li, Hongwei Wang, and Dong Yu.
\newblock Chain-of-note: Enhancing robustness in retrieval-augmented language models.
\newblock In Yaser Al{-}Onaizan, Mohit Bansal, and Yun{-}Nung Chen, editors, \emph{Proceedings of the 2024 Conference on Empirical Methods in Natural Language Processing, {EMNLP} 2024, Miami, FL, USA, November 12-16, 2024}, pages 14672--14685. Association for Computational Linguistics, 2024.
\newblock URL \url{https://aclanthology.org/2024.emnlp-main.813}.

\bibitem[Zang et~al.(2020)Zang, Qi, Yang, Liu, Zhang, Liu, and Sun]{semantic_unit}
Yuan Zang, Fanchao Qi, Chenghao Yang, Zhiyuan Liu, Meng Zhang, Qun Liu, and Maosong Sun.
\newblock Word-level textual adversarial attacking as combinatorial optimization.
\newblock In Dan Jurafsky, Joyce Chai, Natalie Schluter, and Joel~R. Tetreault, editors, \emph{Proceedings of the 58th Annual Meeting of the Association for Computational Linguistics, {ACL} 2020, Online, July 5-10, 2020}, pages 6066--6080. Association for Computational Linguistics, 2020.
\newblock \doi{10.18653/V1/2020.ACL-MAIN.540}.
\newblock URL \url{https://doi.org/10.18653/v1/2020.acl-main.540}.

\end{thebibliography}
